\documentclass[10pt,twocolumn,letterpaper]{article}
\usepackage[T1]{fontenc}
\usepackage{cvpr}              

\definecolor{cvprblue}{rgb}{0.21,0.49,0.74}
\usepackage[pagebackref,breaklinks,colorlinks,allcolors=cvprblue]{hyperref}

\def\paperID{3302} 
\def\confName{CVPR}
\def\confYear{2025}

\title{SUM Parts: Benchmarking Part-Level Semantic Segmentation of Urban Meshes}

\author{
	Weixiao Gao, Liangliang Nan, Hugo Ledoux\\
	Delft University of Technology\\
	{\tt \small {\{w.gao-1,liangliang.nan,h.ledoux\}@tudelft.nl}}
}

\graphicspath{{pics/}{pics/fig1/}{pics/fig1/icons/}{pics/fig2/}{pics/fig3/}{pics/fig4/}{pics/fig5/}{pics/fig6/}{pics/fig7/}{pics/fig8/}{pics/fig9/}{pics/fig10/}{pics/fig11/}{pics/fig12/}{pics/fig13/}{pics/fig14/}{pics/fig15/}{pics/fig16/}{pics/fig17/}{pics/fig18/}{pics/fig19/}{pics/fig20/}{pics/fig21/}{pics/fig22/}{pics/fig23/}{pics/fig24/}}

\begin{document}
\newlength{\defaultintextsep}
\setlength{\defaultintextsep}{\intextsep}
\newlength{\defaultcolumnsep}
\setlength{\defaultcolumnsep}{\columnsep}
\newlength{\defaulttabcolsep}
\setlength{\defaulttabcolsep}{\tabcolsep}
\setlength{\tabcolsep}{2pt} 

\twocolumn[{%
\renewcommand\twocolumn[1][]{#1}%
\maketitle
\begin{center}
    \centering
    \includegraphics[width=\textwidth]{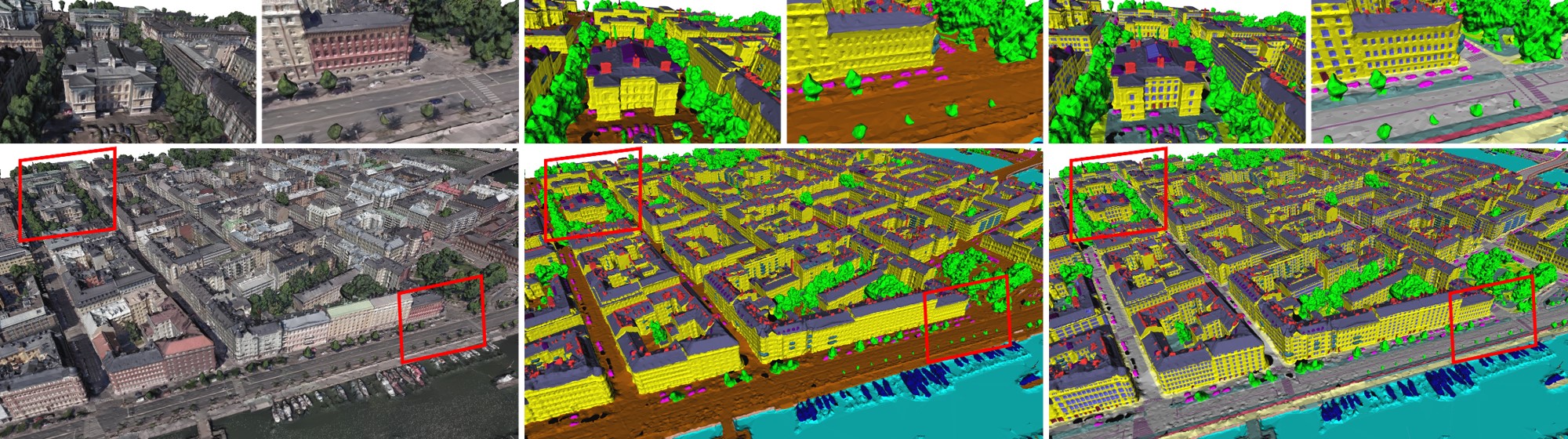}
    \captionof{figure}{SUM Parts provides part-level semantic segmentation of urban textured meshes, covering \(2.5 \, \text{km}^2\) with 21 classes. From left to right: textured mesh, face-based annotations, and texture-based annotations. See~\cref{tab:categories} for class definitions. 
      }
\end{center}%
\label{fig:overview}
}]

\begin{abstract}
Semantic segmentation in urban scene analysis has mainly focused on images or point clouds, while textured meshes—offering richer spatial representation—remain underexplored. This paper introduces SUM Parts, the \textbf{first} large-scale dataset for urban textured meshes with part-level semantic labels, covering about \(2.5\,\text{km}^2\) with 21 classes. The dataset was created using our own annotation tool, which supports both face- and texture-based annotations with efficient interactive selection. We also provide a comprehensive evaluation of 3D semantic segmentation and interactive annotation methods on this dataset. Our project page is available at \href{https://tudelft3d.github.io/SUMParts/}{https://tudelft3d.github.io/SUMParts/}.
\end{abstract}    
\section{Introduction}
\label{sec:intro}
Semantic segmentation is crucial to understanding urban scenes by accurately classifying objects and improving data usability. Recent advances have led to the development of datasets and methods primarily for images and point clouds~\cite{cordts2016cityscapes,martinovic20153d,kitti360_2023}. 
Research on textured meshes has focused mainly on small-scale indoor settings~\cite{s3dis2017,dai2017scannet,chang2017matterport3d}, with limited work on large outdoor environments~\cite{miksik2015semantic,rouhani2017semantic,gao2021sum,weixiao2023pssnet}. 
Furthermore, a critical gap in urban scene understanding is part-level semantic segmentation, which decomposes urban objects into functional components (e.g., windows, chimneys, road markings) following the international CityGML standard~\cite{OGC-CityGML2}.
To address this, we introduce the \textbf{first} large-scale benchmark dataset providing part-level semantic labels for urban textured meshes.

\setlength{\intextsep}{-8pt} 
\setlength{\columnsep}{5pt} 
Obtaining ground truth labels for urban scene understanding often relies on manual annotation, which is time-consuming and expensive~\cite{gao2021sum}. 
Labeling large-scale 3D scenes poses significant challenges compared to 2D image annotation, requiring flexible viewpoint management, specialized interaction methods, and advanced rendering techniques. 
Although considerable research has focused on interactive point cloud annotation~\cite{liu2014new,schmitz2022interactive, kontogianni2023interactive}, efforts to annotate textured meshes~\cite{romanoni2018data,gao2021sum} remain very sparse. Existing studies often label mesh vertices~\cite{dai2017scannet,lang2024iseg} or faces~\cite{romanoni2018data,gao2021sum}, neglecting the richer details provided by textures.
\begin{wrapfigure}{r}{0.4\columnwidth} 
  \begin{center}
  \includegraphics[width=0.4\columnwidth]{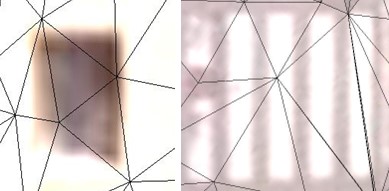} 
  \caption{Mesh textures and wireframes (black).}
  \label{fig:tex_win_road}
  \end{center}
\end{wrapfigure}
Details such as building windows and road markings are better represented in texture images than in unconstrained mesh faces (see~\cref{fig:tex_win_road}). 
To address this, we introduce an efficient interactive tool for textured meshes, enabling both face- and texture-based annotations. 

The main contributions are (1) the first large-scale dataset with part-level semantic labels for urban textured meshes, (2) an efficient interactive annotation tool, and (3) a comprehensive analysis of state-of-the-art 3D semantic segmentation and interactive annotation methods.
\setlength{\intextsep}{\defaultintextsep}
\setlength{\columnsep}{\defaultcolumnsep}
\section{Related work}
\label{sec:related_work}
\textbf{Annotation for semantic segmentation.} Annotation for semantic segmentation is essential in computer vision, leading to the development of various interactive methods for both 2D images and 3D data. 

Early interactive image segmentation methods include region-growing~\cite{adams1994seeded, ning2010interactive}, contour-based~\cite{mortensen1998interactive}, graph-cut~\cite{boykov2001interactive, rother2004grabcut}, and random walk approaches~\cite{grady2006random}. While effective, they often require significant user input or struggle with complex images. Deep learning models like DEXTR~\cite{maninis2018deep} and SAM~\cite{Kirillov_2023_ICCV} have advanced the field but depend on large datasets and may perform poorly on unseen categories. Our interactive annotation method for mesh texture images overcomes these limitations, offering greater generalizability without relying on extensive training data.

Interactive 3D annotation uses data like multi-view images, point clouds, and meshes. Manual methods~\cite{ibrahim2021annotation} are labor-intensive; graph-cut approaches~\cite{liu2014new} depend heavily on user input quality; region-based methods~\cite{gao2021sum} can lead to segmentation errors. Deep learning techniques~\cite{kontogianni2023interactive, yue2023agile3d} require extensive training data and struggle with new data. In contrast, our unsupervised approach requires no prior labeling and operates independently of the original imagery, using approximate user selections and template matching to enhance efficiency and accuracy in complex 3D scenes.

\noindent \textbf{Semantic 3D urban datasets.} 
While several semantic 3D urban datasets exist for LiDAR and photogrammetric point clouds~\cite{hackel2017semantic3d,roynard2018paris,ahn2019,li2020campus3d,can2021semantic,kitti360_2023}, they lack fine-grained part-level semantics essential for comprehensive urban analysis. This is partly due to inherent point cloud limitations, such as low resolution and missing data from occlusions, which hinder capturing detailed structures and small object boundaries~\cite{fei2022comprehensive}. Additionally, point clouds have larger data volumes than textured meshes, leading to longer processing times and higher storage requirements without added informational benefit. In contrast, textured meshes offer better resolution and completeness, but existing mesh datasets~\cite{brostow2009semantic,gao2021sum} typically lack part-level semantic annotations and cover limited categories or scales, focusing on specific objects~\cite{zolanvari2019dublincity,kolle2021h3d}, which limits their applicability. Moreover, existing annotations frequently overlook the rich texture information in meshes. To address these gaps, we introduce the first part-level benchmark dataset of large-scale urban meshes for comprehensive urban analysis.

\section{The SUM parts dataset}\label{sec:sum_dataset}

We aim to use our developed interactive 3D annotation tool to create ground truth for urban textured meshes. Using Helsinki city's mesh~\cite{Helsinki3d} as input, the output includes meshes with face labels and semantic texture masks.
The textured meshes were generated using Bentley's ContextCapture~\cite{contextcap}, reconstructed from oblique aerial imagery with a ground sampling distance of approximately \(7.5\,\text{cm}\).
The annotations were conducted in three representative areas of central Helsinki, comprising 40 tiles of \(62,500\,\text{m}^2\) each, covering a total area of approximately \(2.5\,\text{km}^2\).

\subsection{Annotation}\label{subsec:annotation_tool}
Our annotation aims to achieve precise semantic labeling with significantly improved efficiency for urban meshes.
Our tool features two main modules for part-level semantic annotation: face-based annotation for triangle faces and texture-based annotation for texture pixels. 
We enhance the efficiency of both modules by incorporating interactive selection and template-matching strategies.
We invited five individuals with experience in remote sensing to manually annotate the dataset using our tool: two focused on face-based annotation, two on texture pixel-based annotation, and one reviewed and corrected the annotations. The entire annotation process took approximately 640 hours in total.

\subsubsection{Face-based annotation} 
The face-based annotation aims to assign labels to each face through user interaction, using tools like brushes, strokes, and lassos. To minimize interactions, we developed interactive 3D selection and template-matching algorithms. We first over-segment the mesh into planar segments via region growing, enabling quick selection of large areas, while protrusions are selected semi-automatically based on geometric features. Leveraging the repetitive nature of urban structures, we use structural features for template matching to facilitate rapid annotation.

\textbf{1) Interactive 3D selection.} We propose an interactive protrusion extraction method to address the challenges in over-segmented urban textured meshes, which often struggle with non-planar areas, sharp features, and small-scale structures due to under- or over-segmentation~\cite{weixiao2023pssnet}. Our method aims to efficiently identify protrusions not part of the support plane, similar to foreground-background separation in image segmentation~\cite{rother2004grabcut}. It involves two main steps that work seamlessly to enhance annotation efficiency.
\begin{figure}[!t]
    \resizebox{\columnwidth}{!}{
    \begin{tabular}{cccc}
        \includegraphics[width=0.24\columnwidth]{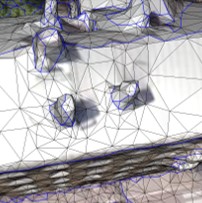} &
        \includegraphics[width=0.24\columnwidth]{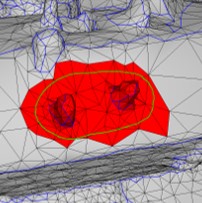} &
        \includegraphics[width=0.24\columnwidth]{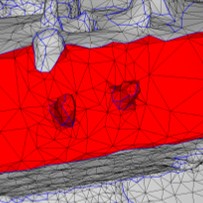} &
        \includegraphics[width=0.24\columnwidth]{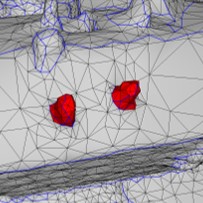} \\
        \includegraphics[width=0.24\columnwidth]{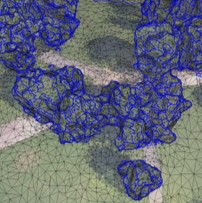} &
        \includegraphics[width=0.24\columnwidth]{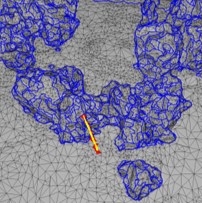} &
        \includegraphics[width=0.24\columnwidth]{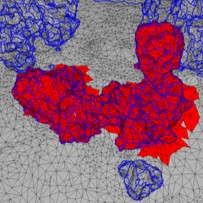} &
        \includegraphics[width=0.24\columnwidth]{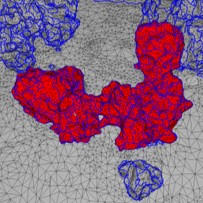} \\
        (a) Input & (b) Lasso/Stroke & (c) Candidates & (d) Protrusions\\
    \end{tabular}
    } 
    \centering
    \caption{Interactive 3D selection. The user performs a lasso (green) or stroke selection (yellow) (b), which generates candidate faces (red) (c). Binary labeling is then applied to these candidate faces to extract protrusions (red) (d). 
    }
    \label{fig:3dsel}
\end{figure}
\indent First, during interactive selection, users employ a lasso or stroke to generate candidate faces for labeling. The algorithm distinguishes these inputs based on the ratio of contour endpoints' distance to the bounding box diagonal and ensures consistent candidate face extraction. For lasso input, all planar segments within the lasso are selected. For stroke input, the selection expands to neighboring faces along the stroke's path, selecting their corresponding planar segments as candidate faces (see~\cref{fig:3dsel}).

Second, we formulate protrusion selection as a binary labeling problem \( l^f = \{\text{support plane}, \text{protrusion}\} \). The candidate segments \( \{f_i\} \) are ordered by area, with the largest serving as the support plane. We construct a dual graph \( \mathcal{G}^f = \{\nu^f, \xi^f\} \) for all candidate faces \(f=\{f_i\}\) each represented by a node in the graph and connected to adjacent faces by graph edges.

The data term \( D^f \) evaluates the likelihood that face \( f_i \) belongs to a protrusion:
\[
D^f(l^f_i) = \eta \times \begin{cases} p_i & \text{if } l^f_i \text{ is support plane}, \\ 1 - p_i & \text{if } l^f_i \text{ is protrusion}, \end{cases}
\]
\noindent where \(\eta\) modulates sensitivity to various geometric characteristics, and \( p_i = d_i + \omega_i \theta_i \) is the protrusion score. Here, \( d_i \) is the maximum distance from face \( f_i \) to the support plane \( P^f_k \); \( \theta_i \) is the minimum angle that quantifies the orientation deviation between the normals of face \( f_i \) and the support plane \( P^f_k \), and \( \omega_i \) is defined as:
\[
\omega_i = \begin{cases} 1 & \text{if } d_i > 1, \\ 1 - d_i & \text{otherwise}. \end{cases}
\]
The smoothness term \( V^f \) measures geometric similarity between adjacent faces:
\[
V^f(l^f_i, l^f_j) = R_{i,j} \cdot 1_{\{l^f_i \neq l^f_j\}},
\]
\noindent where \( z_{\max} \) and \( z_{\min} \) correspond to the range of z-values of all vertices in \( \{f_i\} \), and \( R_{i,j} = 1 - \min\left(1, \dfrac{2|r_i - r_j|}{z_{\max} - z_{\min}}\right) \), and \( r_i \), \( r_j \) are shrinking ball radii computed via the 3D medial axis transform~\cite{sherbrooke1996algorithm}. This accounts for local geometric consistency. 
By combining the above terms, we define the objective function as:
\[
E^f(l^f) = \sum_{i} D^f(l^f_i) + \lambda^f \sum_{\{i,j\}} V^f(l^f_i, l^f_j),
\]
\noindent which we minimize using a graph-cut algorithm~\cite{boykov2001fast}. The parameter \(\lambda^f\) adjusts the weight of the smoothness term, controlling the influence of geometric similarity.

\textbf{2) 3D template matching.}
To leverage repetitive structures in urban scenes and reduce annotation efforts, we employ 3D template matching using structural awareness features from user-selected faces, matching them with similar structures in the scene. This strategy unifies both planar segment and protrusion matching.

\begin{figure}[!t]
     \centering
     \begin{subfigure}{0.49\columnwidth}
         \centering
         \includegraphics[width=\columnwidth]{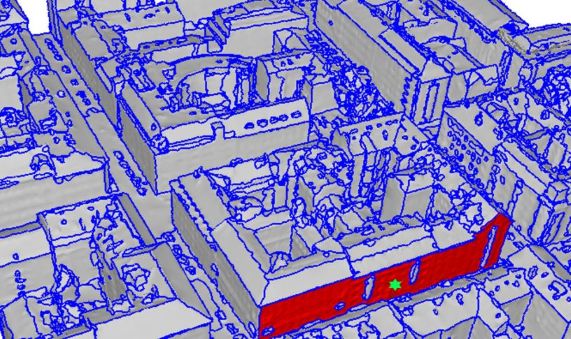}
         \caption{User-selected segment}
         \label{fig:segmatch_a}
     \end{subfigure}
     \begin{subfigure}{0.49\columnwidth}
         \centering
         \includegraphics[width=\columnwidth]{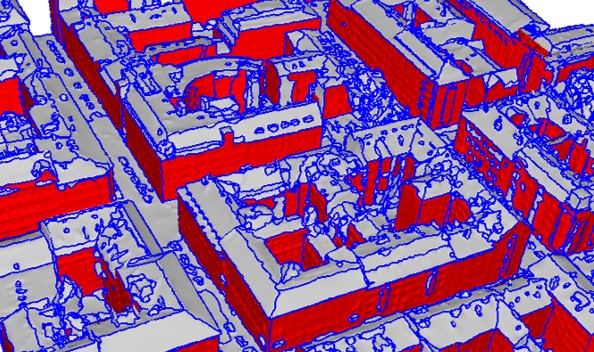}
         \caption{Matched segments}
         \label{fig:segmatch_b}
     \end{subfigure}
     \begin{subfigure}{0.49\columnwidth}
         \centering
         \includegraphics[width=\columnwidth]{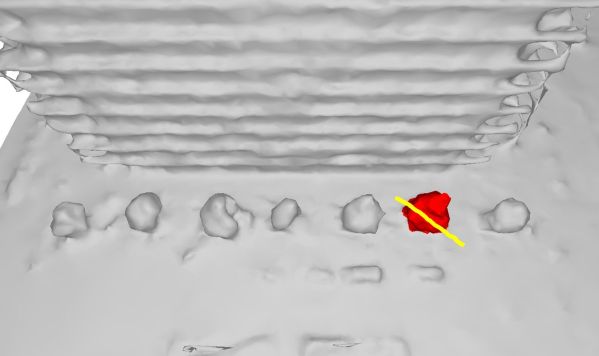}
         \caption{User-selected protrusion}
         \label{fig:promatch_a}
     \end{subfigure}
     \begin{subfigure}{0.49\columnwidth}
         \centering
         \includegraphics[width=\columnwidth]{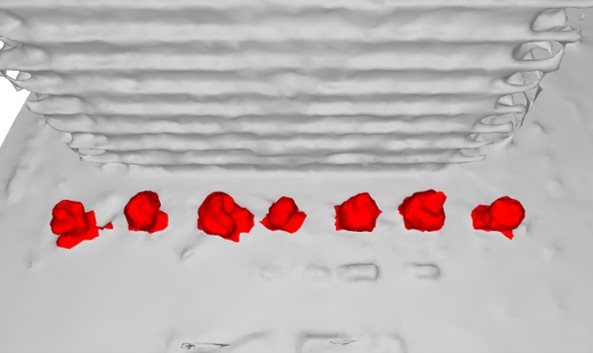}
         \caption{Matched protrusions}
         \label{fig:promatch_b}
     \end{subfigure}
    \caption{3D template matching. When the user selects a planar segment by clicking on it (a), the matched segments are automatically identified (b). A similar matching process also applies to protrusions via a user-drawn stroke, as shown in (c) and (d).}
    \label{fig:3d_match}
\end{figure}
For planar segment matching (see~\cref{fig:segmatch_a} to~\cref{fig:segmatch_b}), we treat the user-selected segment \( P^{(t)} \) as a template and compare it with candidate segments \( \{P^{(c)}_k\} \) based on feature similarity. We assess characteristics such as geometric homogeneity (comparing surface areas), spatial distribution (weighted average heights), orientation (vertical orientations), shape sphericity (based on eigenvalues), and optionally photometric coherence (similarity in color), which constitute a feature vector \( \mathbf{F}^{(\text{seg})} \). A match is determined when the Euclidean norm \( \| \mathbf{F}^{(\text{seg})} \| < \epsilon^{(\text{seg})} \) where \( \epsilon^{(\text{seg})} \) is user-defined depending on input quality.

In protrusion matching (see~\cref{fig:promatch_a} to~\cref{fig:promatch_b}), we use the user-extracted protrusions as templates to find similar structures. We first decompose the template protrusion into planar segments and match them with segments in the scene. The matched segments serve as seeds, which are expanded to neighboring segments to generate candidate regions. We apply spatial and segment scale constraints to limit the expansion:
\[
\left\| O^{(e)}_{k} - O^{(a)}_{j} \right\| < \sqrt{s} \cdot \max_{i} \left\| O^{(t)} - O^{(t)}_{i} \right\|,
\]
\noindent where \( O^{(e)}_{k} \) is the center of the seed segment \( P^{(e)}_k \), \( O^{(a)}_{j} \) is the center of face \( f^{(a)}_j \) in the neighboring segment \( P^{(a)}_k \), \( O^{(t)} \) is the center of the template faces \( f^{(t)} \), \( O^{(t)}_{i} \) is the center of individual template face \( f^{(t)}_i \), and \( s \) is the structural scale parameter controlling the expansion based on the template size. The segment scale constraint ensures that neighboring segments are comparable in size to the templates:
\[
\frac{A^{(a)}_k}{A^{(e)}_k} < s \cdot \frac{\max_j A^{(t)}_j}{\min_j A^{(t)}_j},
\]
\noindent where \( A^{(a)}_k \) and \( A^{(e)}_k \) are the areas of the neighboring segment \( P^{(a)}_k \) and seed segment \( P^{(e)}_k \), respectively, and \( A^{(t)}_j \) are the areas of the template's planar segments.

We then extract candidate protrusions based on structural features such as spatial compactness (comparing the volume occupied by the protrusion relative to its bounding box), surface complexity (assessed by the number of planar segments composing the protrusion), and eigenvalue-based characteristics like linearity, planarity, and sphericity (derived from the covariance of vertex positions). These features form a vector \( \mathbf{F}^{(\text{str})} \), and a match is accepted when the Euclidean norm \(\| \mathbf{F^{(\text{str})}} \| < \epsilon^{(str)}\), where \(\epsilon^{(str)}\) is determined by user interaction and data quality.

\subsubsection{Texture-based annotation}
Mesh textures capture fine details more effectively and avoid the redundancies and discontinuities often found in image-based annotations. However, direct texture annotation is challenging due to discontinuities and computational demands. To address this, we propose a mesh texture annotation strategy based on planar segments, allowing flexible splitting and merging of segments. Our efficient interactive annotation leverages local region extraction and 2D template matching, operating under the assumption that semantic components consist of superpixels with similar geometric and color features.

\begin{figure}[!ht]
     \centering
     \begin{subfigure}{0.49\columnwidth}
         \centering
         \includegraphics[width=\columnwidth]{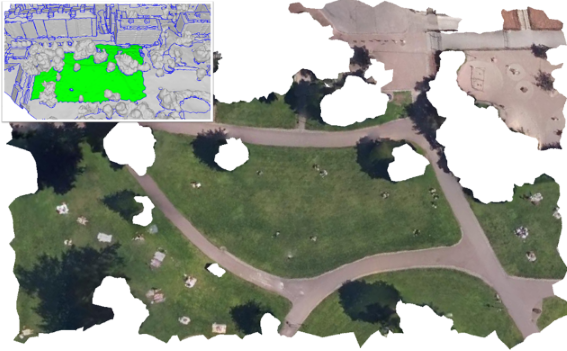}
         \caption{Textured planar segment}
         \label{fig:texsel_a}
     \end{subfigure}
     \begin{subfigure}{0.49\columnwidth}
         \centering
         \includegraphics[width=\columnwidth]{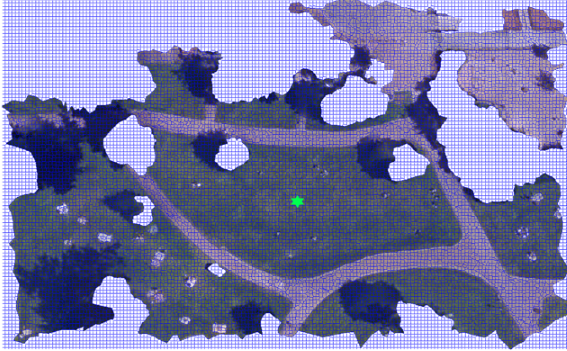}
         \caption{User's click selection}
         \label{fig:texsel_b}
     \end{subfigure}
          \begin{subfigure}{0.49\columnwidth}
         \centering
         \includegraphics[width=\columnwidth]{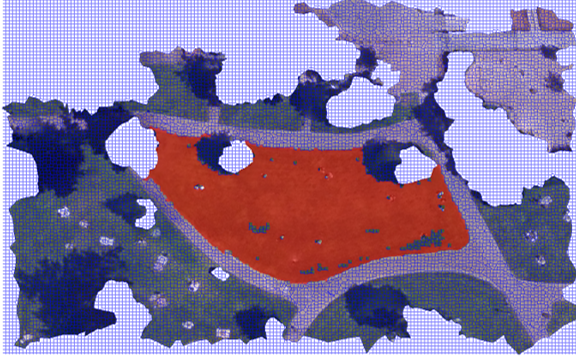}
         \caption{Local expansion}
         \label{fig:texsel_c}
     \end{subfigure}
     \begin{subfigure}{0.49\columnwidth}
         \centering
         \includegraphics[width=\columnwidth]{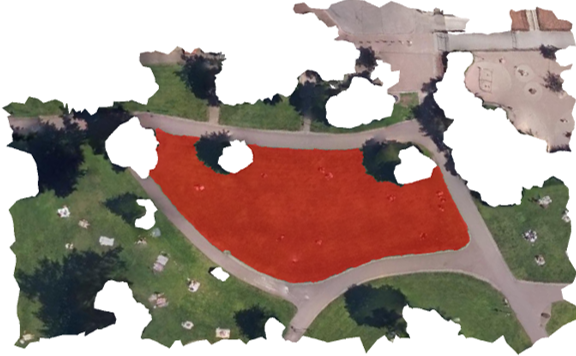}
         \caption{Fine segmentation}
         \label{fig:texsel_d}
     \end{subfigure}
    \caption{Interactive 2D selection. The user selects a texture segment (green) (a). Superpixels are generated (blue), and the user clicks on the region of interest (green star) (b). This triggers local expansion, yielding a coarse segmentation (red) (c), followed by fine segmentation for the final selection (red) (d).}
    \label{fig:2dsel}
\end{figure}

\textbf{1) Interactive 2D selection.}
We aim to capture the region of interest through user clicks. Unlike traditional methods, our approach requires only positive samples. As shown in~\cref{fig:2dsel}, our method consists of two main steps: local expansion and fine segmentation.

In the first local expansion step, we expand the seed superpixels selected by user clicks to encompass the entire area of interest. We first apply Simple Linear Iterative Clustering (SLIC)~\cite{achanta2012slic} to generate homogeneous superpixels from the textured planar segment. We then construct a local adjacency graph \( \mathcal{G}^s = \{\nu^s, \xi^s\} \), where each superpixel is a node connected to its adjacent superpixels. We formulate the expansion as a binary labeling problem \( l^s = \{\text{similar}, \text{non-similar}\} \), aiming to label adjacent superpixels based on their similarity to the initial seed superpixel \( S_0 \). The data term \( D^s \) measures this similarity using the average Wasserstein distance of their Gaussian mixture models (GMMs) over the RGB channels:
\[
D^s(l^s_j) = \alpha \times \begin{cases} 1 - w_j & \text{if } l^s_j \text{ is non-similar}, \\ w_j & \text{if } l^s_j \text{ is similar}, \end{cases}
\]
\noindent where \( w_j = \frac{1}{K} \sum_{k=1}^{K} W(G_k(S_j), G_k(S_0)) \), \( K=3 \) is the number of channels, and \( W \) denotes the Wasserstein distance between the GMMs of the superpixels in each channel. The smoothness term \( V^s \) captures color differences between neighboring superpixels:
\[
V^s(l^s_j, l^s_{j'}) = H_{j,j'} \cdot 1_{\{l^s_j \neq l^s_{j'}\}},
\]
\noindent where \( H_{j,j'} = |\rho_{j} - \rho_{j'}| \), with \( \rho_{j} \) being the color distance (CIEDE2000~\cite{luo2001development}) between superpixel \( S_j \) and the seed \( S_0 \). The energy function combines these terms:
\[
E^s(l^s) = \sum_{j \in \nu^s} D^s(l^s_j) + \lambda^s \sum_{\{j,j'\} \in \xi^s} V^s(l^s_j, l^s_{j'}),
\]
\noindent where \( \lambda^s \) adjusts the weight of \( V^s \). We minimize this energy using graph cuts~\cite{boykov2001fast} to obtain the expanded region.

\setlength{\intextsep}{-10pt} 
\setlength{\columnsep}{5pt} 
Second, we perform fine segmentation to refine the coarse results from the local expansion, 
\begin{wrapfigure}{r}{0.5\columnwidth} 
  \begin{center}
  \includegraphics[width=0.5\columnwidth]{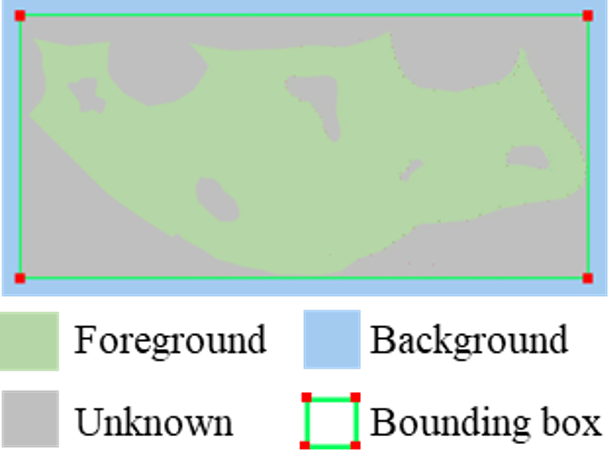} 
  \end{center}
  \label{fig:GrabCut_one}
\end{wrapfigure} 
which may be imprecise at object boundaries due to superpixel resolution and user input. Building on GrabCut~\cite{rother2004grabcut}, we perform detailed pixel-level segmentation using foreground samples while automatically generating background samples from the area beyond the optimal bounding box of the coarse results.
\setlength{\intextsep}{\defaultintextsep}
\setlength{\columnsep}{\defaultcolumnsep}

\begin{figure}[!t]
     \centering
     \begin{subfigure}{0.32\columnwidth}
         \centering
         \includegraphics[width=\columnwidth]{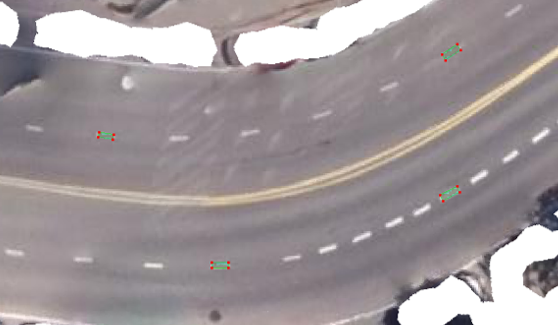}
     \end{subfigure}
     \begin{subfigure}{0.32\columnwidth}
         \centering
         \includegraphics[width=\columnwidth]{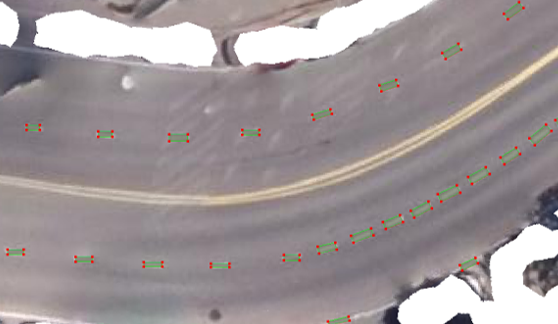}
     \end{subfigure}
     \begin{subfigure}{0.32\columnwidth}
         \centering
         \includegraphics[width=\columnwidth]{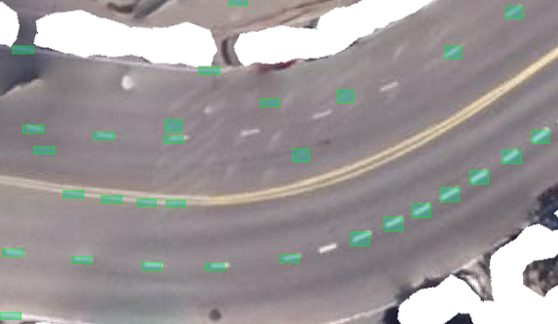}
     \end{subfigure}
     \begin{subfigure}{0.32\columnwidth}
         \centering
         \includegraphics[width=\columnwidth]{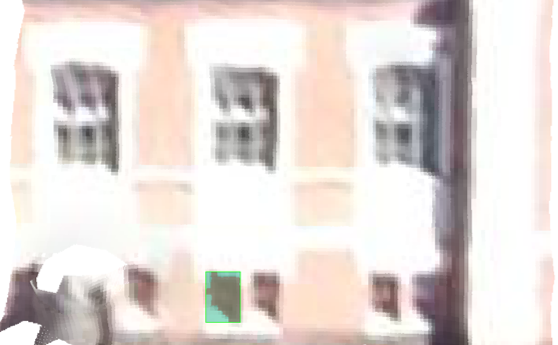}
         \caption{Selected template}
         \label{fig:texmatch_a}
     \end{subfigure}
     \begin{subfigure}{0.32\columnwidth}
         \centering
         \includegraphics[width=\columnwidth]{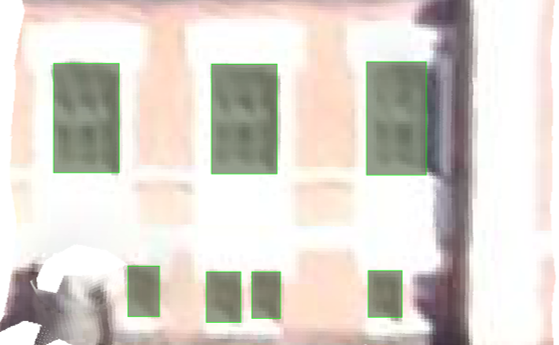}
         \caption{Our matching}
         \label{fig:texmatch_b}
     \end{subfigure}
     \begin{subfigure}{0.32\columnwidth}
         \centering
         \includegraphics[width=\columnwidth]{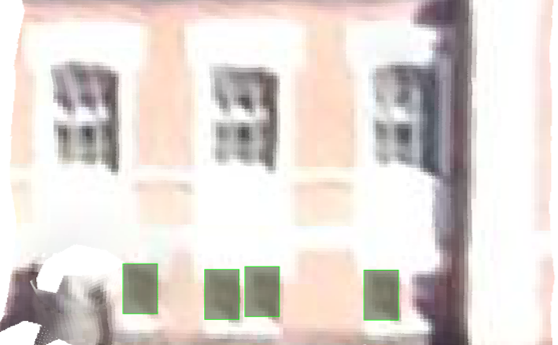}
         \caption{NCC matching}
         \label{fig:texmatch_c}
     \end{subfigure}
    \caption{2D template matching: Extracted regions with green bounding boxes: top shows optimal bounding boxes (rotational invariance), bottom shows vertically aligned bounding boxes (scale invariance) compared to NCC-based methods~\cite{briechle2001template}.}
    \label{fig:tex_match}
\end{figure}

\textbf{2) 2D template matching.}
To improve annotation efficiency for texture images with repetitive structures like windows and road markings, we adopt fast-matching techniques based on 2D structural awareness.
We use user-defined templates to find similar structures within the textured planar segment. The template can be a user-selected region or any arbitrary shape drawn by the user. Normalized Cross-Correlation (NCC)~\cite{briechle2001template} can be used to identify potential matches. However, NCC performs poorly when faced with rotation and scale changes, especially in 3D urban scenes with varying orientations (see~\cref{fig:tex_match}). To overcome this, we propose a region-based template matching approach by extracting structural features from the region \( R^{(t)} \) created by local expansion and matching them to similar regions within the textured planar segments. We calculate the Gaussian Mixture Model (GMM) \( G_k(R^{(t)}) \) of the user-selected region. We then use the Wasserstein distance to filter candidate superpixels:
\[
W(G_k(R^{(t)}), G_k(S^{(c)}_i)) < \epsilon^{\text{seed}},
\]
\noindent where \( \epsilon^{\text{seed}} \) is a user-adjustable threshold, and \( G_k(S^{(c)}_i) \) represents the GMM of candidate superpixel \( S^{(c)}_i \). From the qualified superpixels, we extract candidate regions \( \{ R^{(c)}_i \} \) and compute their similarity to the template region to form a feature vector \( \mathbf{F}^{(\text{reg})} \), which includes: shape index (measuring elongation or flatness), shape regularity (how well the region fills its bounding box), contextual features (similarity in internal and external color distributions). We accept matches where the Euclidean norm \(\| \mathbf{F}^{(\text{reg})}\| < \epsilon^{\text{reg}} \), with \( \epsilon^{\text{reg}} \) depending on user input and image resolution. We also constrain the scale of matching regions using a scaling range based on the number of template pixels:
\[
s^{\text{range}} \in \left[ \frac{N(R^{(t)})}{s^{\text{reg}}}, s^{\text{reg}} \cdot N(R^{(t)}) \right],
\]
\noindent where \( N(R^{(t)}) \) is the number of pixels in the template region, and \( s^{\text{reg}} \) is the scaling factor. Notably, unlike NCC-based template matching, our method offers rotational and scale invariance, allowing it to match targets of varying orientation and size within textured planar segments.

\subsection{Label definition}\label{subsec:label_def}
\begin{figure}[!ht]
\centering
\includegraphics[width=0.47\textwidth]{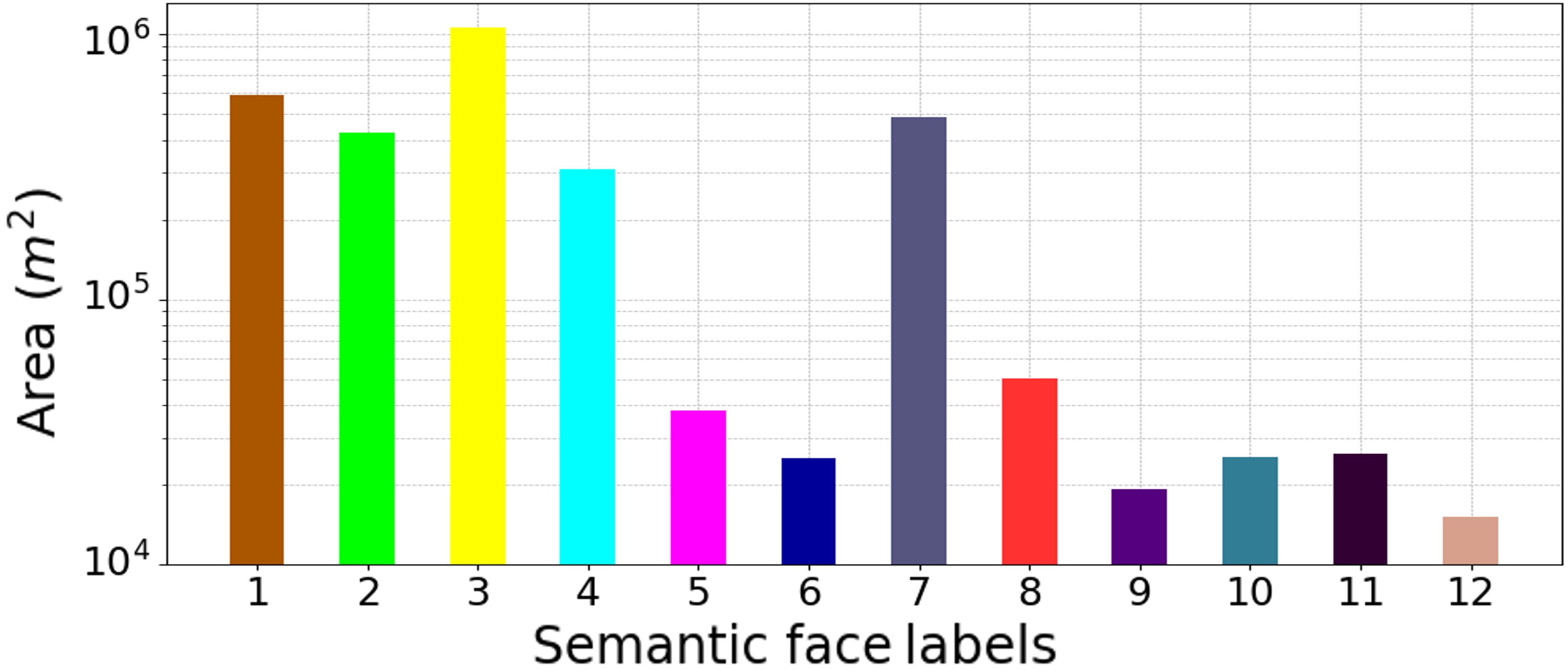}
\includegraphics[width=0.47\textwidth]{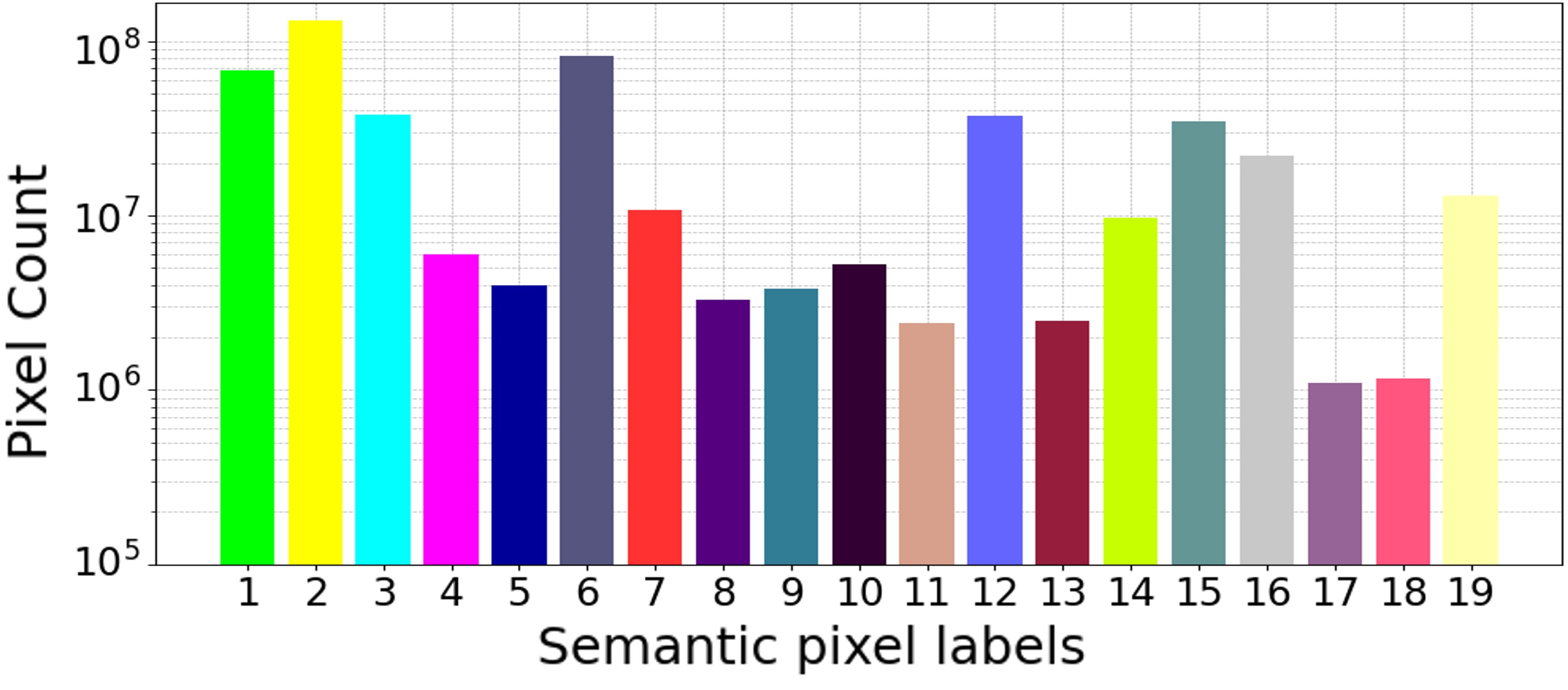} 
\caption{
Statistical distribution of semantic classes across the entire dataset for face (top) and pixel-level labels (bottom).
}
\label{fig:class_dist}
\end{figure}

In the semantic annotation process, we defined two label types: face and pixel labels. 
Each triangle mesh face is assigned one of 13 semantic face labels listed in the top part of~\cref{tab:categories}. Based on these, we defined pixel labels for mesh textures to capture part-level details, introducing 8 new categories shown in the bottom part of~\cref{tab:categories}. We differentiate between two types of labeled scenes: meshes with only face labels, featuring 12 semantic classes (excluding `unclassified'), and meshes with both face and pixel labels, comprising 19 semantic classes (excluding `unclassified', and `terrain' as these are broken down into more specific pixel labels).
\cref{fig:class_dist} show the details of class distribution.
\begin{table}[!t]
\centering
\resizebox{\columnwidth}{!}{
\begin{tabular}{c l p{0.6\columnwidth}}
\hline
\textbf{Color} & \textbf{Name} & \textbf{Explanation} \\
\hline
\raisebox{0ex}{\includegraphics[width=0.01\textwidth]{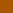}} & \textit{terrain} & Ground surfaces. \\
\raisebox{0ex}{\includegraphics[width=0.01\textwidth]{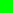}} & \textit{high vegetation} & Tall plants such as trees. \\
\raisebox{0ex}{\includegraphics[width=0.01\textwidth]{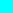}} & \textit{water} & Bodies of water. \\
\raisebox{0ex}{\includegraphics[width=0.01\textwidth]{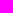}} & \textit{car} & Road vehicles. \\
\raisebox{0ex}{\includegraphics[width=0.01\textwidth]{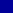}} & \textit{boat} & Watercraft. \\
\raisebox{0ex}{\includegraphics[width=0.01\textwidth]{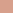}} & \textit{wall} & Vertical barriers. \\
\raisebox{0ex}{\includegraphics[width=0.01\textwidth]{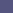}} & \textit{roof surface} & Building roofs. \\
\raisebox{0ex}{\includegraphics[width=0.01\textwidth]{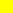}} & \textit{facade surface} & Exterior building walls. \\
\raisebox{0ex}{\includegraphics[width=0.01\textwidth]{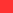}} & \textit{chimney} & Roof vents for smoke. \\
\raisebox{0ex}{\includegraphics[width=0.01\textwidth]{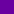}} & \textit{dormer} & Roof projections. \\
\raisebox{0ex}{\includegraphics[width=0.01\textwidth]{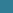}} & \textit{balcony} & Outdoor platforms on buildings. \\
\raisebox{0ex}{\includegraphics[width=0.01\textwidth]{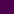}} & \textit{roof installation} & Fixtures on roofs. \\
\midrule
\raisebox{0ex}{\includegraphics[width=0.01\textwidth]{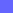}} & \textit{window} & Glass openings in buildings. \\
\raisebox{0ex}{\includegraphics[width=0.01\textwidth]{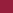}} & \textit{door} & Entrances to buildings. \\
\raisebox{0ex}{\includegraphics[width=0.01\textwidth]{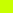}} & \textit{low vegetation} & Short plants like grass. \\
\raisebox{0ex}{\includegraphics[width=0.01\textwidth]{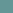}} & \textit{impervious surface} & Non-permeable surfaces. \\
\raisebox{0ex}{\includegraphics[width=0.01\textwidth]{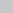}} & \textit{road} & Vehicular paths. \\
\raisebox{0ex}{\includegraphics[width=0.01\textwidth]{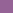}} & \textit{road marking} & Markings on roads. \\
\raisebox{0ex}{\includegraphics[width=0.01\textwidth]{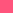}} & \textit{cycle lane} & Bicycle paths. \\
\raisebox{0ex}{\includegraphics[width=0.01\textwidth]{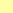}} & \textit{sidewalk} & Pedestrian paths beside roads. \\
\midrule
\raisebox{0ex}{\includegraphics[width=0.01\textwidth]{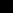}} & \textit{unclassified} & Elements not classified elsewhere. \\
\hline
\end{tabular}
}
\caption{The top 12 are face label definitions, the middle 8 are pixel label definitions, and the last `unclassified' applies to both.}
\label{tab:categories}
\end{table}

\section{Benchmarks}
\subsection{Evaluation of semantic segmentation}\label{subsec:eval_semantic}
\begin{figure}[!t]
    \resizebox{\columnwidth}{!}{
    \begin{tabular}{ccccc}
        \includegraphics[width=0.19\columnwidth]{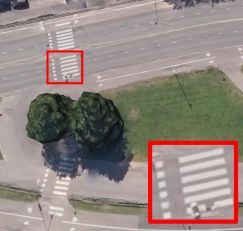} &
        \includegraphics[width=0.19\columnwidth]{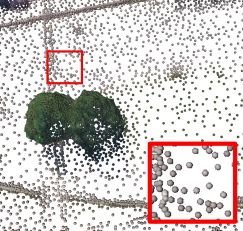} &
        \includegraphics[width=0.19\columnwidth]{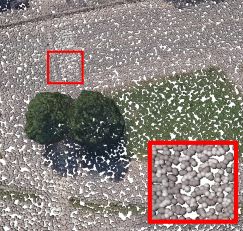} &
        \includegraphics[width=0.19\columnwidth]{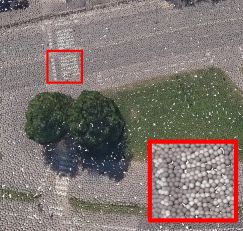} &
        \includegraphics[width=0.19\columnwidth]{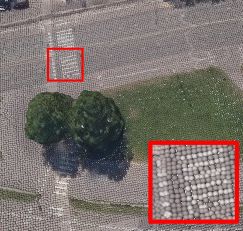} \\
        \includegraphics[width=0.19\columnwidth]{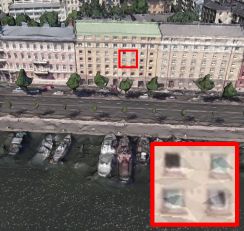} &
        \includegraphics[width=0.19\columnwidth]{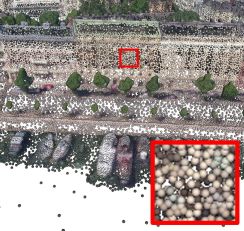} &
        \includegraphics[width=0.19\columnwidth]{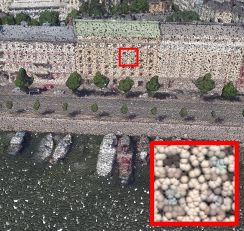} &
        \includegraphics[width=0.19\columnwidth]{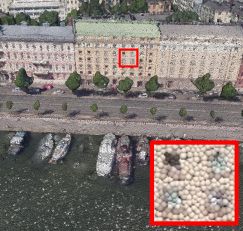} &
        \includegraphics[width=0.19\columnwidth]{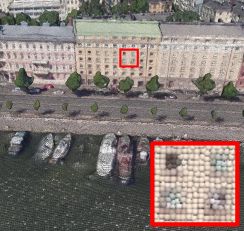} \\
    \end{tabular}
    } 
    \centering
    \caption{Comparison of mesh sampling methods. From left to right: input meshes, face-centered sampling, random sampling, Poisson-disk sampling~\cite{cook1986stochastic}, and our superpixel texture sampling.}
    \label{fig:mesh_sampling_cmp}
\end{figure}

Research shows that semantic segmentation can be performed on point clouds sampled from mesh surfaces~\cite{gao2021sum,selvaraju2021buildingnet}, mapping results back to meshes via nearest neighbor or voting methods. Traditional sampling methods (e.g., face-centered, random, Poisson-disk~\cite{cook1986stochastic}) often miss fine details or exhibit density sensitivity issues in complex environments (see~\cref{fig:mesh_sampling_cmp}). We apply SLIC~\cite{achanta2012slic} over-segmentation on texture images to accurately capture boundaries, then use superpixel centers to generate a texture-based point cloud, ensuring precise semantics, fewer samples, reduced computational load, and efficient pixel-to-point label transfer.

We evaluate state-of-the-art 3D semantic segmentation methods on our datasets, including mesh-based methods like RF-MRF~\cite{rouhani2017semantic}, SUM-RF~\cite{gao2021sum}, and PSSNet~\cite{weixiao2023pssnet}, and point cloud based approaches like PointNet~\cite{qi2017pointnet}, PointNet++~\cite{qi2017pointnet++}, superpoint graphs (SPG)~\cite{landrieu2018large}, SparseConvUnet~\cite{graham20183d}, RandLA-Net~\cite{hu2020randla}, KPConv~\cite{thomas2019kpconv}, PointNext~\cite{qian2022pointnext}, Point Transformer V3 (PointTransV3)~\cite{wu2023point}, and PointVector~\cite{deng2023pointvector}.
It is worth noting that there are currently no semantic segmentation methods specifically designed for meshes with textured-based pixel labels.
These point cloud segmentation methods were evaluated on both the face and pixel labeling tracks. 
We divide our data into three random splits: 24 tiles for training, 8 for validation, and 8 for testing. 
To address the class imbalance, we applied class weights to each method, and to reduce randomness in network predictions, results are averaged over five training runs.

We used established semantic segmentation evaluation metrics, conducting a detailed analysis that included Intersection over Union (IoU) per class, overall accuracy (OA), mean accuracy (mAcc), and mean IoU (mIoU).
For the face labeling track, we determine the final labels of triangles using a voting method based on point cloud prediction results, with the area of each triangle serving as a weight for semantic evaluation metrics.
For the pixel labeling track, we assign final labels to each pixel in the texture image using the nearest neighbor method based on point cloud prediction results, evaluating pixel-level semantic metrics.
\begin{table}[!t]
  \centering
    \begin{tabular}{lccccccc}
    \toprule
    & \multicolumn{4}{c}{Face Labeling} & \multicolumn{3}{c}{Pixel Labeling} \\
    \cmidrule(lr){2-5} \cmidrule(lr){6-8}
    & Fc. & Rd. & Po. & Sp. & Rd. & Po. & Sp.  \\
    \midrule
    PointNet & 5.1  & 8.5  & 5.1  &  \textbf{15.1} & 2.5  & 5.7  & 2.6  \\
    PointNet++ & 27.1  & 17.3  & 18.4  & \textbf{33.1} & 20.6  & 19.5  & \textbf{24.7} \\
    SPG & 29.9  & 29.7  & 31.5  & \textbf{31.7} & 16.0  & 20.0  & \textbf{19.2}\\
    SparseUNet & \textbf{60.5} & 47.6  & 38.6  & 49.9  & \textbf{34.5} & 13.3  & 23.3 \\
    Randla-net & \textbf{57.4} & 49.8  & 49.1  & 54.4 & 36.9  & 39.9  & \textbf{42.1} \\
    KPConv & \textbf{57.5} & 56.4  & 46.5  & 52.9& 38.9  & 26.1  & \textbf{42.6} \\
    PointNext & \textbf{65.3} & 51.3  & 50.4  & 47.7 & 42.9  & \textbf{44.7} & 43.0 \\
    PointTransV3 & \textbf{59.1} & 49.5  & 51.7  & 54.0 & \textbf{38.0} & 36.1  & 37.8 \\
    PointVector & \textbf{70.0} & 56.1  & 52.8  & 57.1 & 44.7  & 45.1  & \textbf{47.9}\\
    Average (mIoU) & \textbf{48.0} & 40.7  & 38.2  & 44.0 & 30.6  & 27.8  & \textbf{31.5} \\
    \bottomrule
    \end{tabular}%
  \caption{Evaluation against different sampling strategies on semantic segmentation using mIoU. `Fc.' represents face-centered sampling, `Rd.' random sampling, `Po.' Poisson-disk sampling~\cite{cook1986stochastic}, and `Sp.' superpixel texture sampling. The highest and average mIoUs for each method are highlighted in bold.
  }
  \label{tab:meshsampling}
\end{table}%

\noindent \textit{\textbf{1) Face labeling track.}}
The face labeling track includes 12 labels, excluding `unclassified'. 
We evaluated the impact of four mesh point cloud sampling strategies on semantic segmentation. To control point cloud density, the number of points for random and Poisson-disk samples matched the superpixel texture sample size, while face-centered samples always matched the number of mesh faces.

\begin{figure}[!t]
    \resizebox{\columnwidth}{!}{
    \begin{tabular}{ccccc}
        \includegraphics[width=0.19\columnwidth]{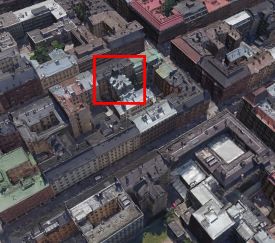} &
        \includegraphics[width=0.19\columnwidth]{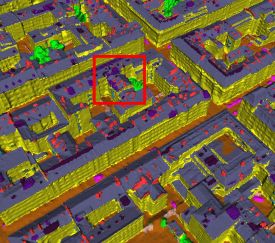} &
        \includegraphics[width=0.19\columnwidth]{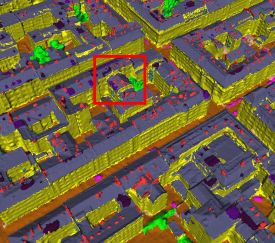} &
        \includegraphics[width=0.19\columnwidth]{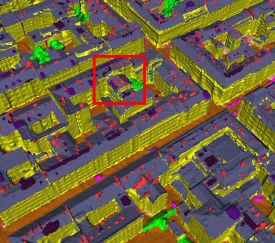} &
        \includegraphics[width=0.19\columnwidth]{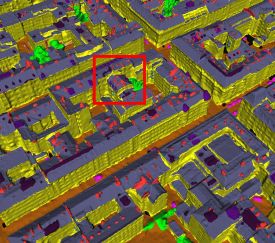} \\
        \includegraphics[width=0.19\columnwidth]{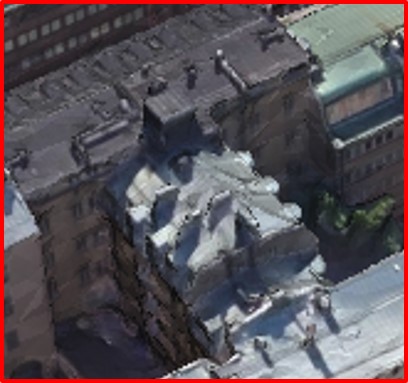} &
        \includegraphics[width=0.19\columnwidth]{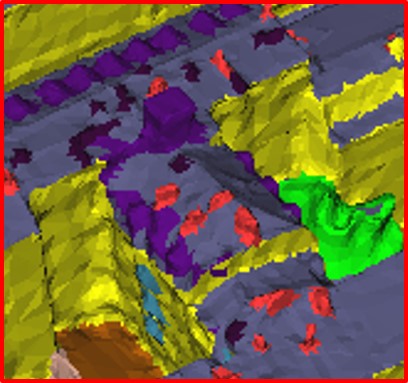} &
        \includegraphics[width=0.19\columnwidth]{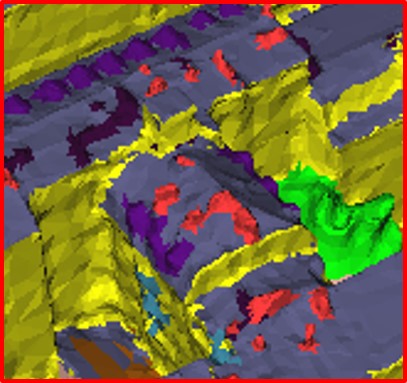} &
        \includegraphics[width=0.19\columnwidth]{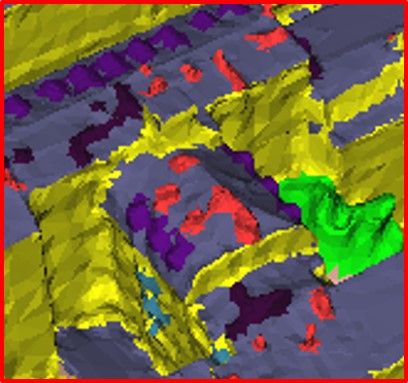} &
        \includegraphics[width=0.19\columnwidth]{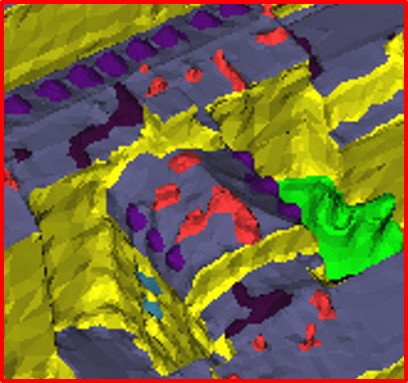} \\
        Input &
		SpUNet\(^{Fc.}\) &
		PtNext\(^{Fc.}\) &
		PtVector\(^{Fc.}\) &
        Truth\\	
        \includegraphics[width=0.19\columnwidth]{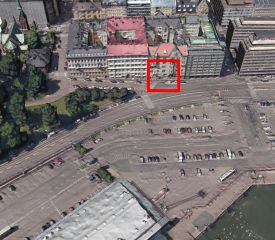} &
        \includegraphics[width=0.19\columnwidth]{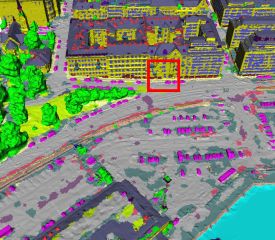} &
        \includegraphics[width=0.19\columnwidth]{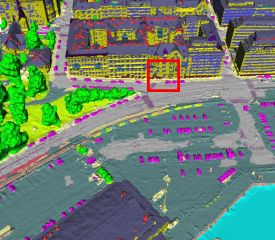} &
        \includegraphics[width=0.19\columnwidth]{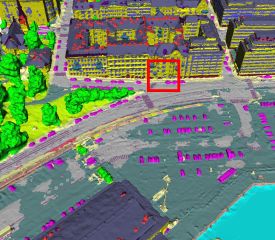} &
        \includegraphics[width=0.19\columnwidth]{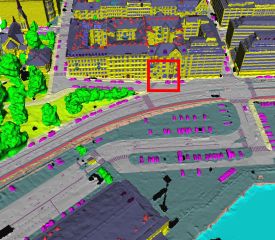} \\
        \includegraphics[width=0.19\columnwidth]{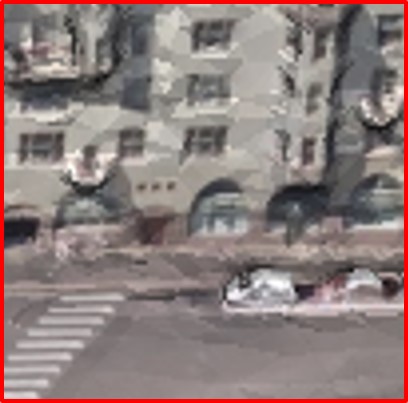} &
        \includegraphics[width=0.19\columnwidth]{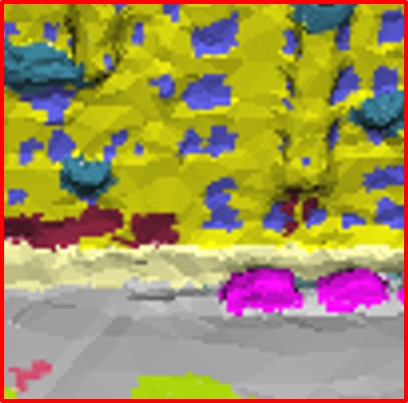} &
        \includegraphics[width=0.19\columnwidth]{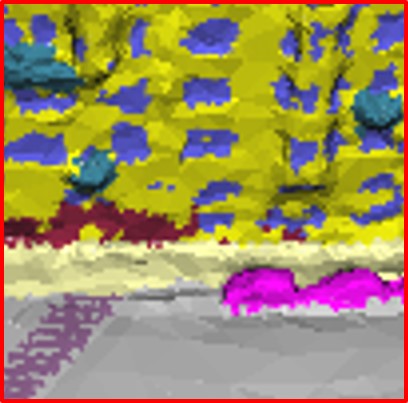} &
        \includegraphics[width=0.19\columnwidth]{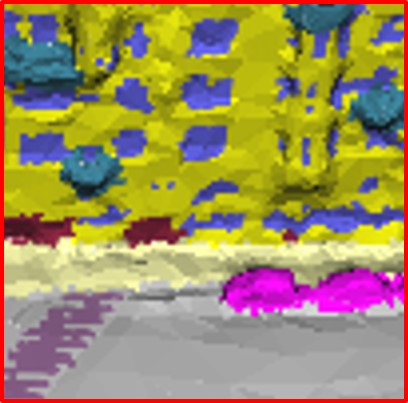} &
        \includegraphics[width=0.19\columnwidth]{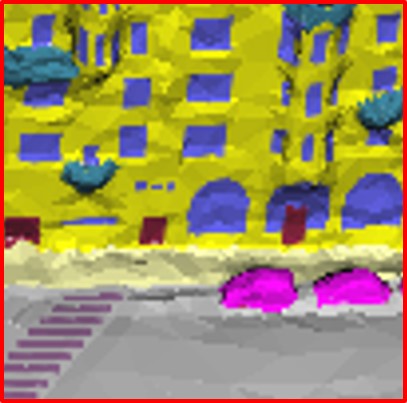} \\
        Input &
		KPConv\(^{Sp.}\) &
		PtNext\(^{Po.}\) &
		PtVector\(^{Sp.}\) &
        Truth\\	
    \end{tabular} 
    }
    \centering
    \caption{Qualitative analysis of face and pixel labeling. The top two rows show the three best methods for face labeling: SparseUNet\(^{Fc.}\)~\cite{graham20183d}, PointNext\(^{Fc.}\)~\cite{qian2022pointnext}, and PointVector\(^{Fc.}\)~\cite{deng2023pointvector}. The bottom two rows show the three best methods for pixel labeling: KPConv\(^{Sp.}\)~\cite{thomas2019kpconv}, PointNext\(^{Po.}\)~\cite{qian2022pointnext}, and PointVector\(^{Sp.}\)~\cite{deng2023pointvector}. `Fc.' represents face-centered sampling, `Po.' Poisson-disk sampling~\cite{cook1986stochastic}, and `Sp.' superpixel texture sampling.}
    \label{fig:tri_tex_pixel_top3}
\end{figure}

\cref{tab:meshsampling} shows that using face-centered point clouds leads to optimal performance for most methods. This is because they adapt well to the geometric characteristics of triangulated meshes, where triangle density is lower in flat areas and higher in non-flat areas. Such a distribution enables deep learning networks to learn rich geometric features.
However, this does not apply to uniform triangular meshes, indicating that the impact of point cloud sampling density on semantic segmentation is far less significant than the impact of point cloud distribution.
The average mIoU in~\cref{tab:meshsampling} also indicates that our proposed superpixel texture sampling method outperforms other mesh sampling methods (except for face-centered point clouds). 
~\cref{tab:semantic_eval} shows that PointVector~\cite{deng2023pointvector} surpasses all competing methods with the highest mAcc of 80.7\% and mIoU of 70.0\%.
~\cref{fig:tri_tex_pixel_top3} presents a qualitative analysis of the top three methods.
\begin{table}[!t]
  \centering
    \begin{tabular}{llcclcc}
    \toprule
    & \multicolumn{3}{c}{Face Labeling Track} & \multicolumn{3}{c}{Pixel Labeling Track} \\
    \cmidrule(lr){2-4} \cmidrule(lr){5-7}
    &Sa.& mAcc & mIoU & Sa. & mAcc & mIoU \\
    \midrule
    RF\_MRF & - & 45.3  & 39.5 & - & - & - \\
    SUM\_RF & - & 53.6  & 46.0 & - & - & -  \\
    PSSNet & - & 56.4  & 47.0  & - & - & - \\
    PointNet & Sp. & 22.0  & 15.1 & Sp. & 9.8  & 2.6 \\
    PointNet++ & Sp. & 46.9  & 33.1 & Sp. & 35.2  & 24.7 \\
    SPG & Sp. & 55.0  & 31.7 & Sp. & 34.5 & 19.2 \\
    SparseUNet & Fc. & 71.7  & 60.5 & Rd. & 45.1 & 34.5 \\
    Randla-net & Fc. & 76.3  & 57.4 & Sp. & 57.7  & 42.1 \\
    KPConv & Fc. & 64.7  & 57.5 & Sp. & 58.3 & 42.6 \\
    PointNext & Fc. & 77.2  & 65.3 & Po. & 57.6  & 44.7 \\
    PointTransV3 & Fc. & 70.2  & 59.1 & Rd. & 54.1  & 38.0 \\
    PointVector & Fc. & \textbf{80.7} & \textbf{70.0} & Sp. & \textbf{63.8} & \textbf{47.9} \\
    \bottomrule
    \end{tabular}%
  \caption{Evaluation of semantic segmentation performance for face labeling and pixel labeling tracks. `Sa.' represents sampling methods, including `Fc.' for face-centered sampling, `Rd.' for random sampling, `Po.' for Poisson-disk sampling~\cite{cook1986stochastic}, and `Sp.' for superpixel texture sampling. `-' indicates not applicable. }
  \label{tab:semantic_eval}
\end{table}%

\noindent \textit{\textbf{2) Pixel labeling track.}}
The pixel labeling track consists of 19 semantic classes, as described in~\cref{subsec:label_def}, including all semantic labels except for `terrain' and `unclassified'.
We evaluated the impact of three sampling methods—random, Poisson-disk, and superpixel texture sampling—on semantic segmentation performance.
Because face-centered point clouds cannot represent semantic components with pixel labels, they were excluded from testing.

The average mIoU in~\cref{tab:meshsampling} shows most methods achieve optimal performance with our proposed sampling, which precisely captures the boundaries of part-level objects.
~\cref{tab:semantic_eval} shows that PointVector~\cite{deng2023pointvector} continues to outperform all other methods, achieving a mIoU of 47.9\%, significantly higher than the others.
We also conducted a qualitative analysis of the top three methods, shown in~\cref{fig:tri_tex_pixel_top3}.

\subsection{Evaluation of interactive annotation}\label{subsec:eval_annot}
\begin{figure}[!t]
    \begin{tabular}{cccc}
        \includegraphics[width=0.22\columnwidth]{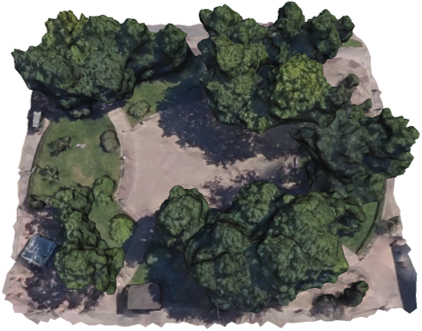} &
        \includegraphics[width=0.22\columnwidth]{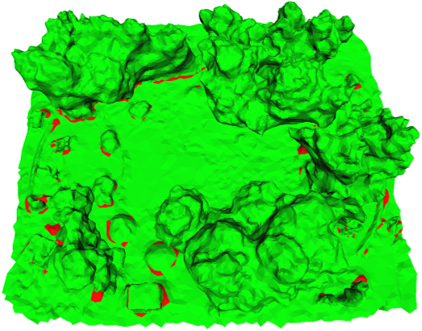} &
        \includegraphics[width=0.22\columnwidth]{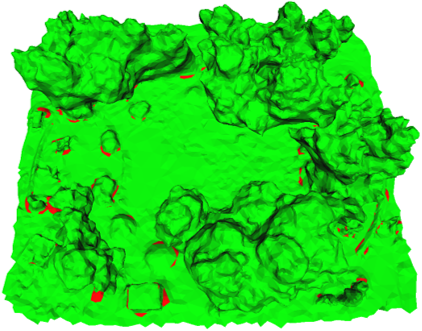} &
        \includegraphics[width=0.22\columnwidth]{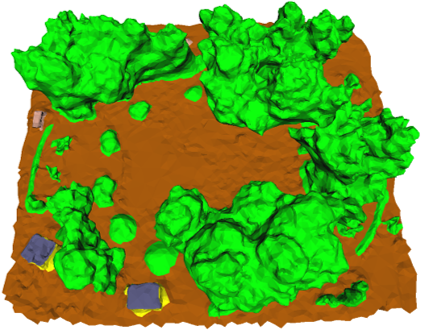} \\
		Input & Segment-based & Ours & Manual \\
    \end{tabular}
    \centering
    \caption{Errors (red) in interactive face annotation. }
    \label{fig:annot_tri_errs}
\end{figure}

We evaluated existing interactive annotation methods for meshes and textured images in real-world scenarios and proposed new evaluation criteria based on user studies. Traditional metrics like the number of clicks to achieve a certain IoU or Average Precision (AP) do not fully capture annotation efficiency due to limitations in evaluation comprehensiveness, interaction complexity, and efficiency measurement. To address these issues, we developed new evaluation metrics including average mean IoU (\( \overline{M} \)), average boundary mean IoU (\( \overline{B} \)), average number of mean user interactions (\(\overline{O}\)), average user annotation time (\(\overline{T}(s)\)), and average percentage of using smart interaction tools (\(\overline{S}(\%)\)). 
\begin{table}[!t]
    \begin{tabular}{lccccc}
    \toprule
      & \(\overline{M}(\%)\) & \(\overline{B}(\%)\) & \(\overline{O}\)& \(\overline{T}(s)\) & \(\overline{S}(\%)\) \\
    \midrule
    Manual & - & - & 6754  & 4183.2 & -\\
    Segment-based & 91.6 & 74.7 & 6119  & 3565.8 & -\\
    Ours & \textbf{92.2} & \textbf{75.0} & \textbf{2992} & \textbf{2992.1} & \textbf{83.0}\\
    \midrule
    Manual & - & - & 653  & 777.1  & - \\
    GrabCut & \textbf{88.1} & 47.1 & 711  & 780.9  & 30.2 \\
    SAM & 84.4 & 29.8 & 716  & \textbf{640.6 } & \textbf{71.7} \\
    SimpleClick & 81.4 & 30.9 & \textbf{252 } & 861.3  & - \\
    Ours & 87.9 & \textbf{49.3} & 582  & 663.3  & 40.3 \\
    \bottomrule
    \end{tabular}%
    \centering
    \caption{Comprehensive performance evaluation of interactive face (top) and texture image annotation (bottom) methods across different test scenarios. The highest values are given in bold.
    }
    \label{tab:annot_eval}
\end{table}%
\indent In our experiments, five users were invited to annotate four representative scenes for face annotation and six for texture annotation using various methods. For interactive face annotation, we compared our method with manual~\cite{rouhani2017semantic} and segment-based interactive annotation~\cite{gao2021sum}. Our method outperformed both in all metrics, reducing the number of interactions and annotation time significantly—about 1.73 times faster than manual annotation and 1.32 times faster than segment-based methods (see~\cref{tab:annot_eval} top and~\cref{fig:annot_tri_errs}). The smart interaction ratio exceeded 80\% in most scenarios, indicating minimal manual intervention was needed. 

\begin{figure}[!t]
    \resizebox{\columnwidth}{!}{
    \begin{tabular}{cccccc}
        \includegraphics[width=0.2\columnwidth]{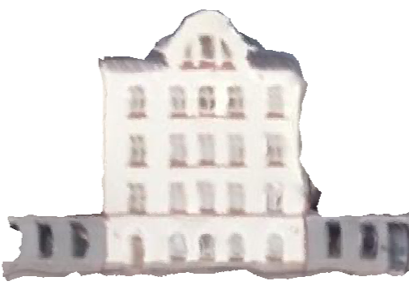} &
        \includegraphics[width=0.2\columnwidth]{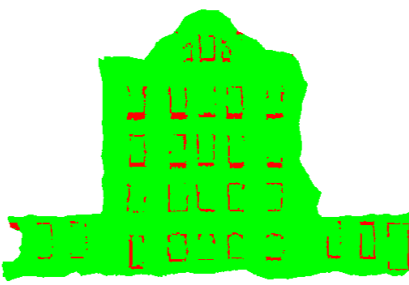} &
        \includegraphics[width=0.2\columnwidth]{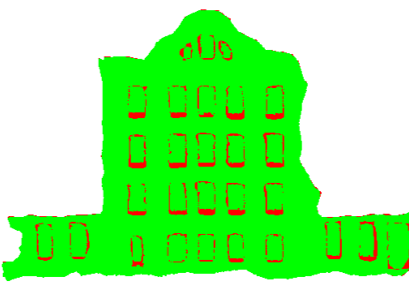} &
        \includegraphics[width=0.2\columnwidth]{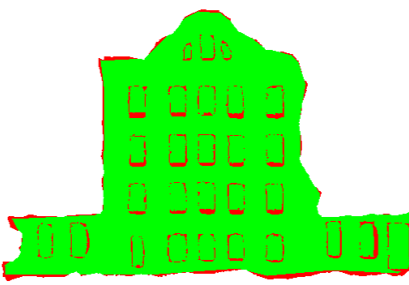} &
        \includegraphics[width=0.2\columnwidth]{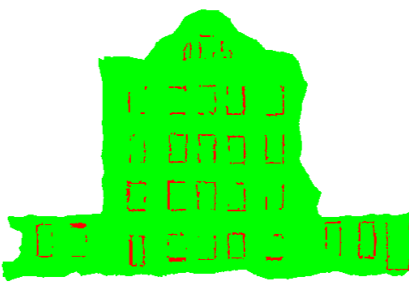} &
        \includegraphics[width=0.2\columnwidth]{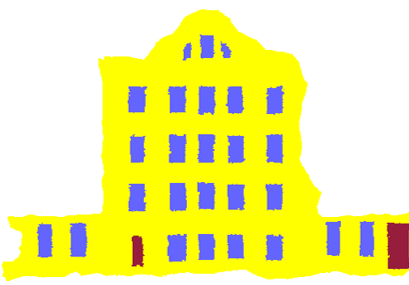} \\
        \includegraphics[width=0.2\columnwidth]{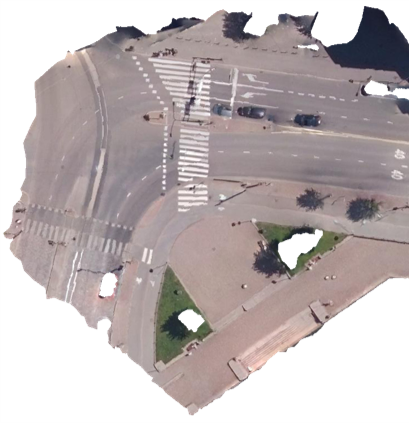} &
        \includegraphics[width=0.2\columnwidth]{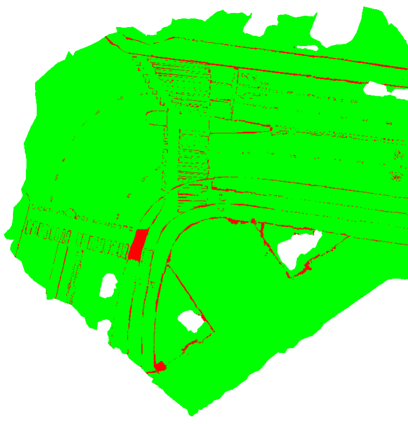} &
        \includegraphics[width=0.2\columnwidth]{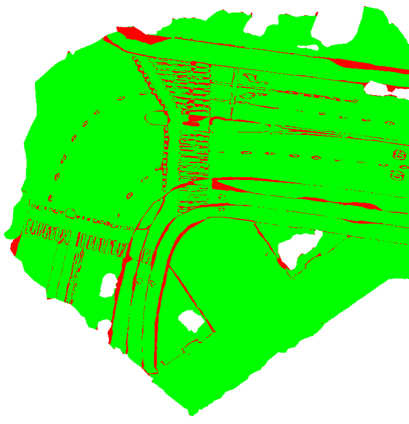} &
        \includegraphics[width=0.2\columnwidth]{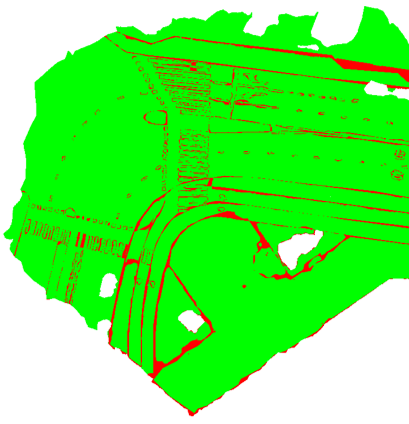} &
        \includegraphics[width=0.2\columnwidth]{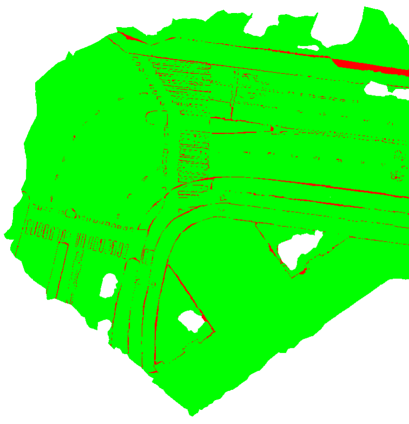} &
        \includegraphics[width=0.2\columnwidth]{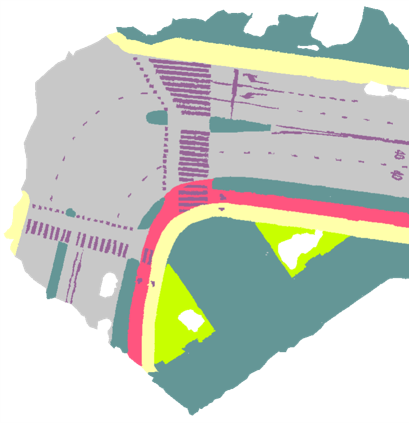} \\
		Input & GrabCut & Simclick & SAM & Ours & Manual \\
    \end{tabular}
    }
    \centering
    \caption{Boundary errors (red) in interactive texture annotation. }
    \label{fig:bound_errs}
\end{figure}

\indent For interactive texture annotation, we compared with manual annotation, GrabCut~\cite{rother2004grabcut}, SAM~\cite{Kirillov_2023_ICCV}, and SimpleClick~\cite{liu2023simpleclick}. Our method surpassed deep learning methods in annotation quality and was comparable to the fastest annotation times of SAM (see~\cref{tab:annot_eval} bottom and~\cref{fig:bound_errs}). It excelled in boundary accuracy due to our fine segmentation and template matching. 
Notably, for objects with regular shapes or repetitive structures, interactive clicking can be inefficient, requiring users to manually trace shape boundaries for higher accuracy. Our method addresses this by enabling efficient annotation of similar structures using reusable templates, reducing repetitive interactions. 

\subsection{Sensitivity analysis}
Conducting ablation studies on energy function-based methods is challenging due to the high interdependence among terms and the complexity of the optimization, complicating the evaluation of individual parts. We conducted a qualitative analysis to assess the impact of parameter adjustments and data quality on our method's outcomes. 

\begin{figure}[!t]
    \resizebox{\columnwidth}{!}{
    \begin{tabular}{cccccc}
        \includegraphics[width=0.2\columnwidth]{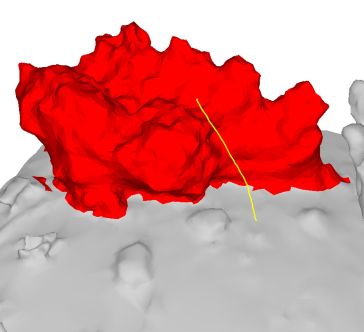} &
        \includegraphics[width=0.2\columnwidth]{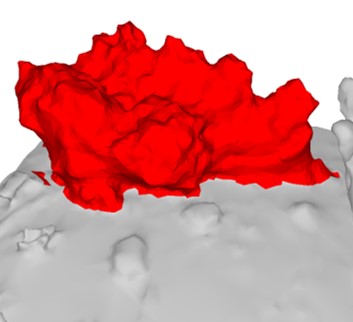} &
        \includegraphics[width=0.2\columnwidth]{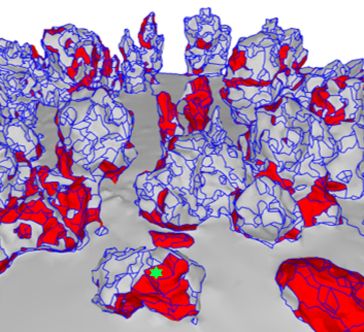} &
        \includegraphics[width=0.2\columnwidth]{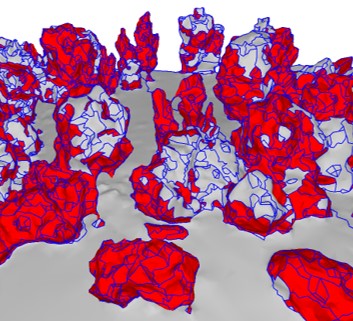} &
        \includegraphics[width=0.2\columnwidth]{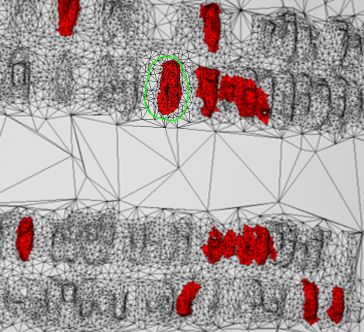} &
        \includegraphics[width=0.2\columnwidth]{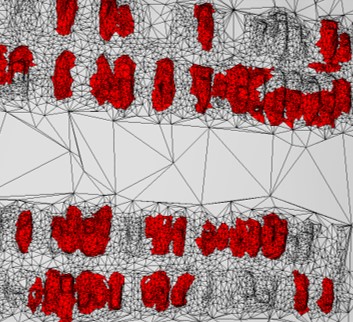} \\
		\(\lambda^f= 0.3\)  & \(\lambda^f= 0.6\)  &
		\(\epsilon^{(seg)}=30\) & \(\epsilon^{(seg)}=80\) &
		\(\epsilon^{(str)}=20\) & \(\epsilon^{(str)}=80\) \\
        \includegraphics[width=0.2\columnwidth]{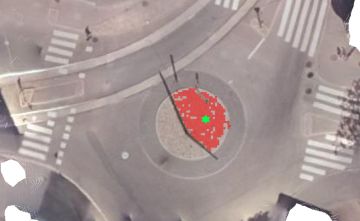} &
        \includegraphics[width=0.2\columnwidth]{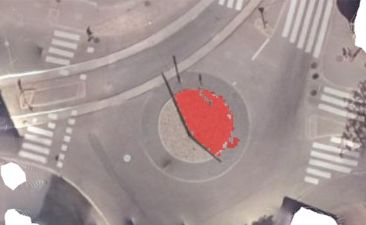} &
        \includegraphics[width=0.2\columnwidth]{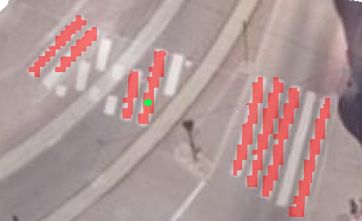} &
        \includegraphics[width=0.2\columnwidth]{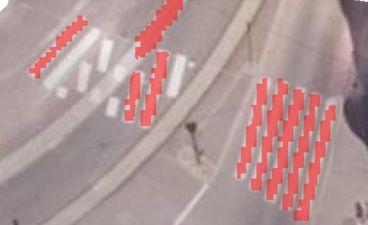} &
        \includegraphics[width=0.2\columnwidth]{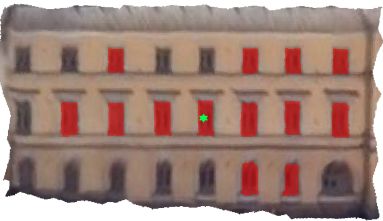} &
        \includegraphics[width=0.2\columnwidth]{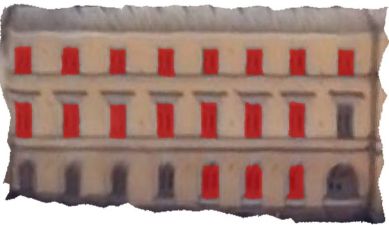} \\
		\(\lambda^s= 0.2\)  & \(\lambda^s= 0.4\)  &
		\(\epsilon^{(seed)}=15\) & \(\epsilon^{(seed)}=30\) &
		\(\epsilon^{(reg)}=30\) & \(\epsilon^{(reg)}=60\) \\
    \end{tabular}
    }
    \centering
    \caption{Sensitivity analysis of parameters in face (top) and texture annotation (bottom) with the same user interactions. }
    \label{fig:params_val}
\end{figure}

\begin{figure}[t!]
    \resizebox{\columnwidth}{!}{
    \begin{tabular}{cccc}
        \includegraphics[width=0.25\columnwidth]{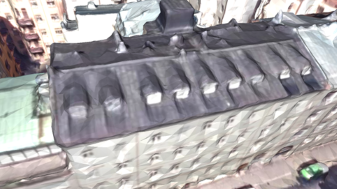} &
        \includegraphics[width=0.25\columnwidth]{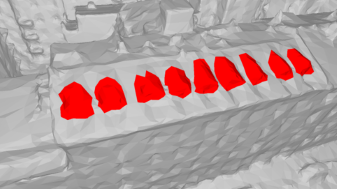} &
        \includegraphics[width=0.25\columnwidth]{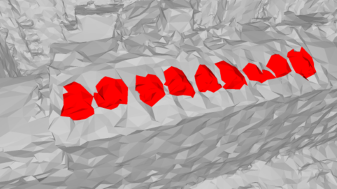} &
        \includegraphics[width=0.25\columnwidth]{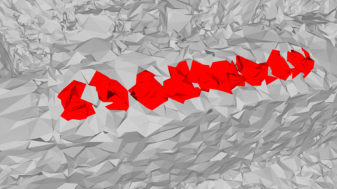} \\
		Mesh  & \(\sigma^m= 0.0\) & \(\sigma^m= 0.2\)& \(\sigma^m= 0.4\) \\
        \includegraphics[width=0.25\columnwidth]{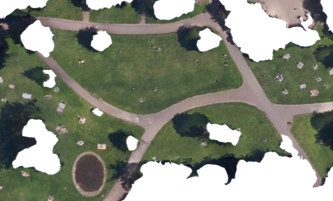} &
        \includegraphics[width=0.25\columnwidth]{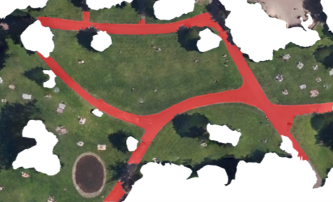} &
        \includegraphics[width=0.25\columnwidth]{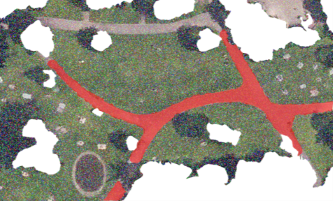} &
        \includegraphics[width=0.25\columnwidth]{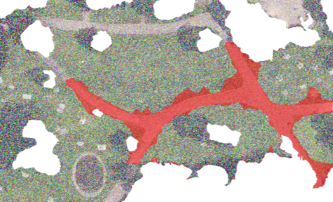} \\
		Texture image & \(\sigma^t= 0\) & \(\sigma^t= 1\) & \(\sigma^t= 3\) \\
    \end{tabular}
    }
    \centering
    \caption{Sensitivity analysis of data quality in protrusion extraction (top) and local region extraction (bottom). \(\sigma^m\) and \(\sigma^t\) represent the standard deviations of Gaussian noise of different inputs.}
    \label{fig:noise_eval}
\end{figure}

In face-based analysis, increasing the balance parameter \( \lambda^f \) improves object boundary clarity while adjusting segment matching thresholds (\( \epsilon^{(\text{seg})} \) and \( \epsilon^{(\text{str})} \)) enhances coverage of repetitive structures (see~\cref{fig:params_val} top). Our method demonstrates significant noise tolerance, accurately extracting protrusions despite added Gaussian noise, which is crucial for effective template matching (see~\cref{fig:noise_eval} top).

In texture-based analysis, a larger balance parameter \( \lambda^s \) smooths local regions, while higher thresholds (\( \epsilon^{\text{seed}} \) and \( \epsilon^{\text{reg}} \)) increase region matches but may reduce seed quality (see~\cref{fig:params_val} bottom). Default parameters perform well with minimal adjustments. Our method, using few interpretable parameters, often outperforms deep learning approaches that rely on inconsistent user clicks, effectively identifying target regions even under high noise and varying user interactions  (see~\cref{fig:noise_eval} bottom).
\section{Summary}
\begin{figure}[!t]
    \resizebox{\columnwidth}{!}{
    \begin{tabular}{cccc}
        \includegraphics[width=0.25\columnwidth]{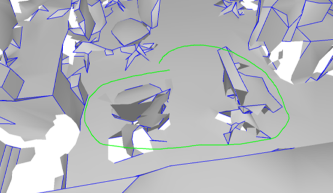} &
        \includegraphics[width=0.25\columnwidth]{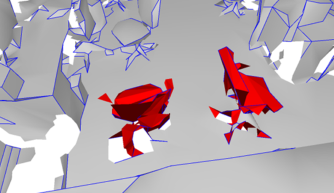} &
        \includegraphics[width=0.25\columnwidth]{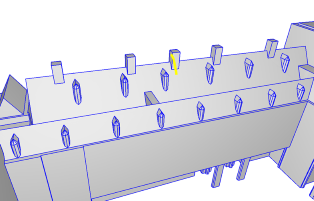} &
        \includegraphics[width=0.25\columnwidth]{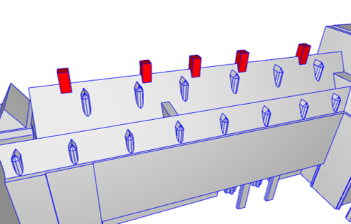} \\
        \includegraphics[width=0.25\columnwidth]{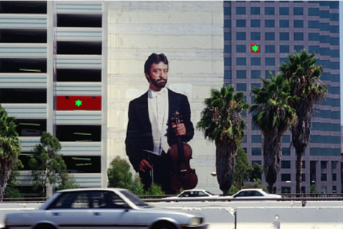} &
        \includegraphics[width=0.25\columnwidth]{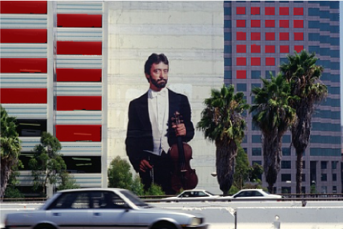} &
        \includegraphics[width=0.25\columnwidth]{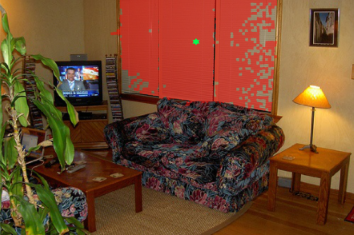} &
        \includegraphics[width=0.25\columnwidth]{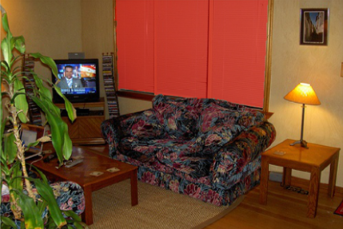} \\
    \end{tabular}
    }
    \centering
    \caption{Applications of our annotation tool for indoor meshes (top left), building models (top right), and images (bottom).}
    \label{fig:annot_apps}
\end{figure}

We introduced SUM Parts, a part-level semantic segmentation dataset for urban meshes covering \(2.5\,\text{km}^2\) with 21 classes. A novel annotation tool facilitates the semantic labeling of mesh faces and texture pixels with efficient 2D/3D selection strategies, streamlining the annotation of 3D urban scenes. Our evaluations show that our approach outperforms existing interactive annotation methods.\\
\noindent \textbf{Applications.} Our interactive annotation method can also handle complex indoor scenes, compact building models, and images (see~\cref{fig:annot_apps}). The SUM Parts dataset advances lightweight semantic city modeling by providing part-level semantics, enabling automated reconstruction of CityGML LoD3 city models~\cite{OGC-CityGML2,Groger12}. 

\noindent \textbf{Limitations.} Our annotation method relies on geometric precision and structural clarity and may be less effective for triangle meshes with topological errors, low resolution, or poor planar over-segmentation. It is also not applicable to natural scenes (e.g., mountains) or complex structures (e.g., palaces) that do not conform to planar and protrusion-based assumptions. In texture annotation, performance can degrade with complex textures, cluttered backgrounds, smaller superpixels (increasing processing time) or larger ones (risking under-segmentation), shadows, or regions with minimal color differentiation.

{
    \small
    \bibliographystyle{ieeenat_fullname}
    \bibliography{main}
}
\clearpage
\maketitlesupplementary
\pagenumbering{gobble}
\section{Details on annotation tool }




\subsection{Face-based annotation}
\paragraph{Protrusion score.}
By measuring the distance and angle from face \( f_i \) to support plane \( P^f_k \), we define protrusion score 
\(
    p_i = d_i + \omega_i \cdot \theta_i, 
\)
where
\[
    d_i = \max_{t \in \{0, 1, 2\}} \left( \text{dist}(v_{t,i}, P^f_k) \right)
\]
is the maximum Euclidean distance from the vertices \( v_t \) of the face \( f_i \) to the support plane \( P^f_k \). 
The angle weight \( \theta_i \) is calculated by measuring the angle 
\(
    \hat{\theta}_i = \cos^{-1} (\mathbf{n}_i \cdot \mathbf{n}_k)
\)
between the normal \( \mathbf{n}_i \) of face \( f_i \) and the normal \( \mathbf{n}_k \) of support planar segment \( P^f_k \), defined as:
\[
    \theta_i = \frac{\min(\hat{\theta}_i, 180^\circ - \hat{\theta}_i)}{90^\circ}. 
\]

\paragraph{Geometric consistency.}
To measure the geometric consistency between adjacent faces, we utilize an interior shrinking ball algorithm derived from the 3D medial axis transform to compute the ball radii for each face~\cite{sherbrooke1996algorithm,peters2018geographical}.
\setlength{\intextsep}{-8pt} 
\setlength{\columnsep}{5pt} 
\begin{wrapfigure}{r}{0.5\columnwidth} 
 \begin{center}
  \includegraphics[width=0.5\columnwidth]{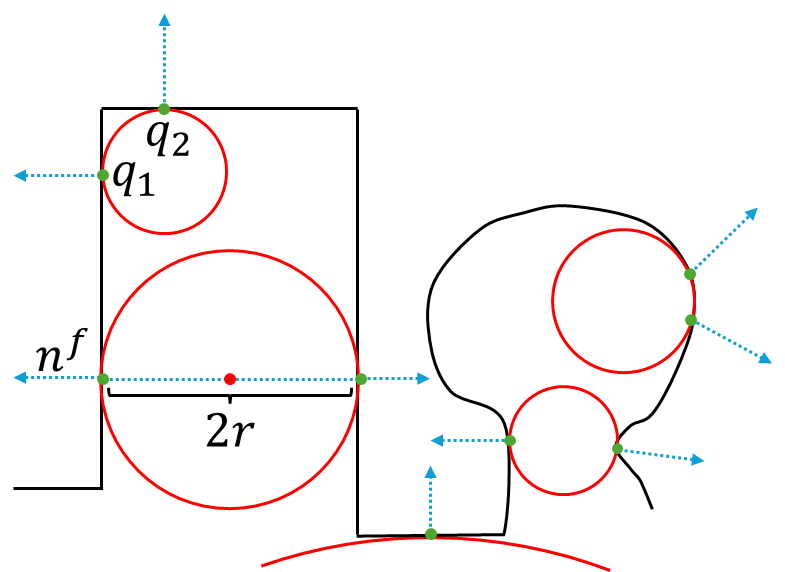} 
  \caption{Cross-sectional view of interior shrinking balls (red) in urban scenarios.}
  \label{fig:mat_ball}
  \end{center}
\end{wrapfigure}
\noindent In urban mesh scenarios, larger shrinking balls typically correspond to major geometric structures such as the terrain or main surfaces of buildings, whereas smaller balls indicate sharp structures or protrusions (as shown in~\cref{fig:mat_ball}).
Consequently, the size of these balls can indirectly reflect the local structural scale, suggesting that adjacent faces within the same geometric structure should have similar radii.
The mesh shrinking ball radius is derived as
\[
    r = \frac{\|q_1 - q_2\|^2}{2 (\mathbf{n}^f \cdot (q_1 - q_2))}, 
\]
where \(r\) refers to the radius \(r_i\) or \(r_j\), and \(\mathbf{n}^f\) denotes the normal \(\mathbf{n}_i\) or \(\mathbf{n}_j\) of the respective faces \(f_i\) or \(f_j\); \( q_1 \) and \( q_2 \) are the tangent points on the faces. 
\setlength{\intextsep}{\defaultintextsep}
\setlength{\columnsep}{\defaultcolumnsep}


\paragraph{Planar segment matching.} 
We define the feature vector \(\mathbf{F^{(seg)}}\) to quantify segment matching similarity, including:
\begin{itemize}
\item Geometric homogeneity: Differences in area between geometrically similar segments are calculated as:
\[
    \Delta A^{(seg)} = \frac{\left| \text{\textit{area}}^{(c)} -  \text{\textit{area}}^{(t)} \right|}{ \text{\textit{area}}^{(t)}}, 
\]
where \(  \text{\textit{area}}^{(c)} \) and \(  \text{\textit{area}}^{(t)} \) are the areas of the candidate and template segments, respectively.

\item Spatial distribution: Vertical distribution similarity is measured by comparing weighted average heights:
\[
    \Delta H^{(seg)} =\left| \frac{\sum_{i=1}^{m'} z_{i} \cdot a_{i}}{ \text{\textit{area}}^{(c)}} - \frac{\sum_{j=1}^{m''} z_{j} \cdot a_{j}}{ \text{\textit{area}}^{(t)}}\right|. 
\]
\noindent where \(z_i\) and \(a_i\) denote the z-coordinate and area of each face \(f_i\) in the candidate segment \(P^{(c)}_k\). \(z_j\) and \(a_j\) denote those in the template segment \(P^{(t)}\). 

\item Spatial orientation: Similarity in vertical orientation is assessed between segments \(P^{(c)}_k\) and \(P^{(t)}\)~\cite{ROUHANI2017124}.
\item Shape sphericity: Calculated using eigenvalues from triangle vertices of the segment \(P^{(c)}_k\) and \(P^{(t)}\) to evaluate similarity~\cite{weinmann2013feature}.
\item Photometric coherence: Color similarity is assessed using CIELAB~\cite{luo2001development} color distance and greenness~\cite{mckinnon2017comparing} difference.
\end{itemize}

\paragraph{Protrusion matching.} 
For seed expansion, in addition to spatial and segment scale constraints, we introduce optional topology constraints based on adjacency to optimize user focus and simplify inspection. In urban scenes, small protrusions (e.g., cars, balconies, dormers) reduce global matching efficiency by increasing inspection workload and computation (e.g., matching cars globally takes 5s, whereas planar facades take only 0.4s, see~\cref{fig:segmatch_b}). Therefore, we set topology constraints as the default for practical efficiency by confining the search space to planar segments of the template support surface (e.g., limiting annotations to the current facade for facade installations).

\noindent The feature vector \(\mathbf{F^{(str)}}\) includes:
\begin{itemize}
\item Spatial compactness: The compactness of a protrusion is quantified by considering its volume. We expect similar protrusions to have comparable values, defined as
\[
    \Delta V^{(str)} = \left| \frac{\text{\textit{vol}}^{(c)}}{\text{\textit{vol}}^{(c)}_{box}} - \frac{\text{\textit{vol}}^{(t)}}{\text{\textit{vol}}^{(t)}_{box}}\right|,
\]
where \(\text{\textit{vol}}^{(c)}\) and \(\text{\textit{vol}}^{(c)}_{box}\) represent the volume of \(f^{(c)}\) and its bounding box volume, respectively. \(\text{\textit{vol}}^{(t)}\) and \(\text{\textit{vol}}^{(t)}_{box}\) are the corresponding values for \(f^{(t)}\).
\item Surface complexity: We assume complex 3D shapes decompose into multiple planar segments. Surface complexity similarity is measured by the ratio of the number of planar segments in the template and candidate protrusions, defined as
\[
    \Delta N^{(str)} = \left(\frac{\max(n^{(t)}, n^{(c)})}{\min(n^{(t)}, n^{(c)})}\right)^\mu, 
\]
where \(n^{(t)}\) and \(n^{(c)}\) respectively represent the number of planar segments for the template and candidate protrusions, and \(\mu = {\min(n^{(t)}, n^{(c)})}\).
\item Structural features: Measuring the similarity of protrusions involves comparing their structural features through eigenvalue analysis including linearity, planarity, and sphericity~\cite{weinmann2013feature}. We determine similarity by the \( \ell_1 \) distance in the feature space, including differences in linearity \(\Delta L^{(str)}\), planarity \(\Delta P^{(str)}\), and sphericity \(\Delta S^{(str)}\).
\end{itemize}

\subsection{Texture-based annotation}
\paragraph{Gaussian mixture model (GMM).} 
\(G_k\) denotes the GMM for the \(k\)-th channel, defined as 
\[
    G_k(S) = \sum_{m=1}^{M} \pi_{km} \mathcal{N}(x; \mu_{km}, \Sigma_{km}),
\]
where \(S\) represents the superpixel \(S_0\) or \(S_j\). \(x\) is a pixel sample point of \(S\), and \(M\) is the number of components in GMM (\(M\) set to 5 in all experiments in this paper).
\( \mathcal{N}(s; \mu, \Sigma) \) denotes the multivariate normal distribution, with \(\mu\) representing the mean for superpixels \(S_0\) or \(S_j\), and \(\Sigma\) denotes their respective covariance matrices.

\paragraph{Local color consistency.} 
For local color consistency, where 
\(
    \rho_{j} = \Delta E_{00}(U_0, U_j)  
\)
is the color distance (i.e., CIEDE2000~\cite{luo2001development}) from the superpixel \( S_j \) to its seed \( S_0 \).
To more accurately capture the intrinsic structure and variability within superpixels' color distributions, we employ a GMM to compute the average Lab color, represented by \( U = \sum_{m=1}^M \pi_{m} \mu_{m}\), where \( U \) represents \( U_0 \) or \( U_j \), with \( \pi_{m} \) as the mixing weight and \( \mu_{m} \) as the mean for the \( m \)-th Gaussian component in the Lab color space.
Additionally, seed samples for \( U_0 \) are taken from its first-order neighborhood, whereas samples for \( U_j \) come from its own pixels.

\paragraph{Region-based template matching.} 
The feature vector \(\mathbf{F^{(reg)}}\) includes:
\begin{itemize}
    \item Shape index: To assess shape similarity, we use a shape index reflecting elongation or flatness, which is defined as:
    \(
        r = \frac{\min(w, h)}{\max(w, h)},
    \)
    where \(w\) and \(h\) represent the width and height of the object's bounding box, respectively. 
    The similarity between regions is calculated as:
    \begin{math}
        \Delta I^{(reg)} = |r_c - r_t|,
    \end{math}
    where \(r\) represents the ratio \(r_c\) of the candidate region or \(r_t\) of the template region.
    \item Shape regularity: We assess shape regularity to calculate the similarity between areas. Similar to structural matching, compactness is used to describe how well a shape fills its bounding box, defined as:
    \[
        \Delta A^{(reg)} = \left| \frac{\text{\textit{area}}^{(c)}}{\text{\textit{area}}^{(c)}_{box}} - \frac{\text{\textit{area}}^{(t)}}{\text{\textit{area}}^{(t)}_{box}}\right|,
    \]
    where \(\text{\textit{area}}^{(c)}\) and \(\text{\textit{area}}^{(t)}\) represent the area of the candidate and template regions, respectively, and \(\text{\textit{area}}^{(c)}_{box}\) and \(\text{\textit{area}}^{(t)}_{box}\) are the areas of their bounding boxes. 
    \item Contextual features: Similar regions should have similar internal and external color distributions. We evaluate these differences using the Wasserstein distance, calculated as:
    \(
        \Delta D^{(reg)} = \left| W(G_k(R^{(c)}_{in}),G_k(R^{(c)}_{out})) - W(G_k(R^{(t)}_{in}),G_k(R^{(t)}_{out})) \right|, 
    \)
    where \(R^{(c)}_{in}\) and \(R^{(c)}_{out}\) denote the interior and exterior pixel collections of the candidate region, respectively, and \(R^{(t)}_{in}\) and \(R^{(t)}_{out}\) for the template region. The external region \(R_{out}\) includes pixels covered but not selected during local expansion. 
\end{itemize}

\paragraph{Scalability.} Our workflow is fully compatible with deep learning-based frameworks like Semantic-SAM~\cite{semantic_sam_2024} and Mask DINO~\cite{maskdino_2023}. Combining them demonstrates the potential to accelerate template generation through prompt-based segmentation at various granularities and refine template matching with instance/object detection.
Additionally, our 2D paint canvas (see~\cref{fig:2dsel}) converts texture segments into images that are compatible with these segmentation methods. This, combined with our annotated dataset, allows direct training on 3D textured surfaces, setting the stage for future improvements in efficiency and accuracy.

\section{Details on benchmark results}
\subsection{Evaluation of interactive annotation}
\paragraph{Evaluation metrics.} 
Traditional metrics, such as click counts to achieve specific Intersection over Union (IoU) or Average Precision (AP), are quantifiable but do not fully capture the true efficiency of the annotation process. The main shortcomings of these methods include:
\begin{itemize}
    \item Evaluation limitations: Relying solely on IoU or AP does not fully capture annotation comprehensiveness. For example, a 90\% IoU may still require multiple boundary adjustments for accuracy.
    \item Interaction limitations: Click-based interactions alone cannot perfectly annotate boundaries, often requiring tools like lassos or polygons. Additionally, standardized click positions do not account for individual user variations, hindering realistic efficiency assessment.
    \item Efficiency limitations: Average click counts do not reflect actual interaction efficiency due to varying user speeds. Measuring total annotation time provides a more accurate assessment of efficiency.
\end{itemize}

To address these issues, we developed an evaluation system comprising Intersection over Union (IoU), Boundary IoU (BIoU), number of operations (Oper), annotation time (Time), and smart interaction ratio (SR). BIoU assesses boundary annotation accuracy~\cite{cheng2021boundary, weixiao2023pssnet}. Oper counts mouse clicks and keyboard keystrokes. Time measures annotation duration in seconds. SR quantifies the frequency of non-manual interactions (counting only click-based selections, excluding other operations). Our evaluation is based on user studies with \(u\) users across \(n\) test scenes, each with \(c\) categories. The average metrics are calculated as follows:
\begin{itemize}
    \item Evaluating a single scenario. For a given scenario \( s \), annotated by \( u \) users across \( c \) categories, \(\overline{mIoU}\), \(\overline{mBIoU}\), \(\overline{mOper}\), \(\overline{mTime}\), and \(\overline{mSR}\), can be obtained by averaging the \(IoU\), \(mBIoU\), \(mOper\), \(mTime\), and \(mSR\) values across all users and categories.
    \item Evaluating multiple scenarios. To obtain \(\overline{M}\), \(\overline{B}\), \(\overline{O}\), \(\overline{T}\), and \(\overline{S}\), the averages of for multiple scenarios, we take the average of each scenario’s \(\overline{mIoU}\), \(\overline{mBIoU}\), \(\overline{mOper}\), \(\overline{mTime}\), \(\overline{mSR}\) and then average these values: 
    \begin{align}
        \overline{M} &= \frac{1}{n} \sum_{s=1}^n \overline{mIoU}_s &  
        \overline{B} &= \frac{1}{n} \sum_{s=1}^n \overline{mBIoU}_s \notag \\ 
        \overline{O} &= \frac{1}{n} \sum_{s=1}^n \overline{mOper}_s &   
        \overline{T} &= \frac{1}{n} \sum_{s=1}^n \overline{mTime}_s \notag \\
        \overline{S} &= \frac{1}{n} \sum_{s=1}^n \overline{mSR}_s \notag
    \end{align}
\end{itemize}
In the user study, we recorded each user's annotation progress and interactions in real-time, requiring at least 95\% scene completion based on mesh area or texture pixels.

\begin{figure*}[!ht]
    \begin{tabular}{cccc}
        \includegraphics[width=0.23\textwidth]{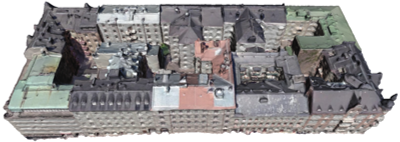} &
        \includegraphics[width=0.23\textwidth]{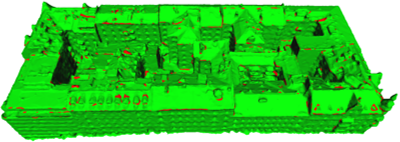} &
        \includegraphics[width=0.23\textwidth]{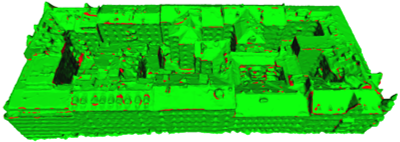} &
        \includegraphics[width=0.23\textwidth]{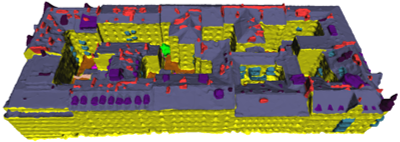} \\
       Input & Segment-based (errors)~\cite{gao2021sum} & Ours (errors) & Manual~\cite{rouhani2017semantic} \\
    \end{tabular}
    \centering
    \caption{Qualitative analysis of interactive mesh face annotations and their error maps (shown in red) for the courtyard complex.}
    \label{fig:annot_tri_errs_1}
\end{figure*}
\begin{figure*}[!ht]
    \begin{tabular}{cccc}
        \includegraphics[width=0.23\textwidth]{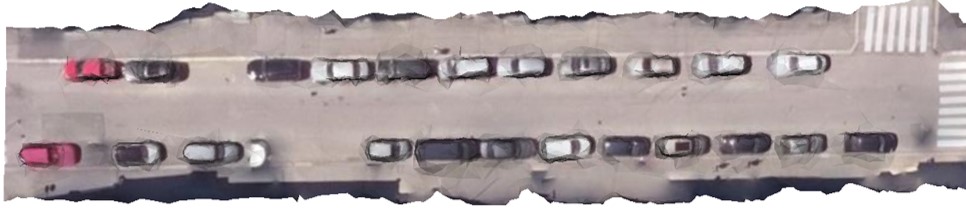} &
        \includegraphics[width=0.23\textwidth]{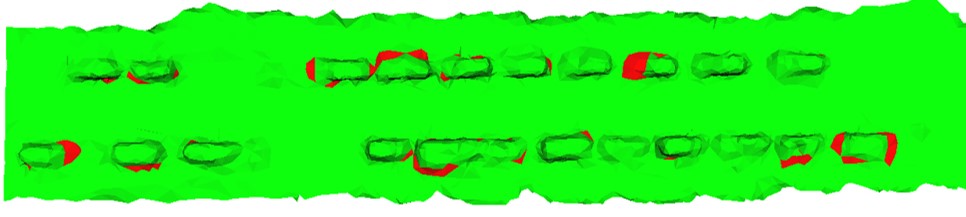} &
        \includegraphics[width=0.23\textwidth]{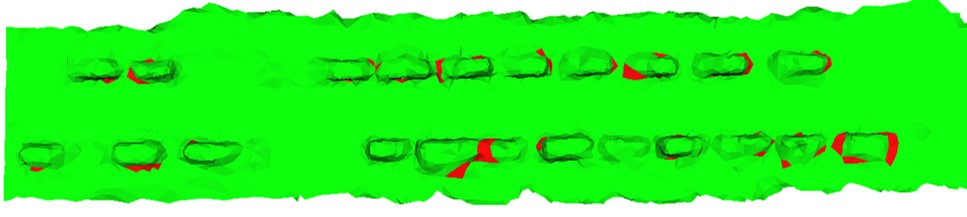} &
        \includegraphics[width=0.23\textwidth]{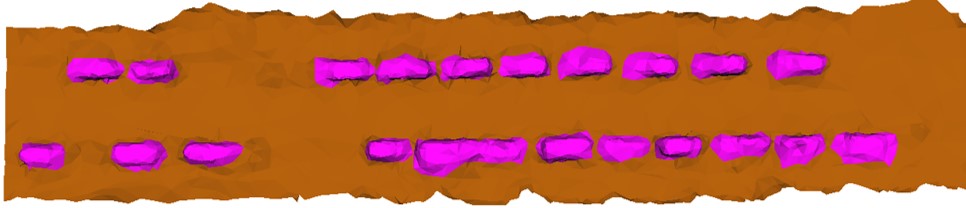} \\
         Input & Segment-based (errors)~\cite{gao2021sum} & Ours (errors) & Manual~\cite{rouhani2017semantic} \\
    \end{tabular}
    \centering
    \caption{Qualitative analysis of interactive mesh face annotations and their error maps (shown in red) for the street with vehicles.}
    \label{fig:annot_tri_errs_2}
\end{figure*}

\paragraph{Comparisons.} 
~\cref{tab:tri_all} shows that our method outperforms segment-based annotation~\cite{gao2021sum} in object region and boundary quality. Across all four test scenarios, it significantly reduces both interaction counts and annotation time. 
We also provide additional qualitative analysis, as shown in~\cref{fig:annot_tri_errs_1} and~\cref{fig:annot_tri_errs_2}.
\begin{table}[ht!]
\centering
\begin{tabular}{@{}lccccc@{}}
\toprule
    Metric & Method & Cour. & Stre. & Park. & Harb. \\ \midrule
    \multirow{2}{*}{\(\overline{mIoU}_s(\%)\)} 
    & Seg & 89.1 & 94.0 & 92.1 & 91.1 \\
    & Ours & \textbf{89.5} & \textbf{94.2} & \textbf{92.9} & \textbf{92.2} \\
    \midrule
    \multirow{2}{*}{\(\overline{mBIoU}_s(\%)\)} 
    & Seg. & \textbf{72.7}& \textbf{85.4} & 70.6 & 70.3 \\
    & Ours. & 72.4 & 84.5 & \textbf{71.9} & \textbf{71.0} \\
    \midrule
    \multirow{3}{*}{\(\overline{mOper}_s\)} 
    & Man. & 18154 & 1589 & 3559 & 3714 \\
    & Seg. & 17645 & 1407 & 2529 & 2894 \\
    & Ours & \textbf{13231} & \textbf{909} & \textbf{1797} & \textbf{1957} \\
    \midrule
    \multirow{3}{*}{\(\overline{mTime}_s(s)\)} 
    & Man. & 11401.0 & 969.5 & 2146.5 & 2215.8 \\
    & Seg. & 10441.3 & 757.0 & 1441.4 & 1623.7 \\
    & Ours & \textbf{9107.2} & \textbf{498.0} & \textbf{1105.9} & \textbf{1257.5} \\
    \midrule
    \(\overline{mSR}_s(\%)\) & Ours & \textbf{66.5} & \textbf{94.9 }& \textbf{85.8} & \textbf{84.8} \\
\bottomrule
\end{tabular}
\caption{Performance evaluation of interactive mesh face annotation methods across four scenarios: Cour. (courtyard complex), Stre. (streets with vehicles), Park. (park with trees), Harb. (harbor with ships). Methods include Man.~\cite{rouhani2017semantic} and Seg.~\cite{gao2021sum}. Highest values are shown in bold.
}
\label{tab:tri_all}
\end{table}

From~\cref{tab:tex_all}, our method has slightly lower \(\overline{M}\) than GrabCut~\cite{rother2004grabcut}, but achieves higher \(\overline{mIoU}_s\) in most scenarios and excels in boundary quality.
While SimpleClick~\cite{liu2023simpleclick} is more efficient in interaction count, our method outperforms others in most scenarios. Though slower than SAM~\cite{Kirillov_2023_ICCV}, our method still surpasses other methods in interaction time. 
The need for manual corrections enhances annotation quality without significant time cost, and our approach delivers more accurate boundaries with similar correction workloads compared to deep learning methods. 
We achieve this through:
(1) Better quality from the user-defined template that enables pixel-level boundary control, outperforming deep learning-based clicks by approximately +3.5$\sim$6.5\% mIoU and +18.4$\sim$19.5\% boundary mIoU (~\cref{tab:annot_eval}), especially for regular shapes like windows in~\cref{fig:bound_errs}.
(2) Higher efficiency offered by reusable, scale- and rotation-invariant templates, which reduces the interaction count by -18.7\% compared to SAM (582 vs. SAM's 716) and annotation time by -23\% compared to SimpleClick (663.3s vs. SimpleClick's 861.3s), benefiting repetitive structures (~\cref{tab:annot_eval}).
Although SAM is slightly faster and SimpleClick requires fewer interactions, our intentional design using handcrafted templates instead of intensive smart clicks prioritizes higher-quality annotations while maintaining a similar total annotation time.
\begin{table}[ht!]
    \centering
      \resizebox{\columnwidth}{!}{
    \begin{tabular}{@{}lccccccc@{}}
    \toprule
    Metric & Method & Fac1. & Fac2. & Par1. & Par2. & Rod1. & Rod2. \\ 
    \midrule
    \multirow{4}{*}{\(\overline{mIoU}_s(\%)\)} 
        & Gra. & 80.2 & 87.3 & 94.0 & 90.4 & 91.3\(^{\ast}\) & \textbf{85.4} \\
        & SAM & \textbf{81.0} & 86.4 & 85.3 & 86.4 & 86.5 & 80.7 \\
        & Sip. & 73.7 & 77.5 & 87.9 & 86.2 & 84.0 & 79.1 \\
        & Ours & 79.2 & \textbf{88.7} & \textbf{95.0} & \textbf{91.2}& \textbf{91.3} & 82.1 \\
        \midrule
        \multirow{4}{*}{\(\overline{mBIoU}_s(\%)\)} 
        & Gra. & 27.8 & 45.4 & 63.7 & 45.8 & 49.9 & 50.1 \\
        & SAM & 24.7 & 33.9 & 25.8 & 23.3 & 34.1 & 37.0 \\
        & Sip. & 19.7 & 26.8 & 38.0 & 34.4 & 31.8 & 34.6 \\
        & Ours & \textbf{28.1} & \textbf{50.5} &\textbf{67.8} &\textbf{48.6}& \textbf{50.5} & \textbf{50.6}\\
        \midrule
        \multirow{5}{*}{\(\overline{mOper}_s\)} 
        & Man. & 515 & 124 & 497 & 297 & 718 & 1764 \\
        & Gra. & 717 & 156 & 462 & 243 & 960 & 1729 \\
        & SAM & 715 & 319 & 363 & 319 & 1020 & 1684 \\
        & Sip. & \textbf{297} & 119 & \textbf{77} & \textbf{78} & \textbf{400} & \textbf{539} \\
        & Ours & 487 & \textbf{105} & 497 & 213 & 720 & 1468 \\
        \midrule
        \multirow{5}{*}{\(\overline{mTime}_s(s)\)} 
        & Man. & 816.9 & 185.6 & 636.7 & 280.1 & 801.7 & 1941.4 \\
        & Gra. & 920.1 & 242.6 & 494.9 & 270.8 & 836.3 & 1920.6 \\
        & SAM & \textbf{565.5} & \textbf{150.6} & 460.9 & 389.5 & 800.6 & \textbf{1476.5} \\
        & Sip. & 1128.5 & 338.9 &\textbf{197.7} & 230.8 & 1526.4 & 1745.5 \\
        & Ours & 631.1 & 155.8 & 565.9 & \textbf{189.1} & \textbf{767.9} & 1670.3 \\
        \midrule
        \multirow{4}{*}{\(\overline{mSR}_s(\%)\)} 
        & Gra. & 7.1 & 11.8 & 59.3 & 66.5 & 9.9 & 26.7 \\
        & SAM &\textbf{94.7} & \textbf{92.8} & \textbf{78.8} & \textbf{78.6} & \textbf{49.2} & 36.0 \\
        & Ours & 20.3 & 21.2 & 51.6 & 75.8 & 32.2 & \textbf{40.8}\\
    \bottomrule
    \end{tabular}
    }
    \caption{Performance evaluation of interactive texture annotation methods across six scenarios: Man. (manual), Gra. (GrabCut~\cite{rother2004grabcut}), SAM (Segment Anything~\cite{Kirillov_2023_ICCV}), Sip. (SimpleClick~\cite{liu2023simpleclick}), Fac1./Fac2. (facades 1 \& 2), Par1./Par2. (parks 1 \& 2), Rod1./Rod2. (roads 1 \& 2). \(^{\ast}\)GrabCut on Rod1. achieved 91.29\%, slightly below our method's 91.31\%. Highest values are in bold.}
    \label{tab:tex_all}
\end{table}

\begin{figure*}[!ht]
      \resizebox{\textwidth}{!}{
    \begin{tabular}{cccccc}
        \includegraphics[width=0.16\textwidth]{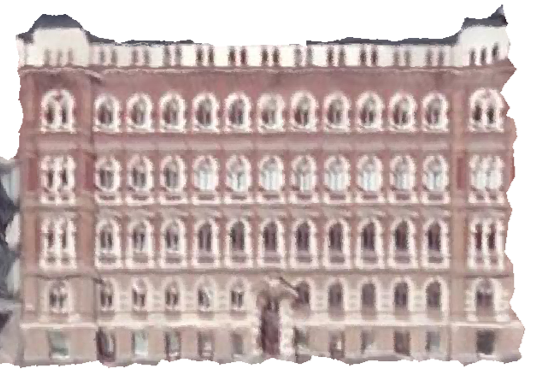} &
        \includegraphics[width=0.16\textwidth]{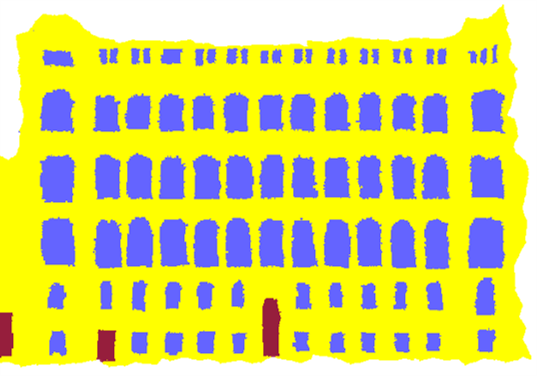} &
        \includegraphics[width=0.16\textwidth]{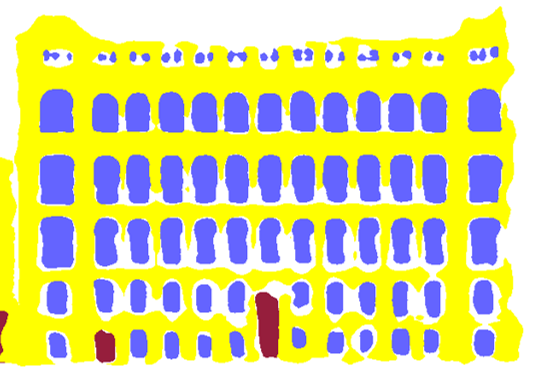} &
        \includegraphics[width=0.16\textwidth]{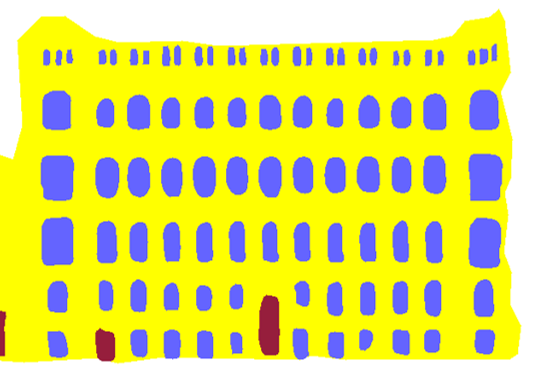} &
        \includegraphics[width=0.16\textwidth]{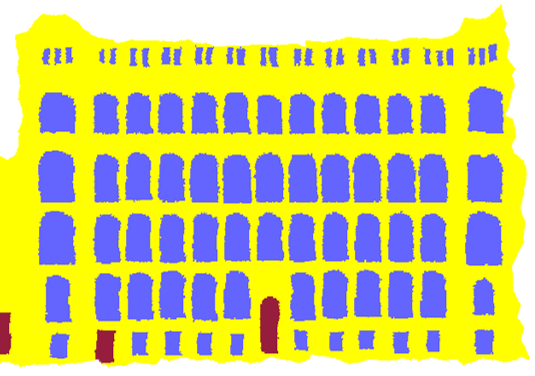} &
        \includegraphics[width=0.16\textwidth]{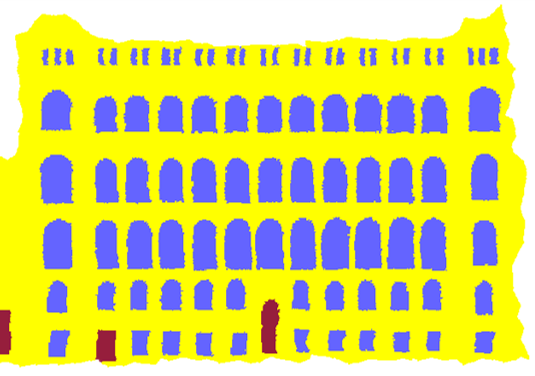} \\
        Input & GrabCut~\cite{rother2004grabcut} & Simclick~\cite{liu2023simpleclick} & SAM~\cite{Kirillov_2023_ICCV} & Ours & Manual \\
    \end{tabular}
    }
    \centering
    \caption{Qualitative analysis of interactive texture annotation results for the facade.}
    \label{fig:tex_anno_results_1}
\end{figure*}

\begin{figure*}[!ht]
      \resizebox{\textwidth}{!}{
    \begin{tabular}{cccccc}
        \includegraphics[width=0.16\textwidth]{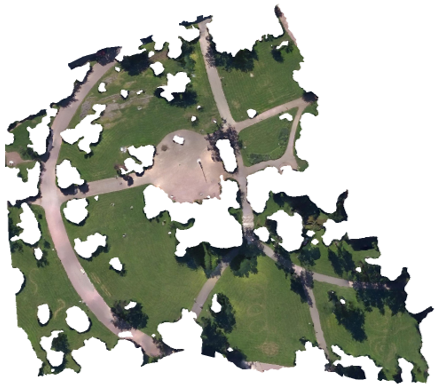} &
        \includegraphics[width=0.16\textwidth]{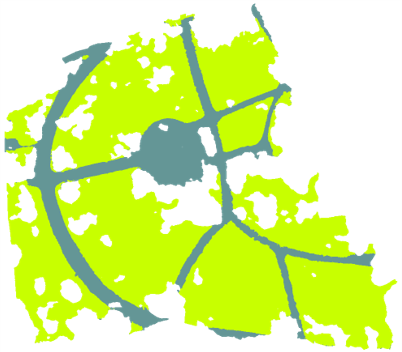} &
        \includegraphics[width=0.16\textwidth]{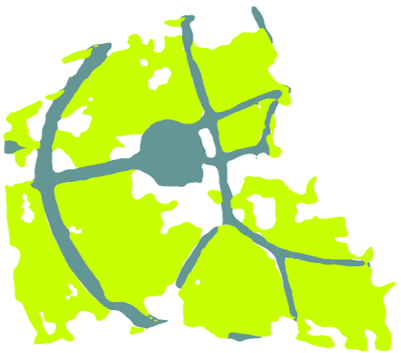} &
        \includegraphics[width=0.16\textwidth]{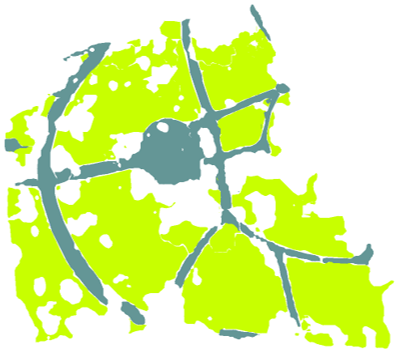} &
        \includegraphics[width=0.16\textwidth]{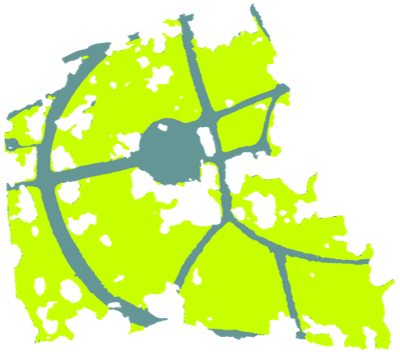} &
        \includegraphics[width=0.16\textwidth]{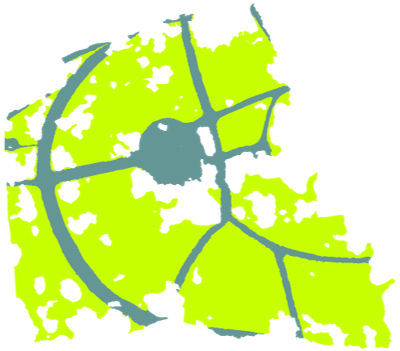} \\
		Input & GrabCut~\cite{rother2004grabcut} & Simclick~\cite{liu2023simpleclick} & SAM~\cite{Kirillov_2023_ICCV} & Ours & Manual \\
    \end{tabular}
    }
    \centering
    \caption{Qualitative analysis of interactive texture annotation results for the park.}
    \label{fig:tex_anno_results_2}
\end{figure*}

\begin{figure*}[!ht]
    \resizebox{\textwidth}{!}{
    \begin{tabular}{ccc}
        \includegraphics[width=0.32\textwidth]{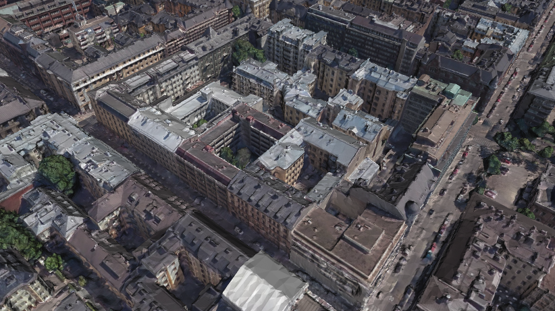}&
        \includegraphics[width=0.32\textwidth]{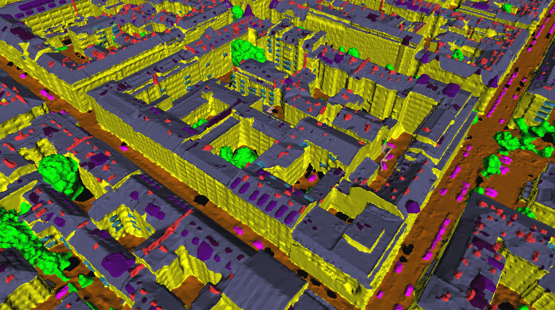}&
        \includegraphics[width=0.32\textwidth]{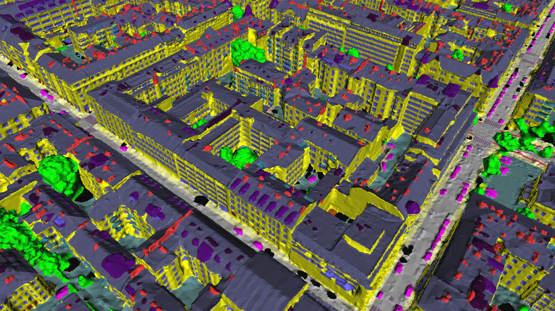}\\
        \includegraphics[width=0.32\textwidth]{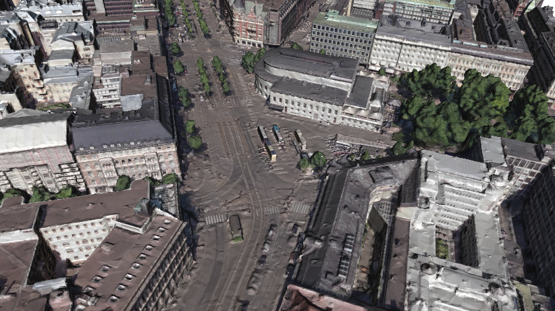}&
        \includegraphics[width=0.32\textwidth]{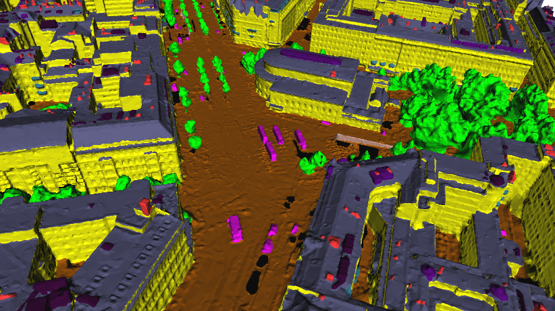}&
        \includegraphics[width=0.32\textwidth]{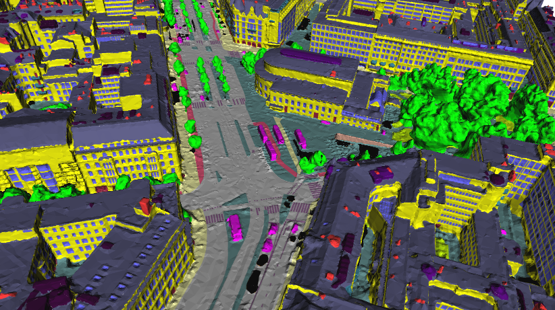}\\
        \includegraphics[width=0.32\textwidth]{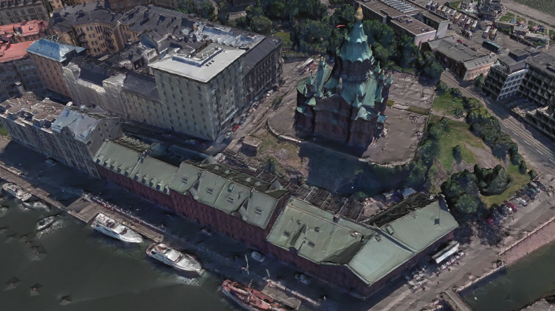}&
        \includegraphics[width=0.32\textwidth]{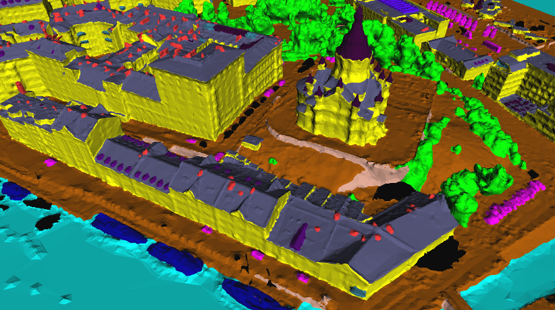}&
        \includegraphics[width=0.32\textwidth]{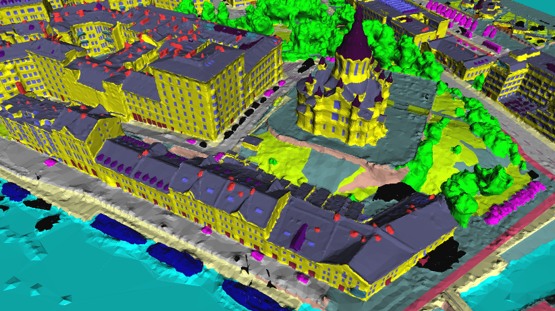}\\
        \includegraphics[width=0.32\textwidth]{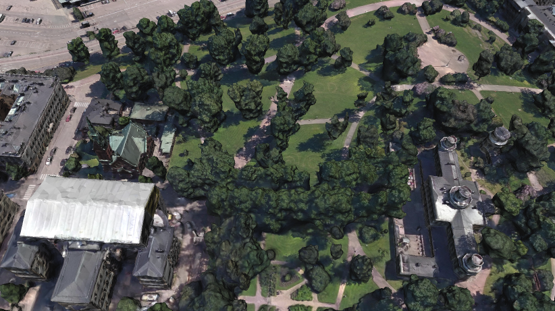}&
        \includegraphics[width=0.32\textwidth]{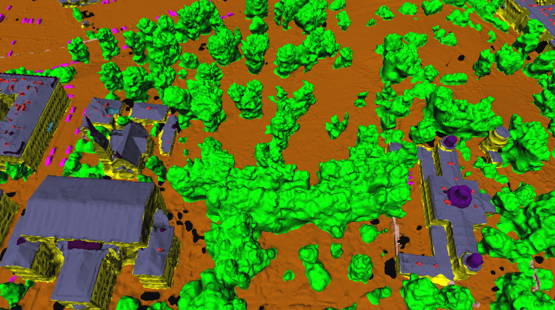}&
        \includegraphics[width=0.32\textwidth]{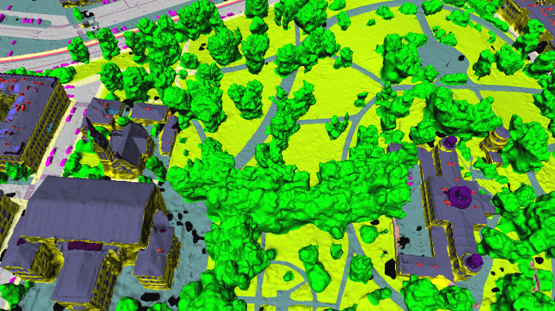}\\
        \includegraphics[width=0.32\textwidth]{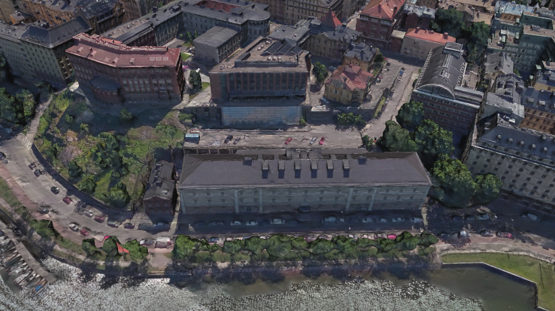}&
        \includegraphics[width=0.32\textwidth]{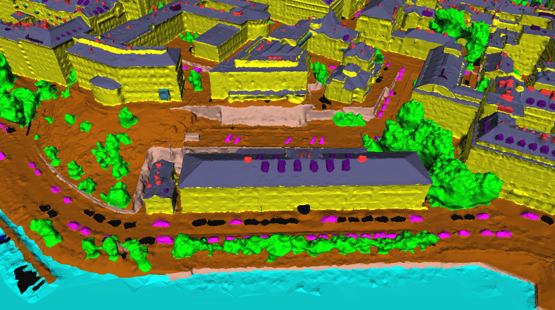}&
        \includegraphics[width=0.32\textwidth]{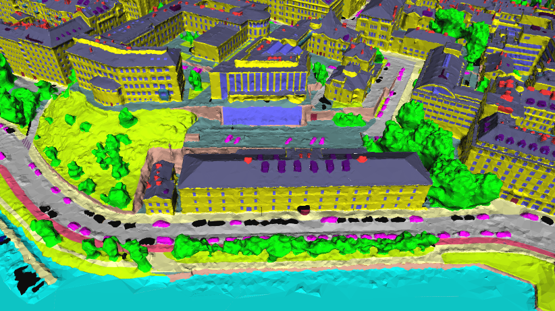}\\
         Input &
         Face-based annotation&
         Texture-based annotation\\
    \end{tabular}
    }
    \centering 
    \caption{Examples of part-level annotated semantic urban meshes are displayed from the first to the third column, showing textured meshes, face-based semantic meshes (13 classes), and texture-based semantic meshes (19 classes), respectively.
  }
\label{fig:annotation_results}
\end{figure*}

For regular-shaped objects, interactive clicking is suboptimal. As shown in~\cref{fig:tex_anno_results_1} and~\cref{fig:tex_anno_results_2}, single clicks lack boundary precision, and multiple clicks do not significantly improve accuracy. Repetitive structures increase the annotation burden due to frequent clicking. Instead, users achieve high precision by drawing rectangles or polygons for elements like windows or doors. Our method enables efficient annotation of similar structures by creating a graphical template once. In summary, if manual corrections in semi-automatic annotations take as much or more time than fully manual annotations, the method loses its utility. 
Additional qualitative results from our annotation methods are presented in~\cref{fig:annotation_results}.

\paragraph{Ablation studies on template matching.} 
Our feature design is grounded in geometric priors (shape properties and structural distribution), label-free operation, and computational efficiency, validated through hierarchical ablation studies (mIoU) as follows.
For matching: (1) Planar segments (e.g., roofs, best 88.4\% in~\cref{fig:annot_tri_errs_1}). Removing geometric homogeneity (-6.1\%),  spatial distribution (-13.8\%), orientation (-13.3\%), and shape sphericity (-1.8\%) caused performance drops. (2) Protrusions (e.g., cars, best 97.0\% in~\cref{fig:annot_tri_errs_2}). When spatial compactness (-2.6\%), surface complexity (-1.2\%), and structural features (-1.6\%) were removed, precise matching suffered significantly. (3) Regions (e.g., windows, best 90.7\% in~\cref{fig:tex_anno_results_1}). Eliminating shape index (-5.2\%), regularity (-38.1\%), and contextual features (-20.3\%) severely impaired boundary alignment and color consistency.
These results highlight the essential role of each feature and their combined effectiveness, confirming our method's superior performance.

\subsection{Evaluation of semantic segmentation}
\begin{figure*}[!ht]
    \centering
    \begin{tabular}{cccc}
        \includegraphics[width=0.23\textwidth]{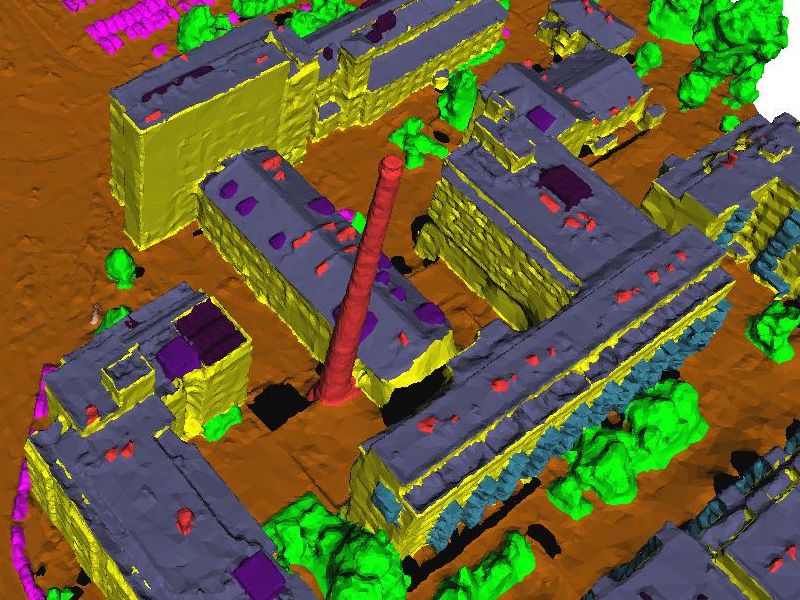} &
        \includegraphics[width=0.23\textwidth]{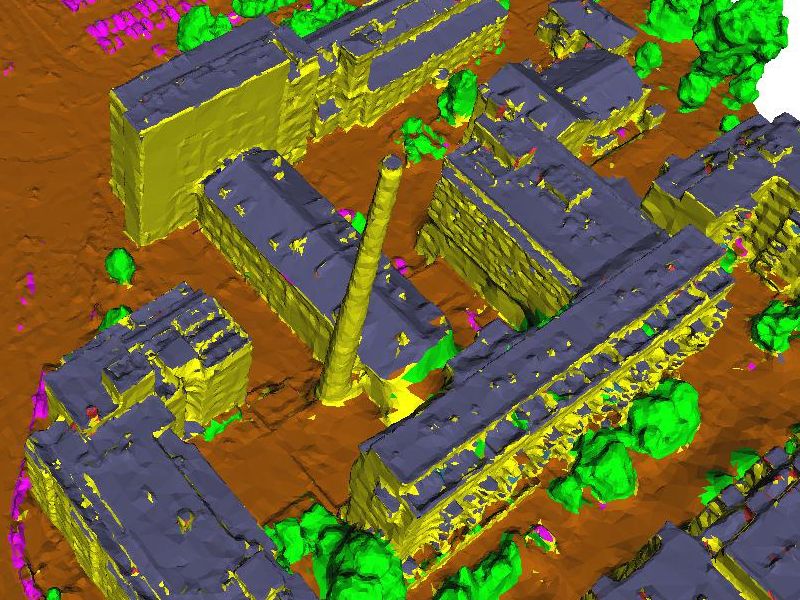} &
        \includegraphics[width=0.23\textwidth]{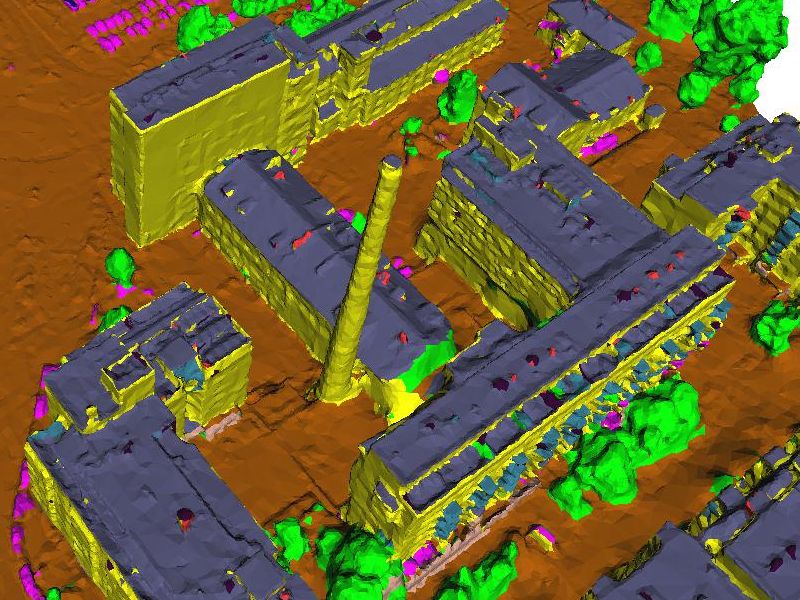} &
        \includegraphics[width=0.23\textwidth]{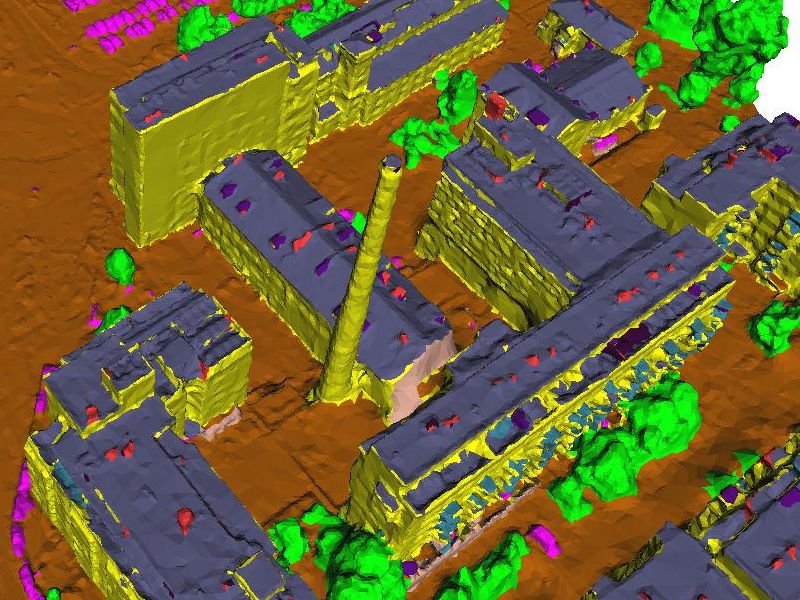} \\	
        \includegraphics[width=0.23\textwidth]{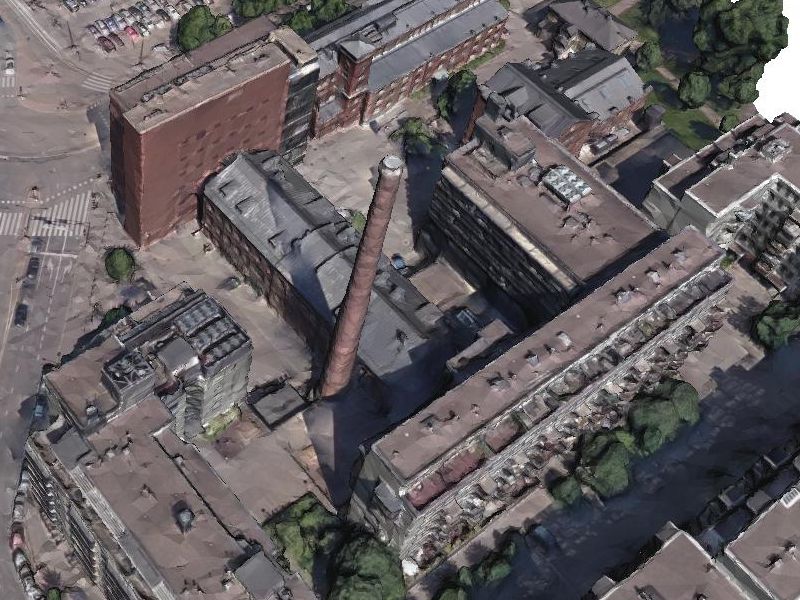} &
        \includegraphics[width=0.23\textwidth]{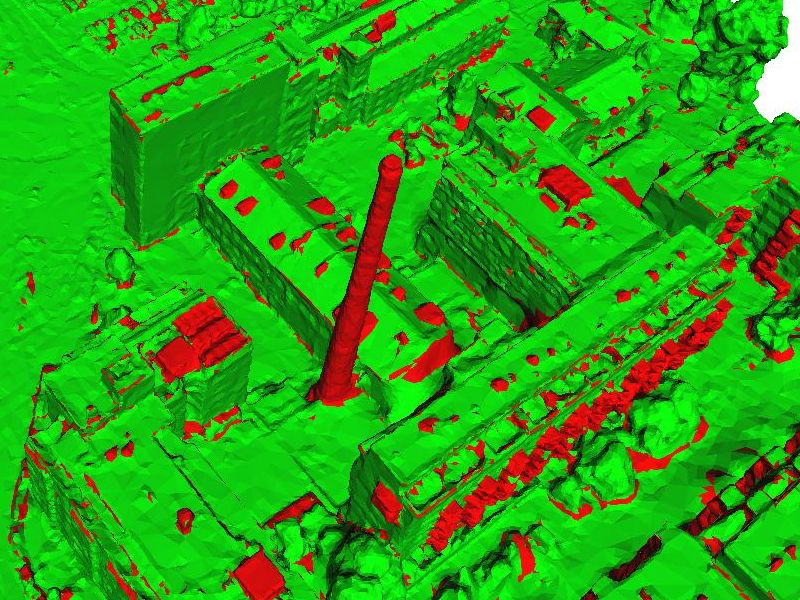} &
        \includegraphics[width=0.23\textwidth]{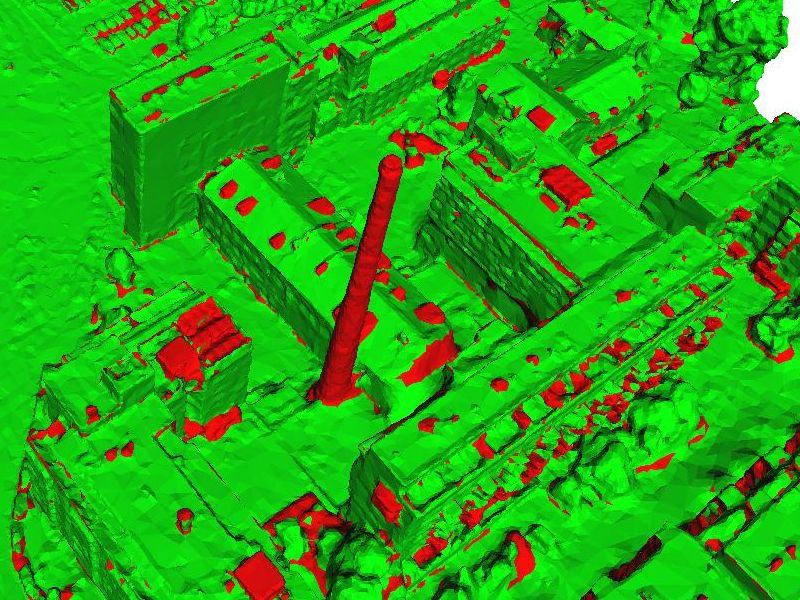} &
        \includegraphics[width=0.23\textwidth]{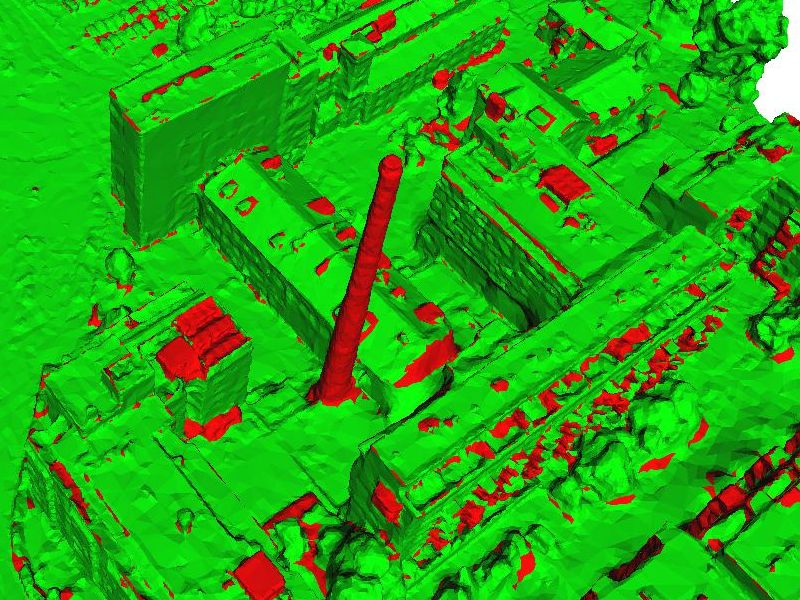} \\
        Truth (top), Mesh (bottom)   &
		RF\_MRF &
		SUM\_RF &
        PSSNet \\
        \includegraphics[width=0.23\textwidth]{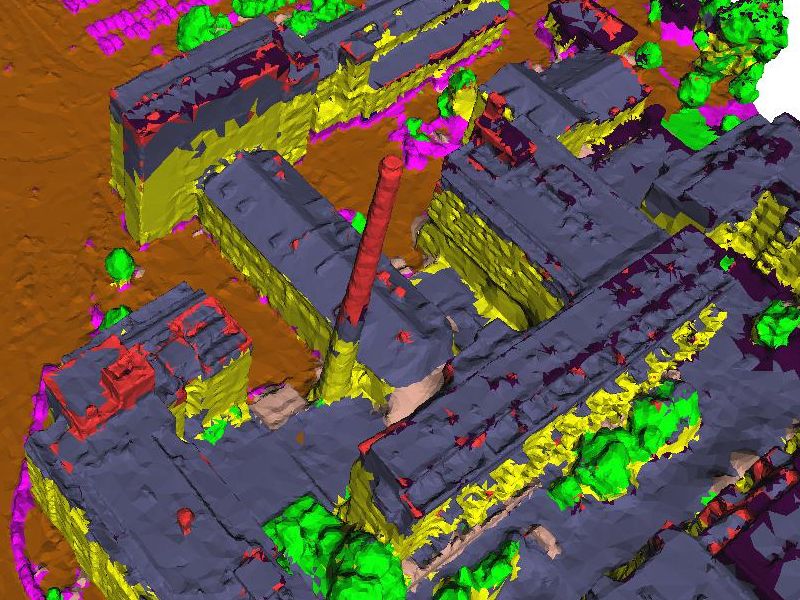} &
        \includegraphics[width=0.23\textwidth]{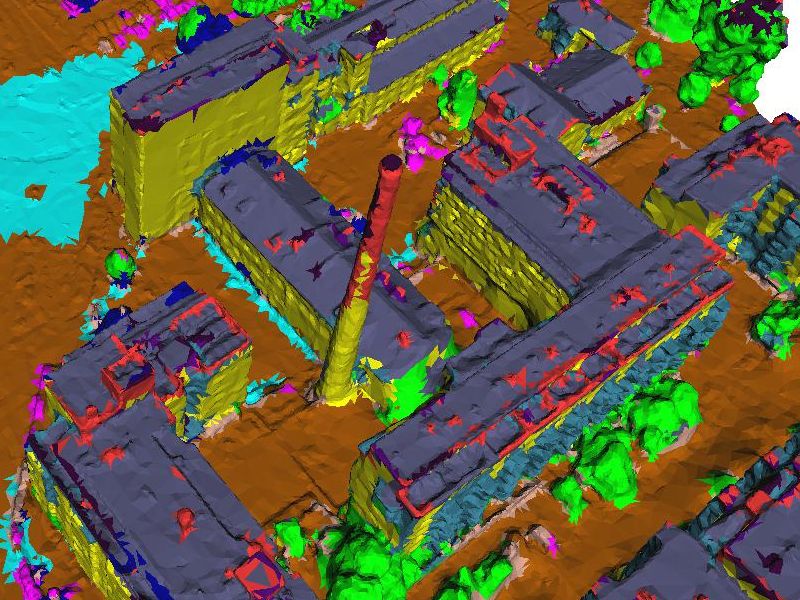} &
        \includegraphics[width=0.23\textwidth]{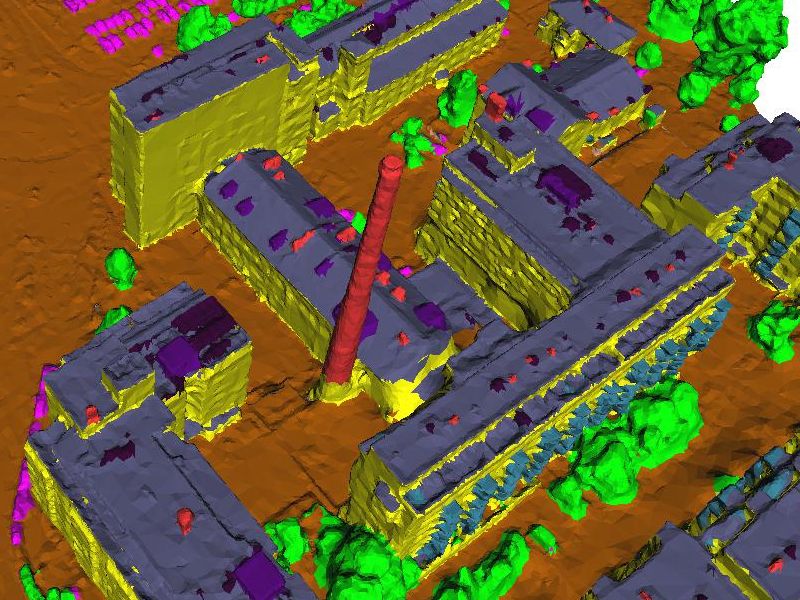} &
        \includegraphics[width=0.23\textwidth]{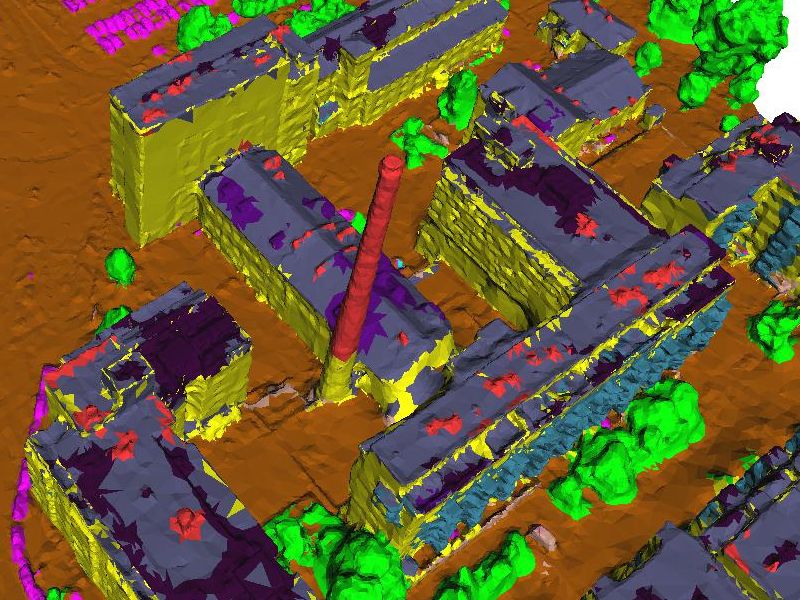} \\
        \includegraphics[width=0.23\textwidth]{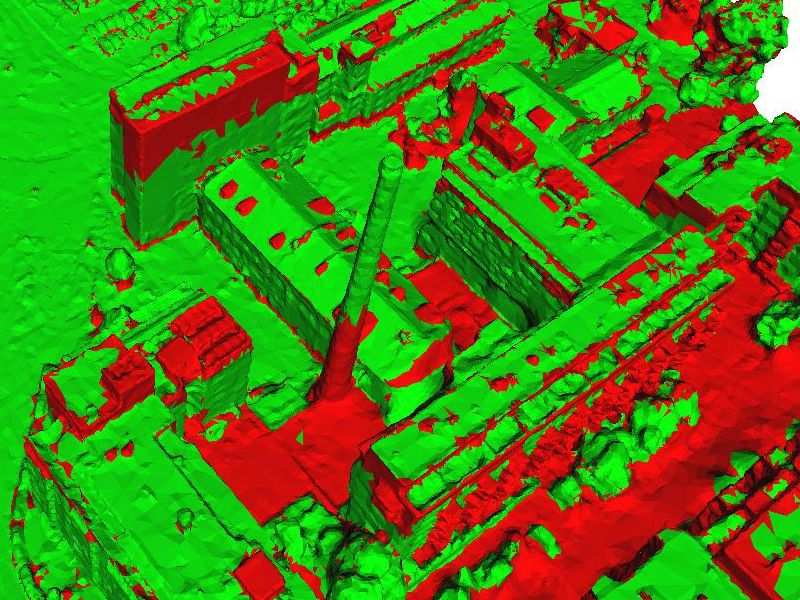} &
        \includegraphics[width=0.23\textwidth]{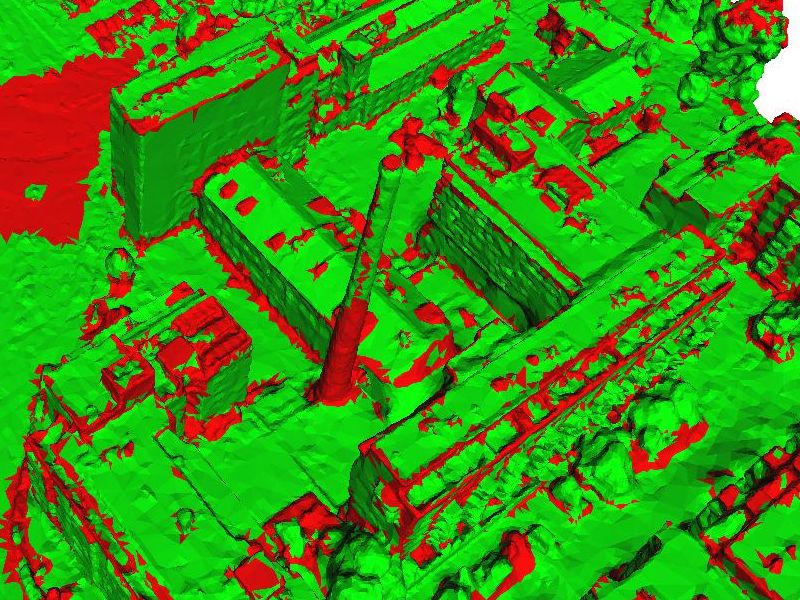} &
        \includegraphics[width=0.23\textwidth]{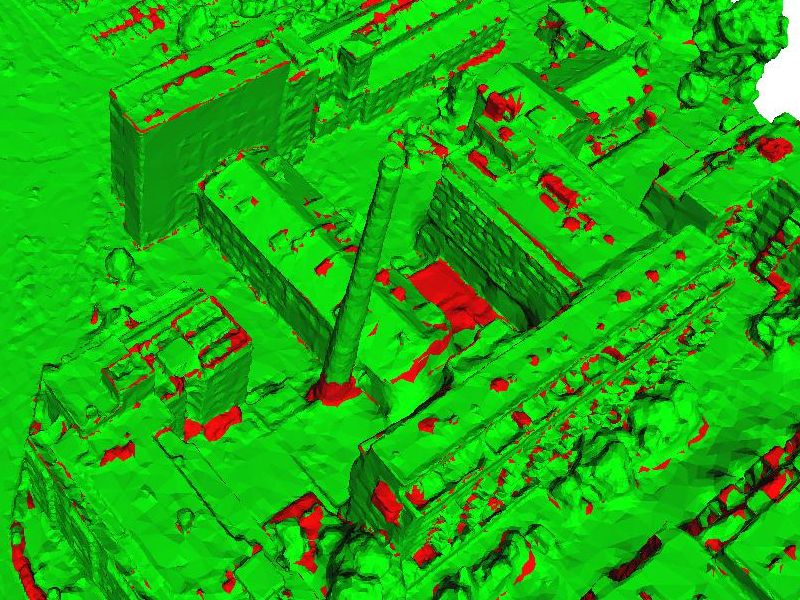} &
        \includegraphics[width=0.23\textwidth]{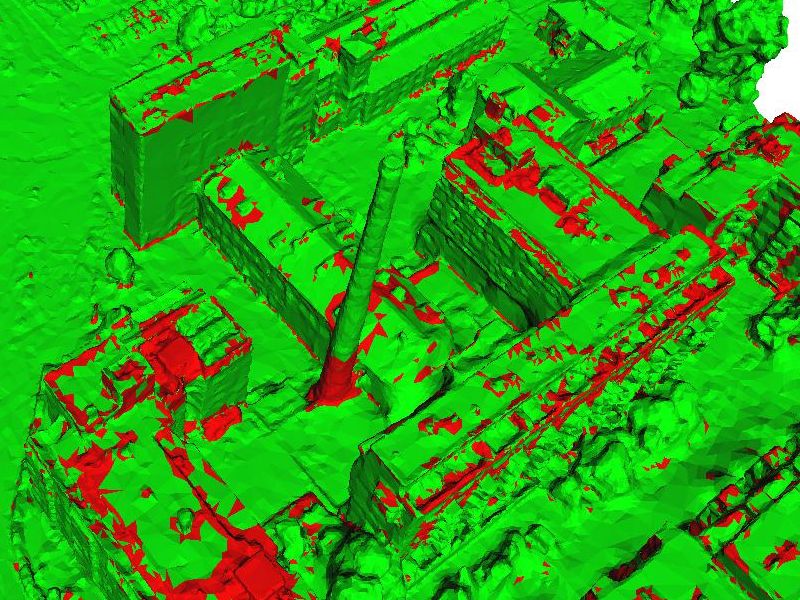} \\
        PointNet++\(^{Sp.}\)  &
		SPG\(^{Sp.}\) &
		SparseUNet\(^{Fc.}\) &
        Randla-net\(^{Fc.}\) \\
        \includegraphics[width=0.23\textwidth]{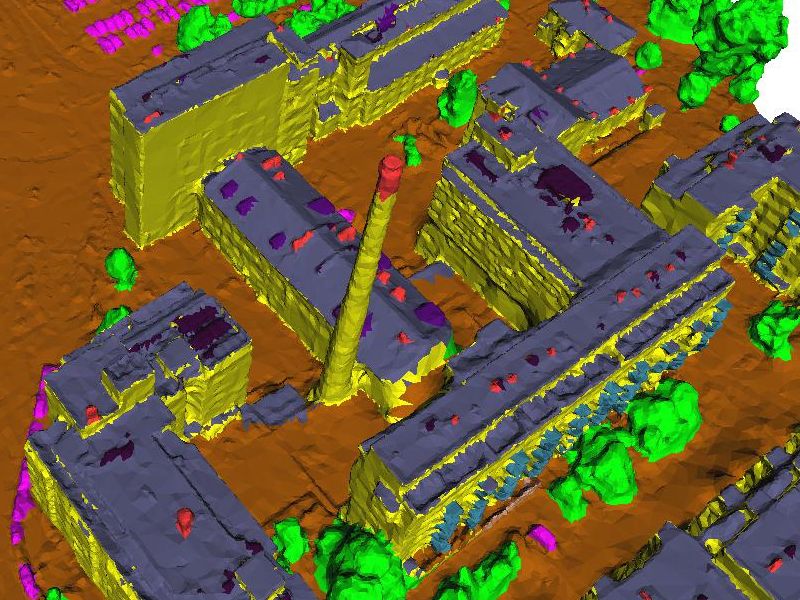} &
        \includegraphics[width=0.23\textwidth]{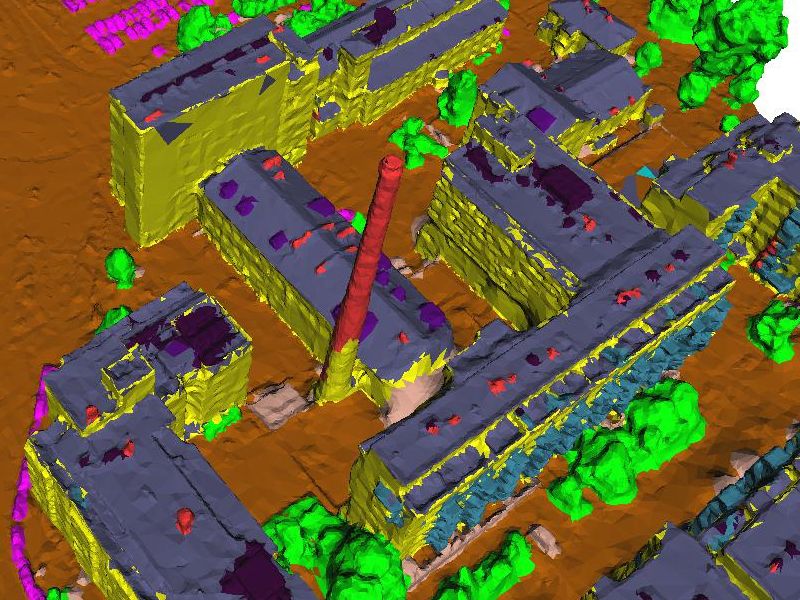} &
        \includegraphics[width=0.23\textwidth]{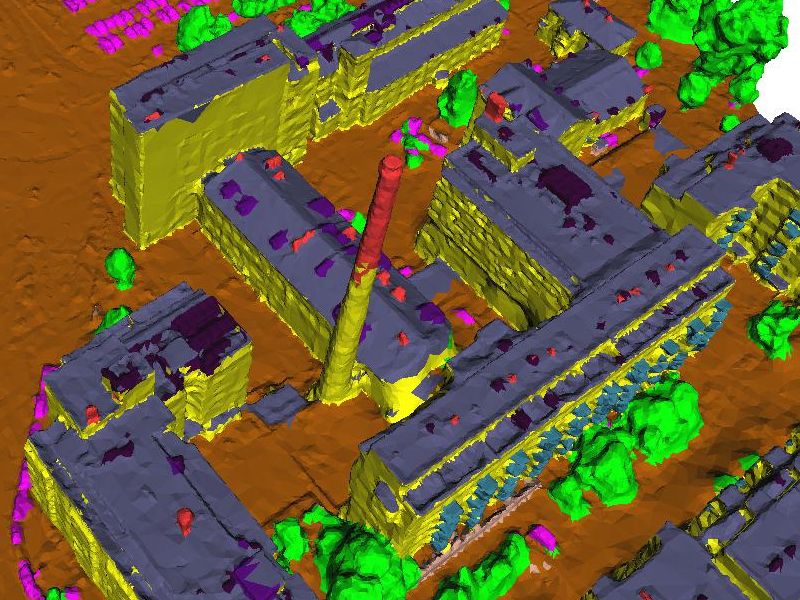} &
        \includegraphics[width=0.23\textwidth]{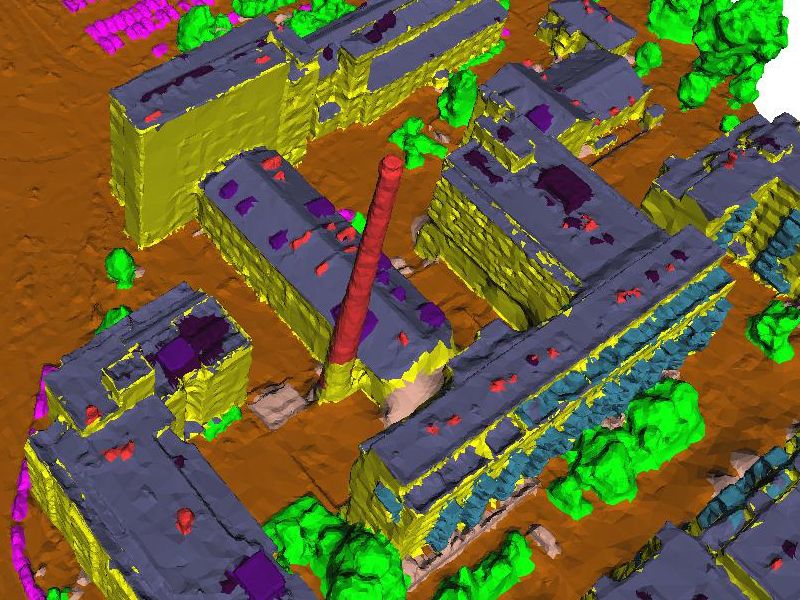} \\
        \includegraphics[width=0.23\textwidth]{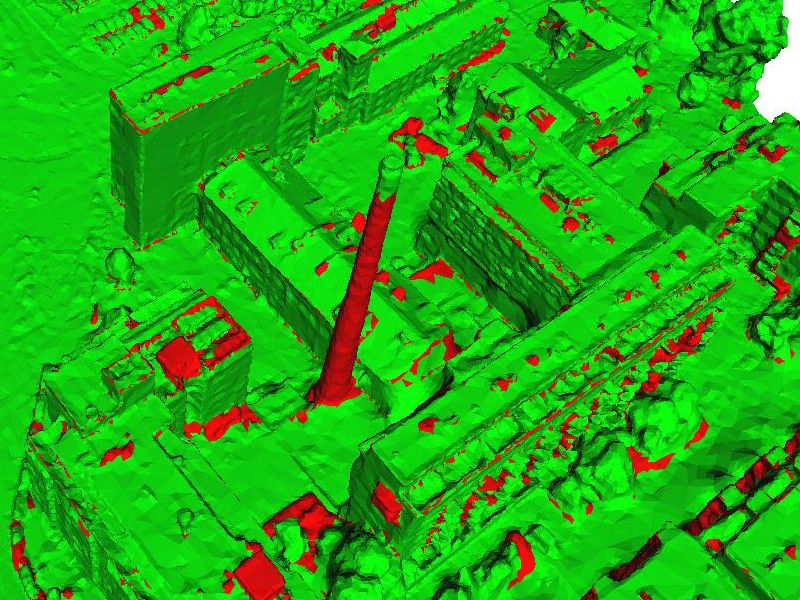} &
        \includegraphics[width=0.23\textwidth]{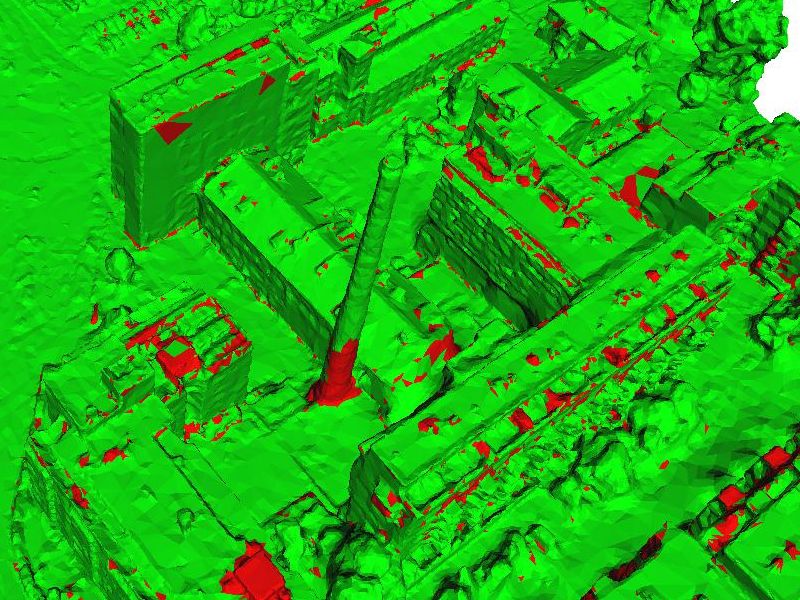} &
        \includegraphics[width=0.23\textwidth]{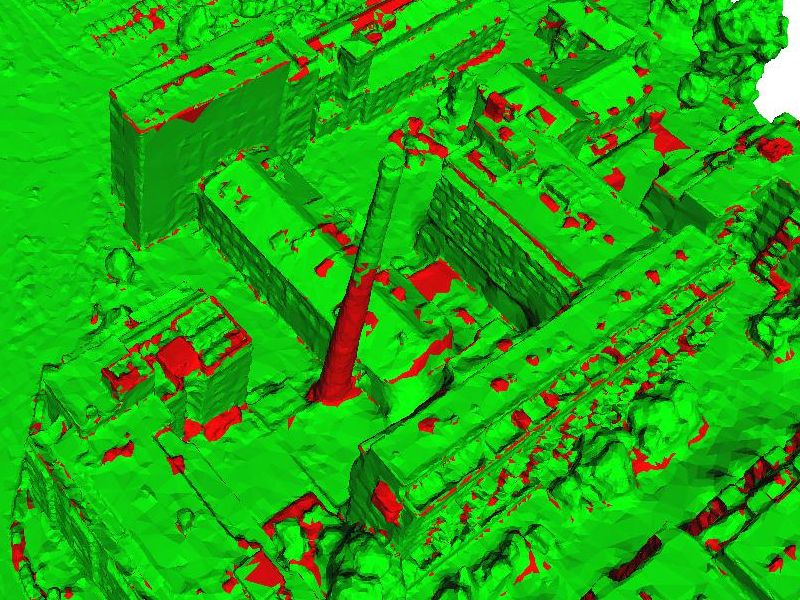} &
        \includegraphics[width=0.23\textwidth]{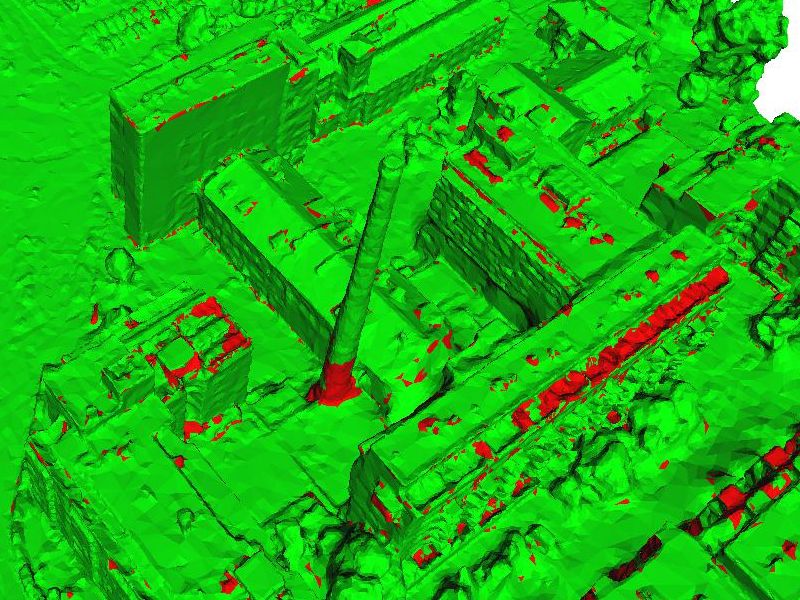} \\
        KPConv\(^{Fc.}\)  &
		PointNext\(^{Fc.}\) &
		PointTransV3\(^{Fc.}\) &
        PointVector\(^{Fc.}\) \\
    \end{tabular}  
    \caption{Qualitative analysis of semantic segmentation and error maps in the face labeling track for all methods except PointNet~\cite{qi2017pointnet} in the first scenario. \(^{Fc.}\) and \(^{Sp.}\) denote face-centered and superpixel sampling, respectively.}
    \label{fig:face_label_quality_scene1}
\end{figure*}

\begin{figure*}[!ht]
    \centering
    \begin{tabular}{cccc}
        \includegraphics[width=0.23\textwidth]{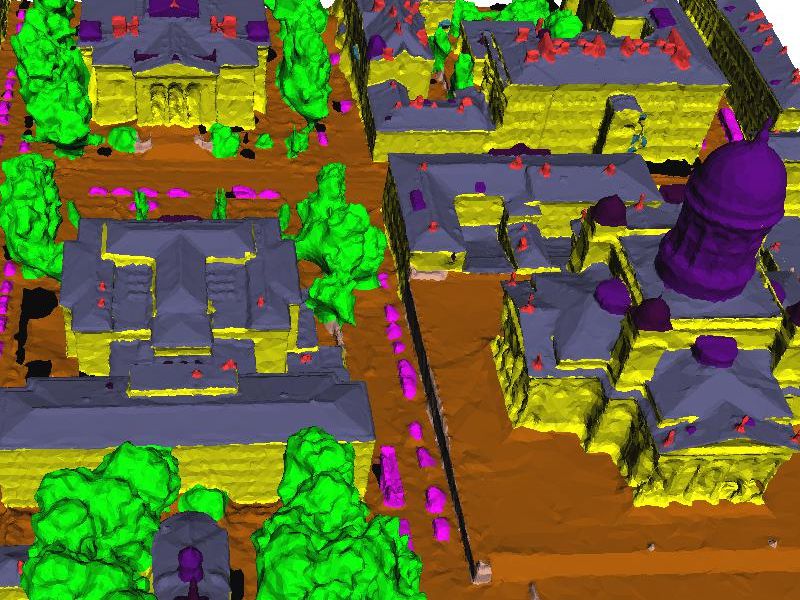} &
        \includegraphics[width=0.23\textwidth]{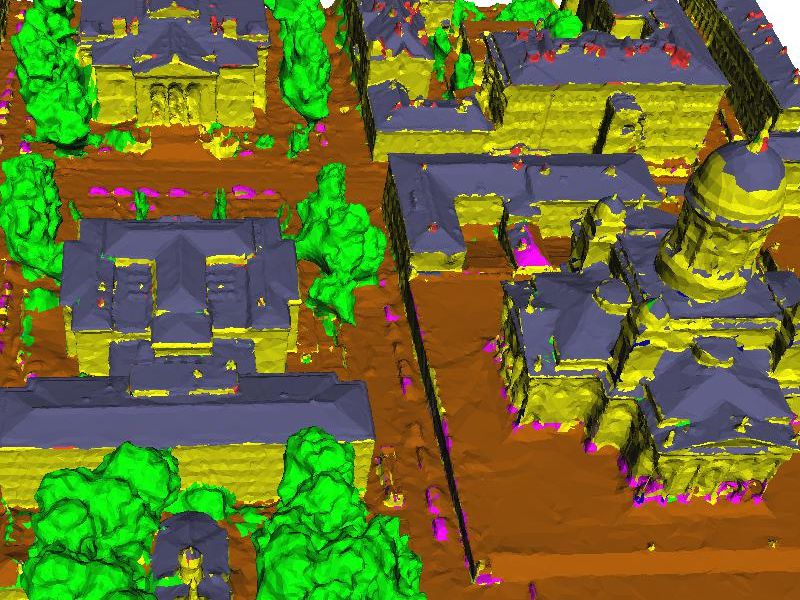} &
        \includegraphics[width=0.23\textwidth]{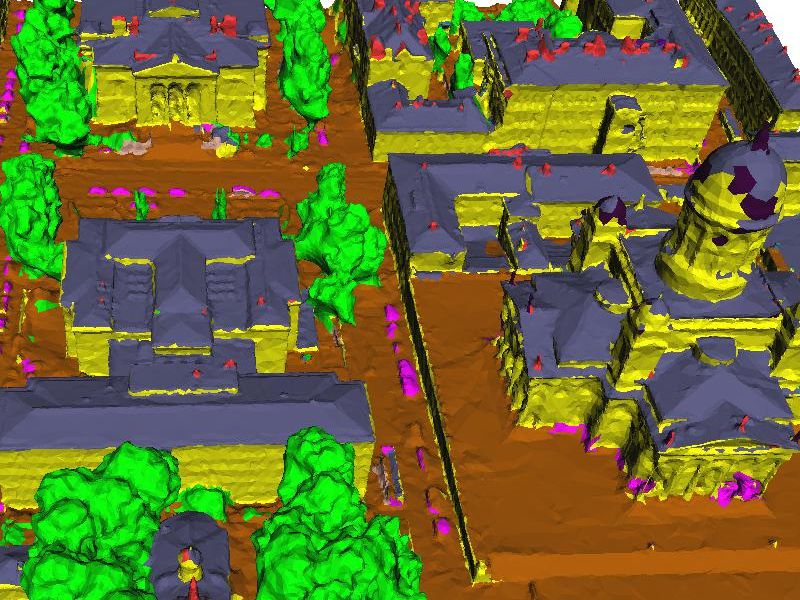} &
        \includegraphics[width=0.23\textwidth]{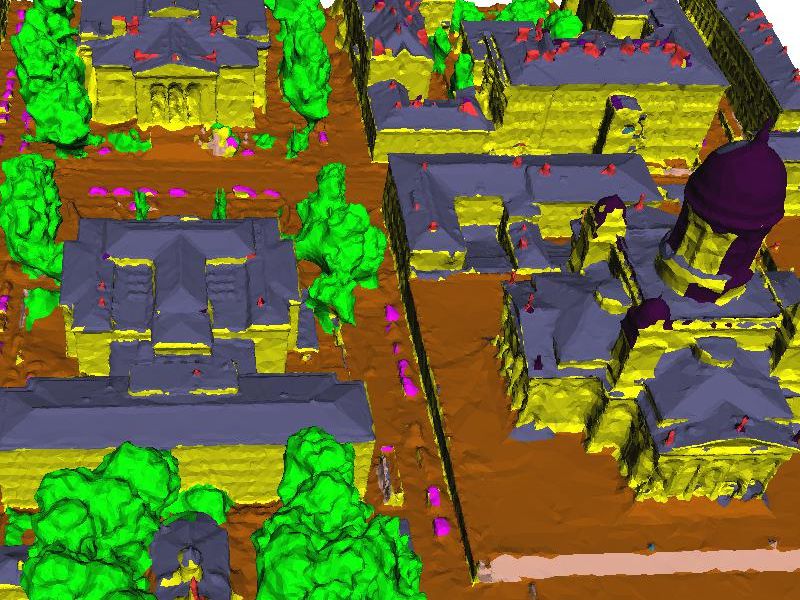} \\	
        \includegraphics[width=0.23\textwidth]{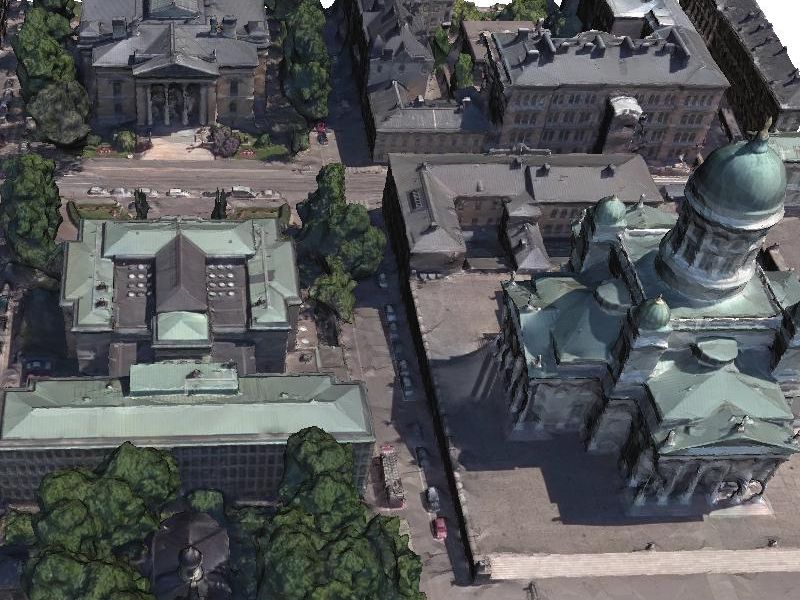} &
        \includegraphics[width=0.23\textwidth]{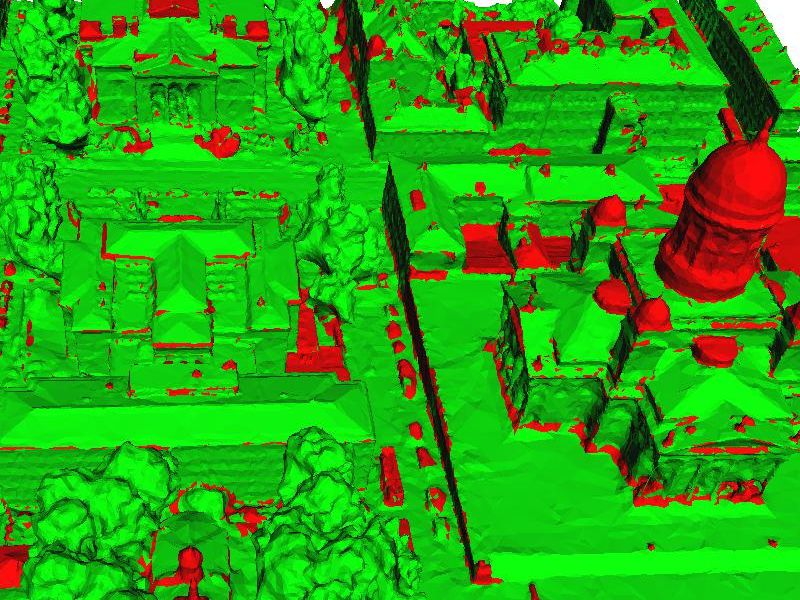} &
        \includegraphics[width=0.23\textwidth]{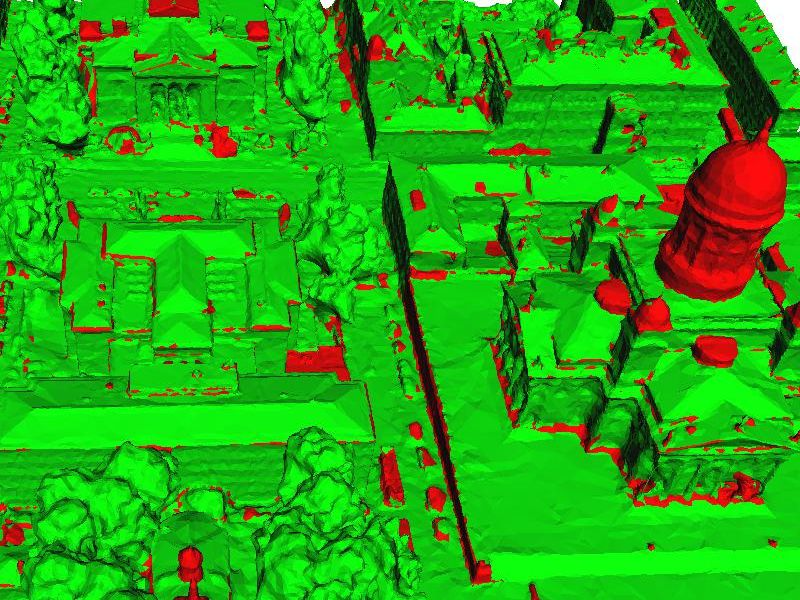} &
        \includegraphics[width=0.23\textwidth]{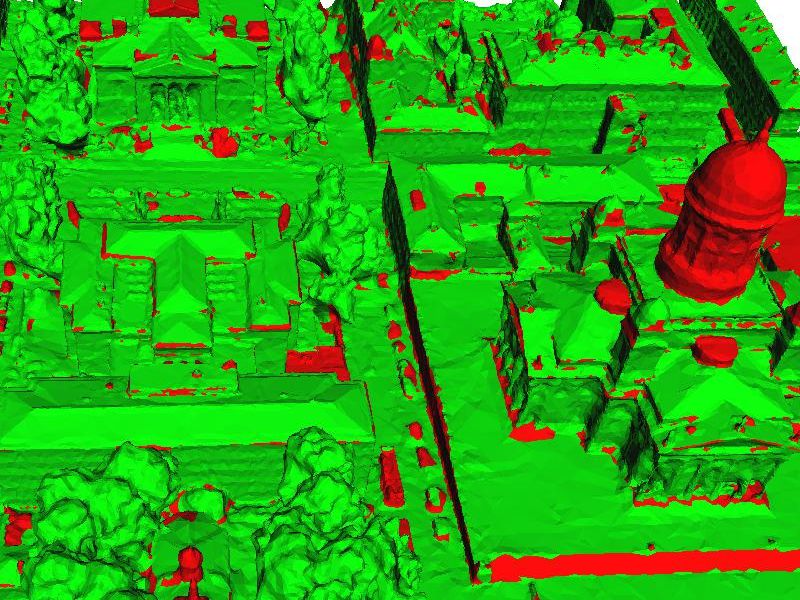} \\
        Truth (top), Mesh (bottom)   &
		RF\_MRF &
		SUM\_RF &
        PSSNet \\
        \includegraphics[width=0.23\textwidth]{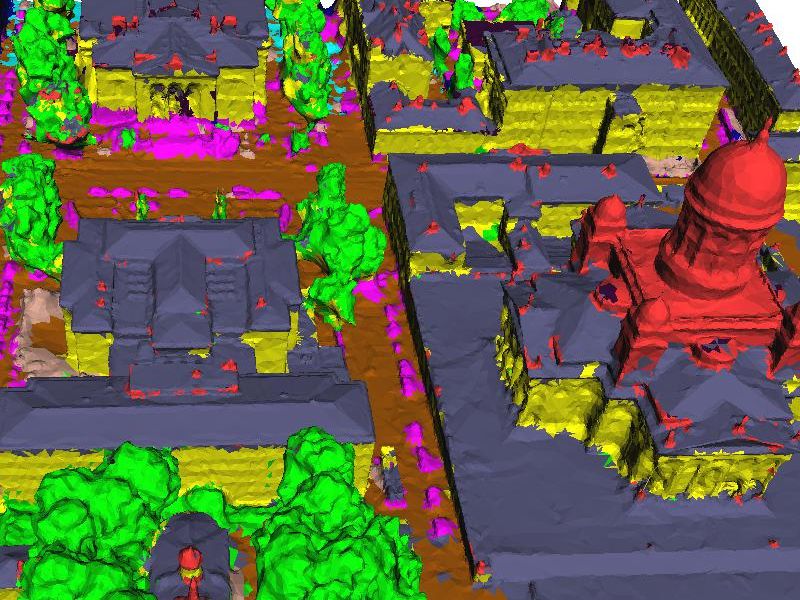} &
        \includegraphics[width=0.23\textwidth]{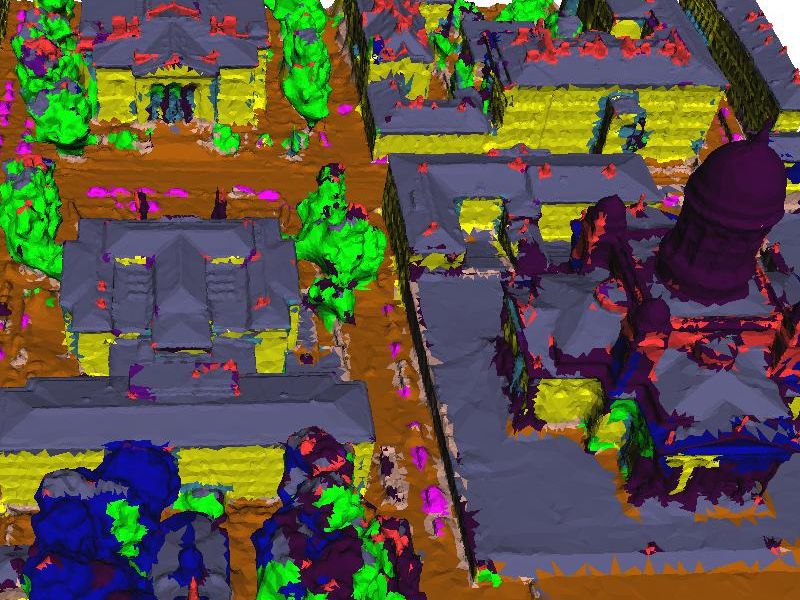} &
        \includegraphics[width=0.23\textwidth]{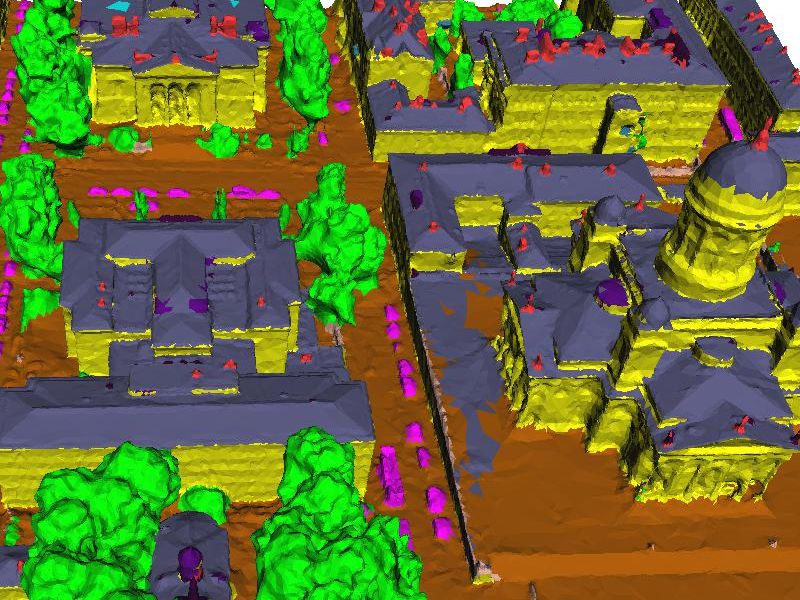} &
        \includegraphics[width=0.23\textwidth]{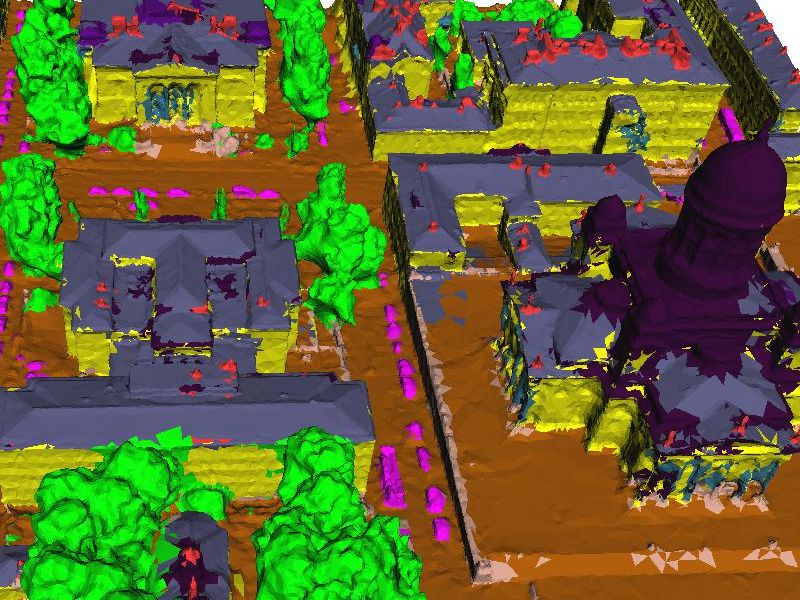} \\
        \includegraphics[width=0.23\textwidth]{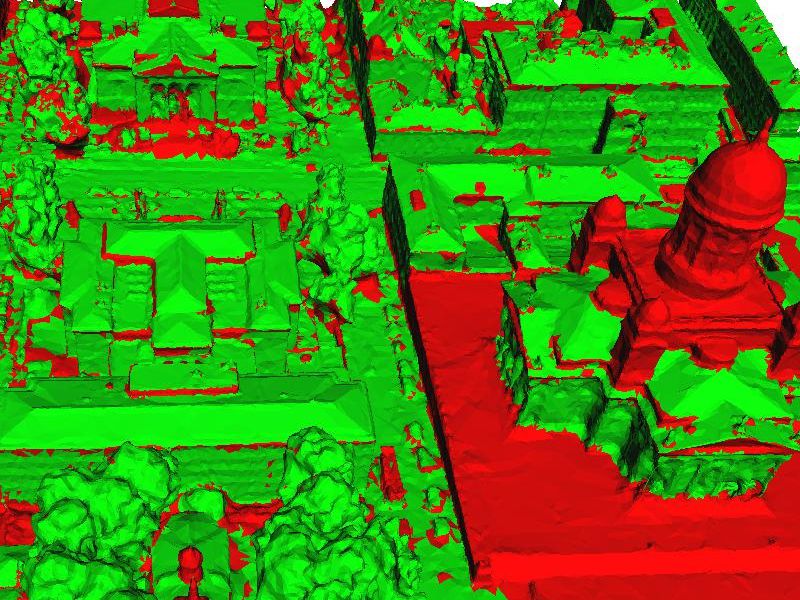} &
        \includegraphics[width=0.23\textwidth]{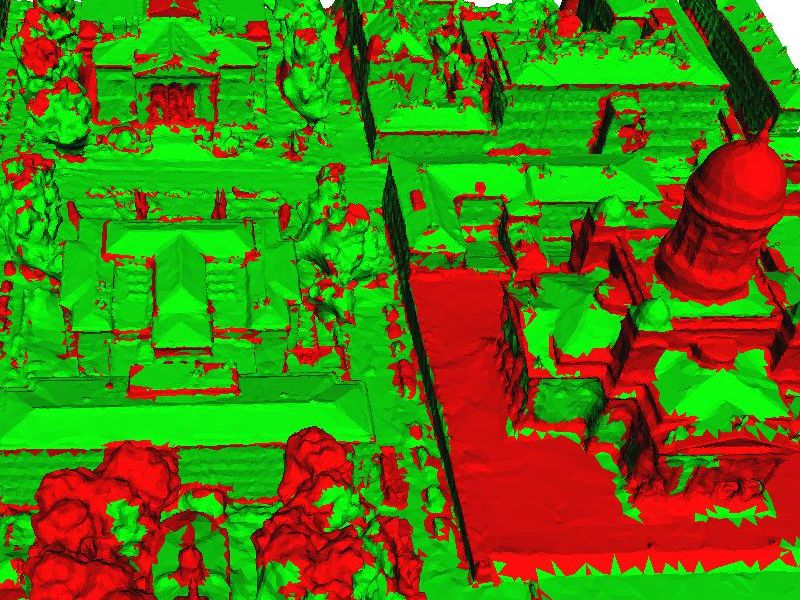} &
        \includegraphics[width=0.23\textwidth]{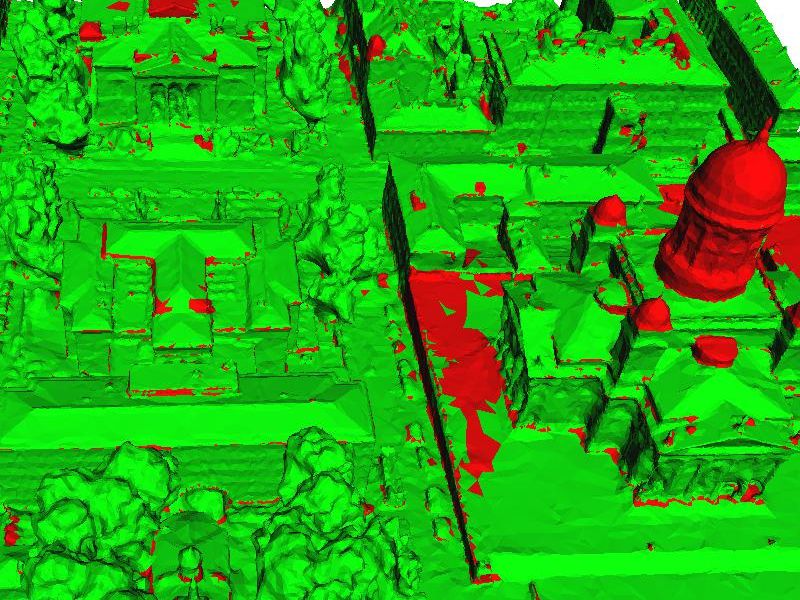} &
        \includegraphics[width=0.23\textwidth]{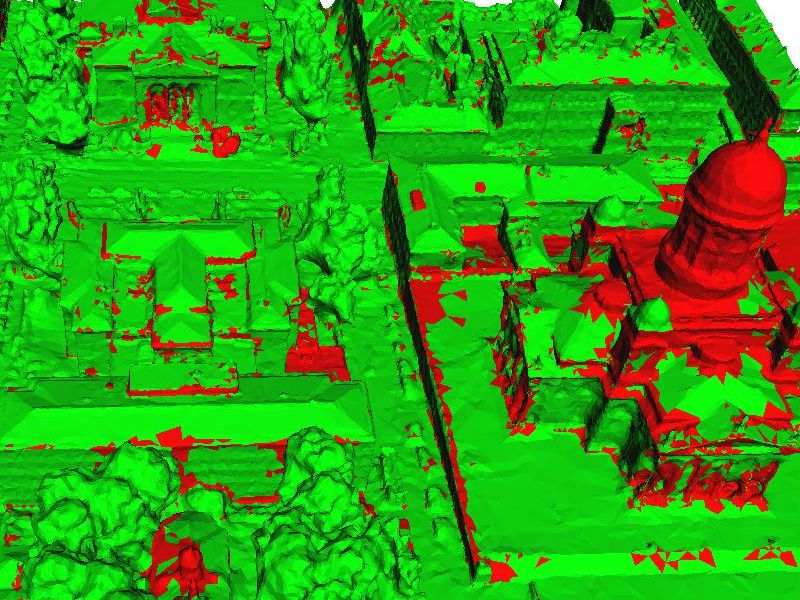} \\
        PointNet++\(^{Sp.}\)  &
		SPG\(^{Sp.}\) &
		SparseUNet\(^{Fc.}\) &
        Randla-net\(^{Fc.}\) \\
        \includegraphics[width=0.23\textwidth]{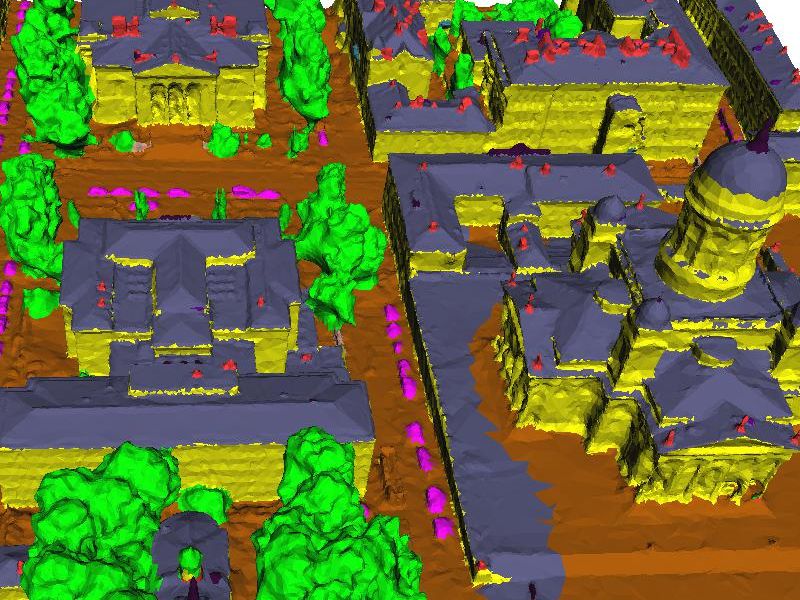} &
        \includegraphics[width=0.23\textwidth]{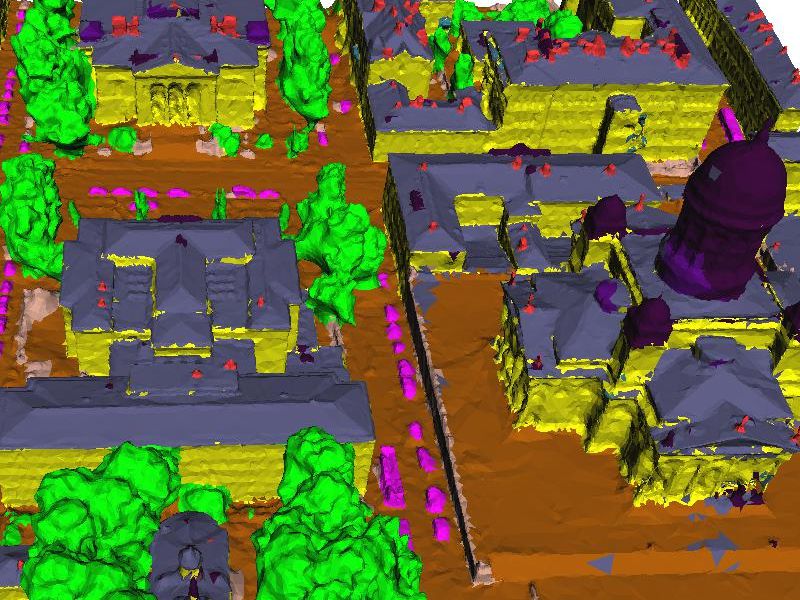} &
        \includegraphics[width=0.23\textwidth]{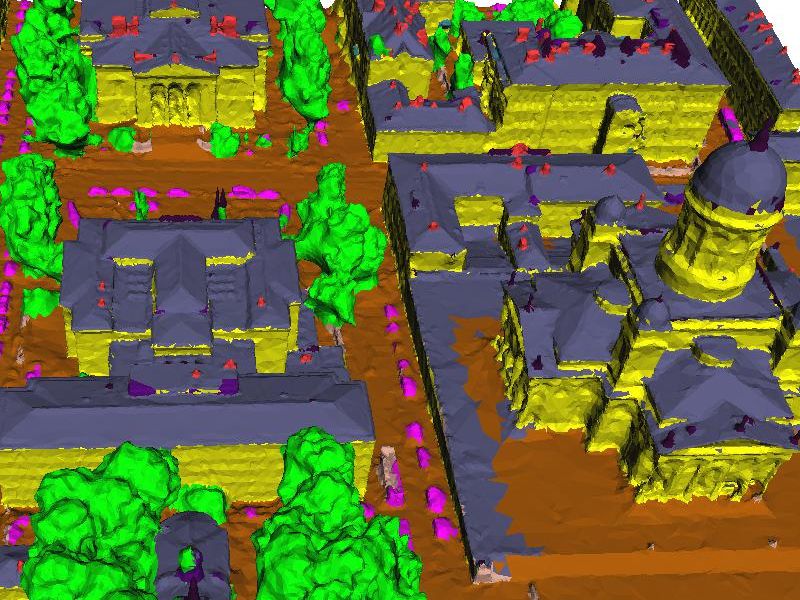} &
        \includegraphics[width=0.23\textwidth]{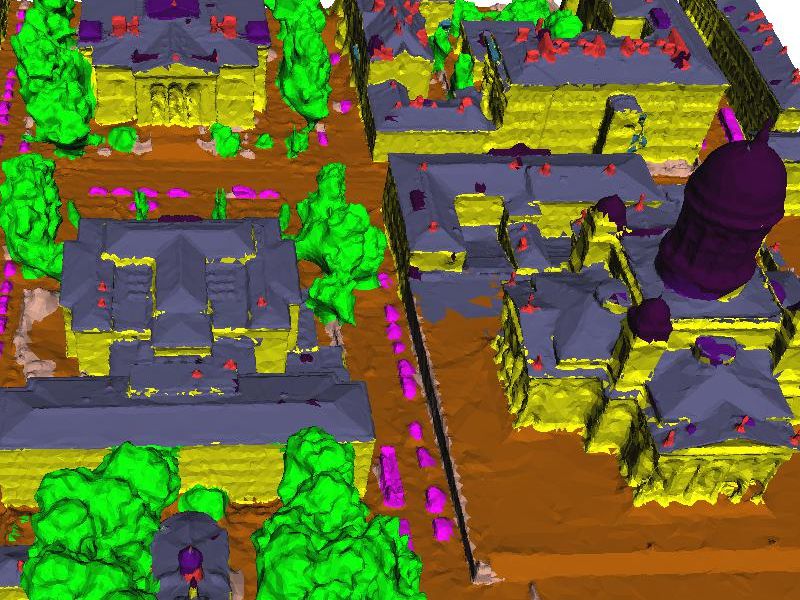} \\
        \includegraphics[width=0.23\textwidth]{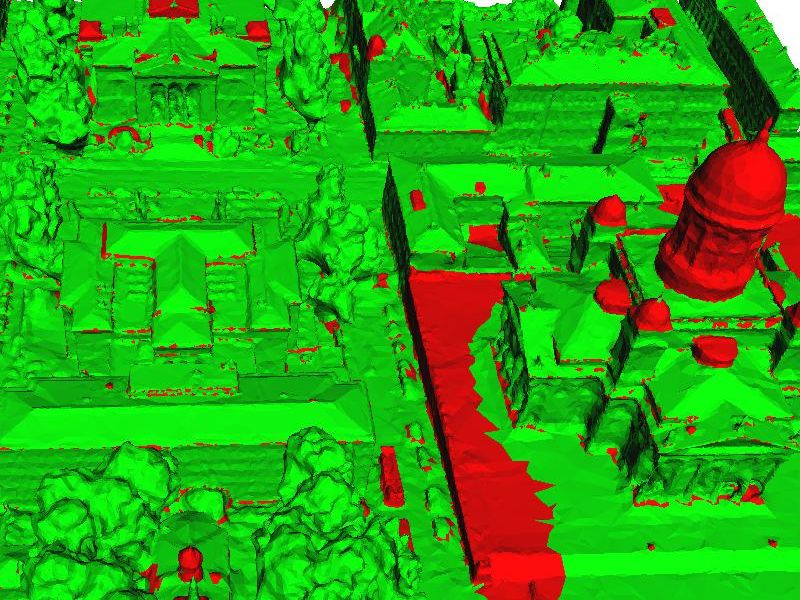} &
        \includegraphics[width=0.23\textwidth]{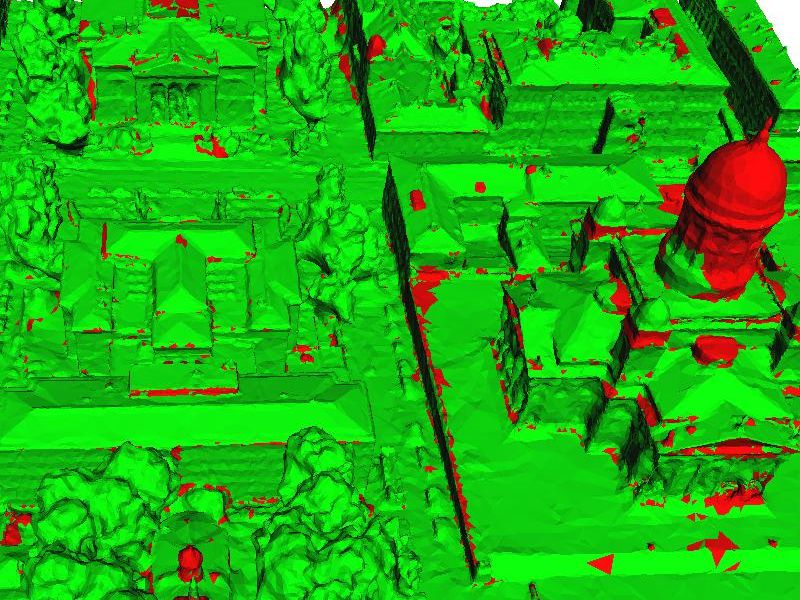} &
        \includegraphics[width=0.23\textwidth]{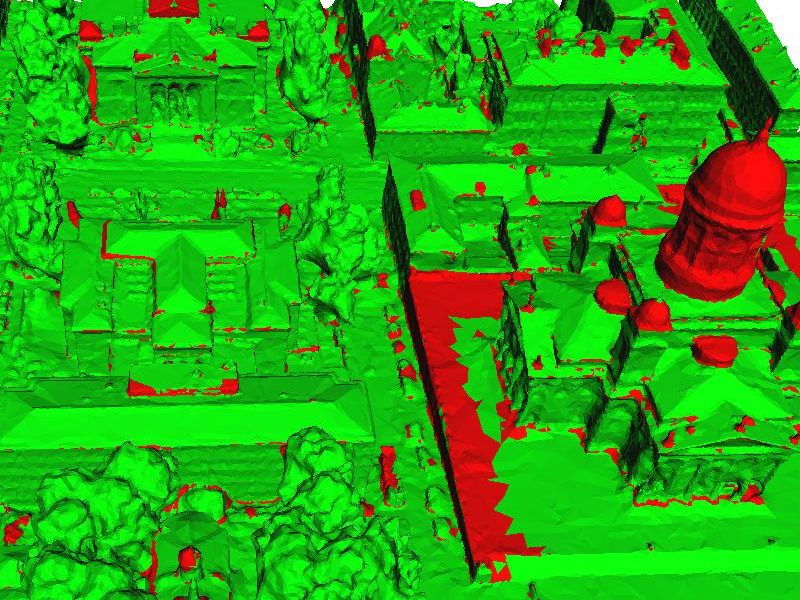} &
        \includegraphics[width=0.23\textwidth]{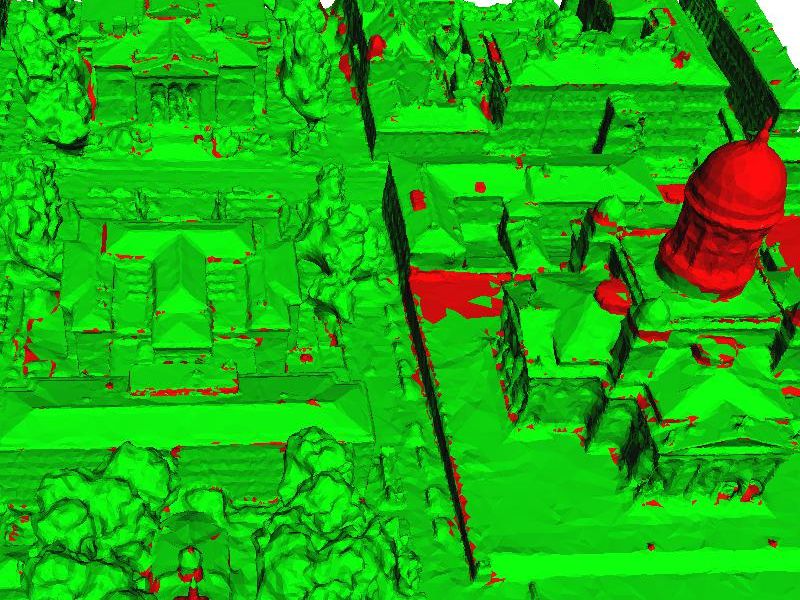} \\
        KPConv\(^{Fc.}\)  &
		PointNext\(^{Fc.}\) &
		PointTransV3\(^{Fc.}\) &
        PointVector\(^{Fc.}\) \\
    \end{tabular}
    \caption{Qualitative analysis of semantic segmentation and error maps in the face labeling track for all methods except PointNet~\cite{qi2017pointnet} in the second scenario. \(^{Fc.}\) and \(^{Sp.}\) denote face-centered and superpixel sampling, respectively.}
    \label{fig:face_label_quality_scene2}
\end{figure*}

\noindent \textit{\textbf{1) Face labeling track.}} ~\cref{tab:face_labels} provides a detailed comparison of results for all face-labeled classes.
Due to class imbalance, most methods show better performance in categories with more samples and poorer performance in categories with fewer samples. 
We conducted qualitative analyses on all methods except PointNet for two scenarios, as shown in~\cref{fig:face_label_quality_scene1} and~\cref{fig:face_label_quality_scene2}.

\begin{figure*}[!t]
    \centering
    \resizebox{\textwidth}{!}{
    \begin{tabular}{ccccc}
        \includegraphics[width=0.23\textwidth]{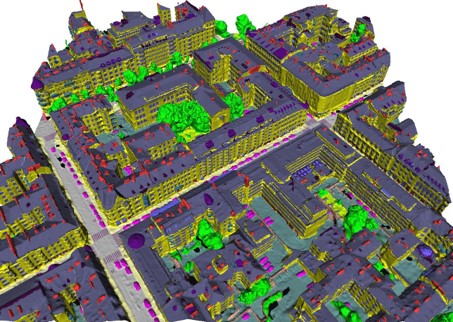} &
        \includegraphics[width=0.23\textwidth]{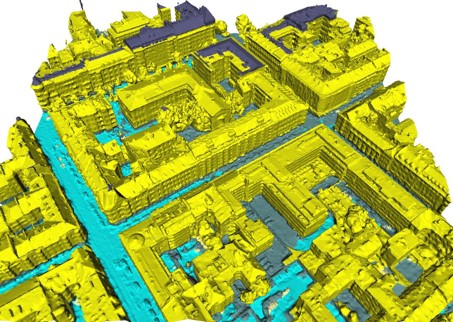} &
        \includegraphics[width=0.23\textwidth]{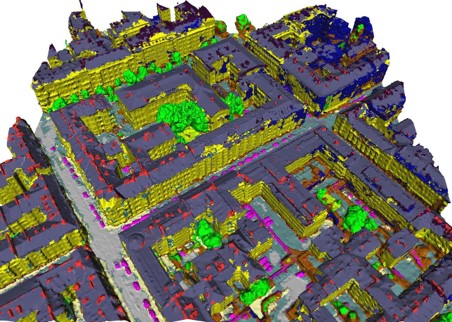} &
        \includegraphics[width=0.23\textwidth]{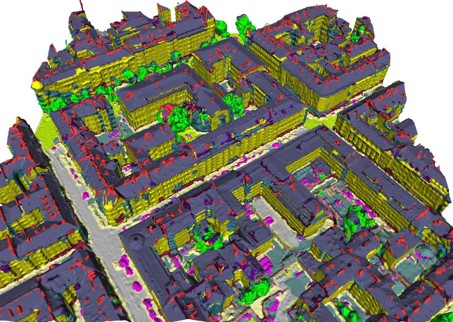} &
        \includegraphics[width=0.23\textwidth]{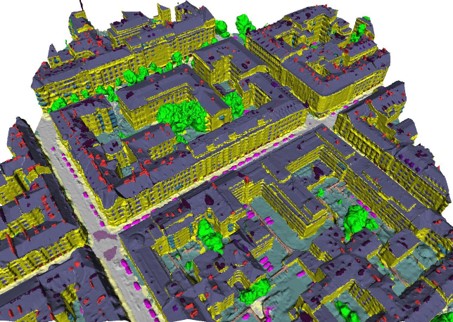} \\	
        \includegraphics[width=0.23\textwidth]{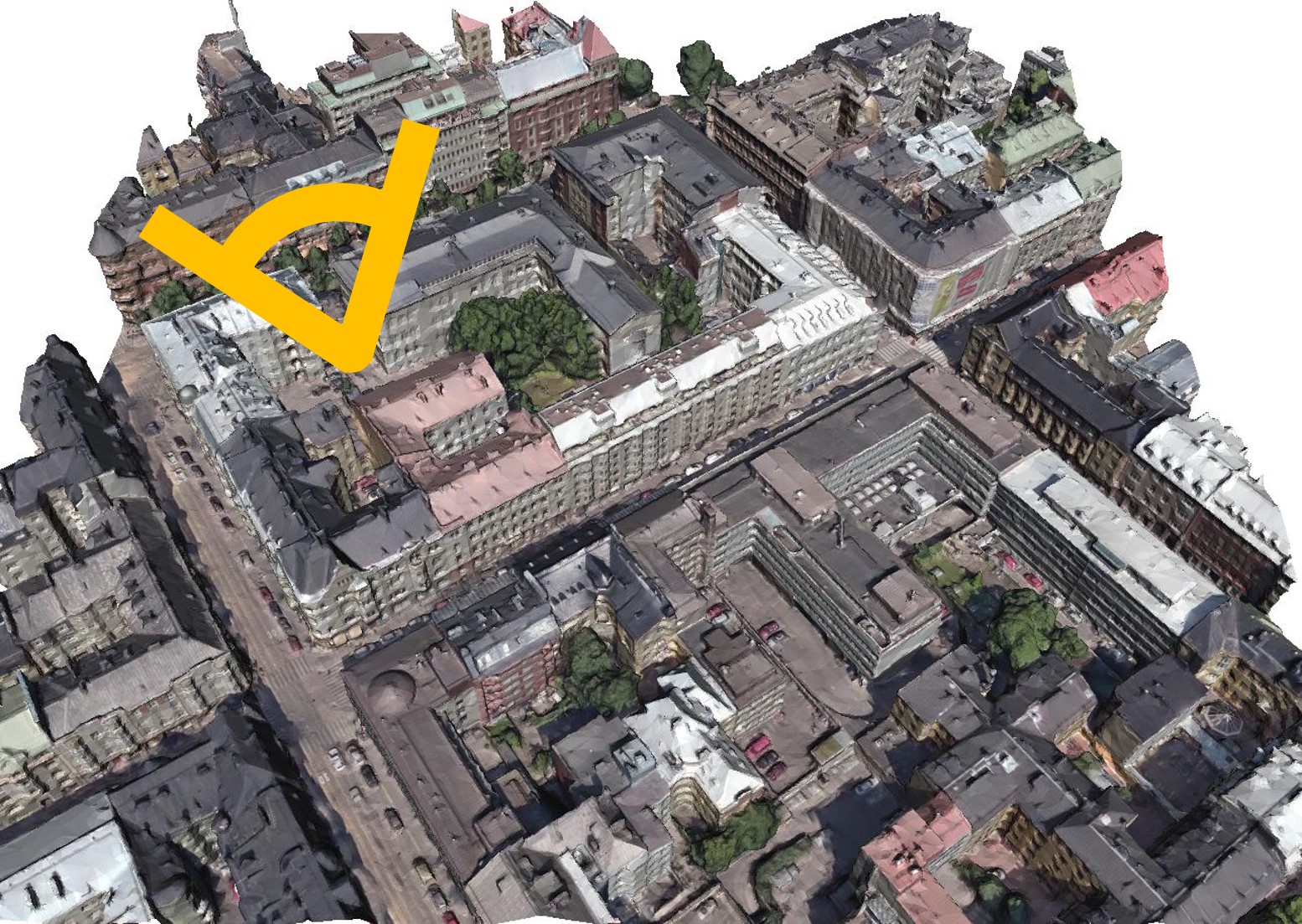} &
        \includegraphics[width=0.23\textwidth]{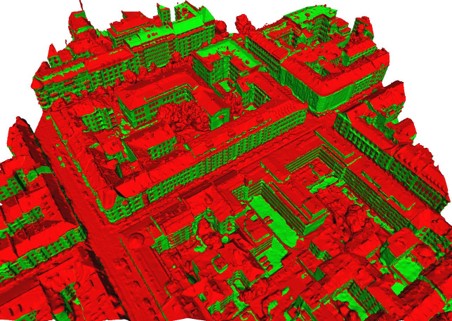} &
        \includegraphics[width=0.23\textwidth]{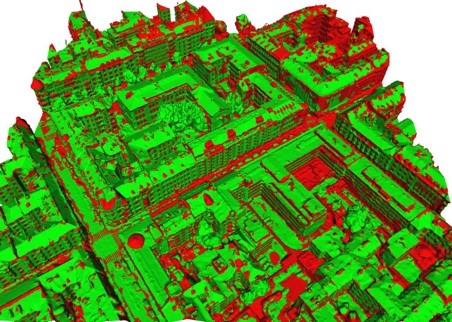} &
        \includegraphics[width=0.23\textwidth]{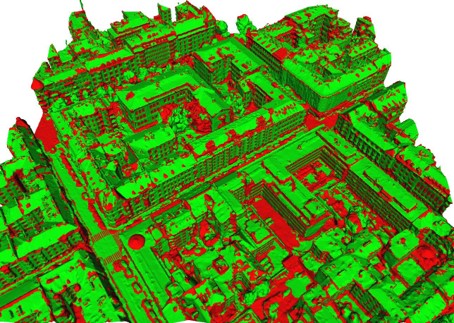} &
        \includegraphics[width=0.23\textwidth]{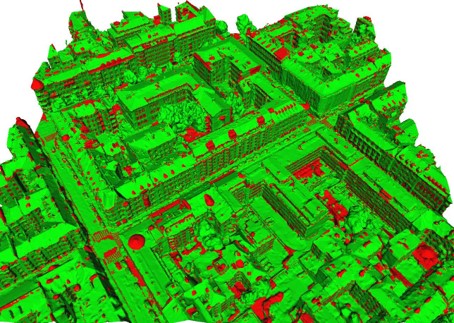} \\	
        Truth (top), Mesh (bottom)   &
		PointNet\(^{Sp.}\) &
		PointNet++\(^{Sp.}\) &
        SPG\(^{Sp.}\) & 
        SparseUNet\(^{Rd.}\) \\
        \includegraphics[width=0.23\textwidth]{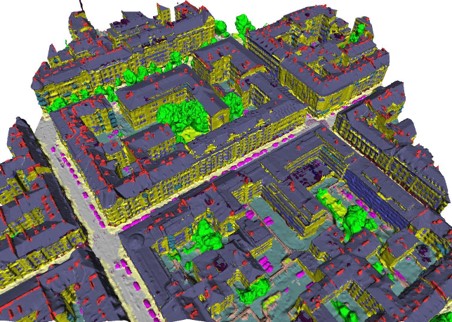} &
        \includegraphics[width=0.23\textwidth]{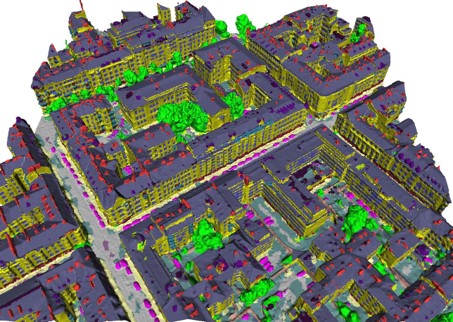} &
        \includegraphics[width=0.23\textwidth]{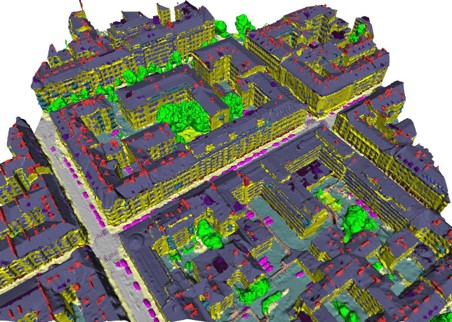} &
        \includegraphics[width=0.23\textwidth]{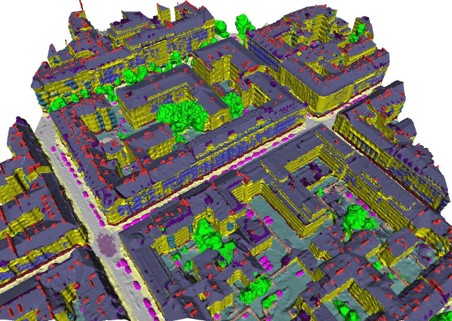} &
        \includegraphics[width=0.23\textwidth]{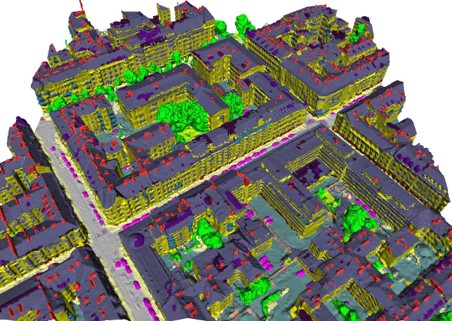} \\	
        \includegraphics[width=0.23\textwidth]{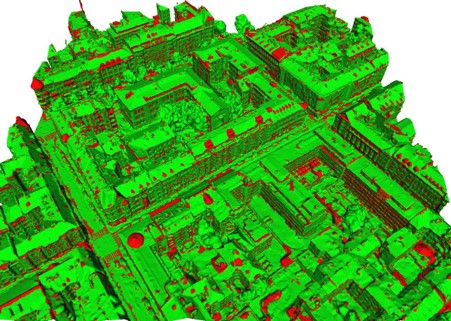} &
        \includegraphics[width=0.23\textwidth]{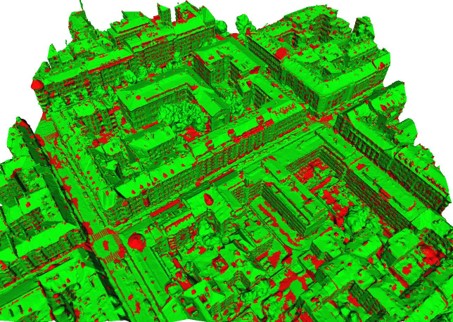} &
        \includegraphics[width=0.23\textwidth]{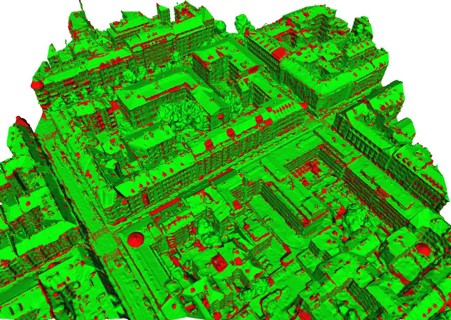} &
        \includegraphics[width=0.23\textwidth]{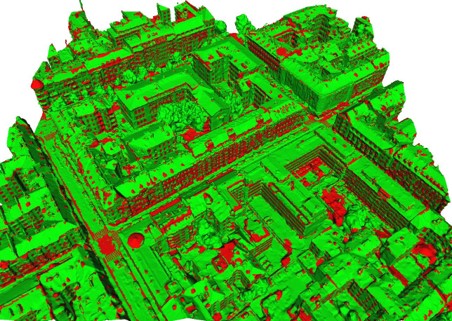} &
        \includegraphics[width=0.23\textwidth]{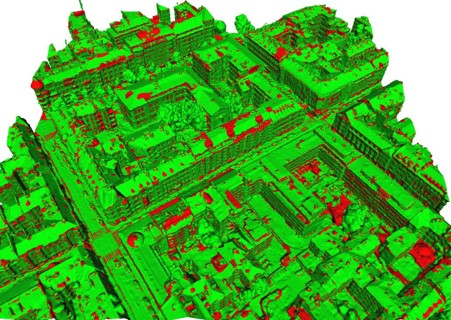} \\	
		Randla-net\(^{Sp.}\) &
		KPConv\(^{Sp.}\) &
        PointNext\(^{Po.}\) & 
        PointTransV3\(^{Rd.}\) &
        PointVector\(^{Sp.}\)\\
        \includegraphics[width=0.23\textwidth]{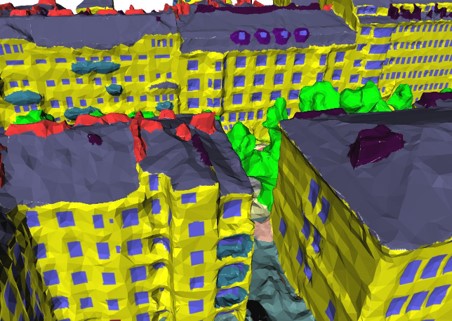} &
        \includegraphics[width=0.23\textwidth]{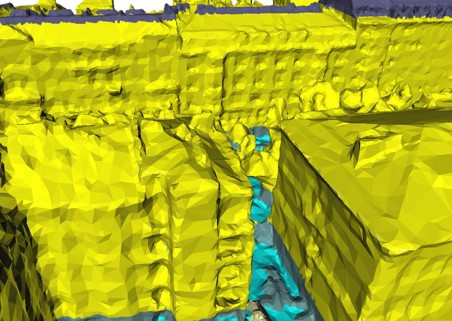} &
        \includegraphics[width=0.23\textwidth]{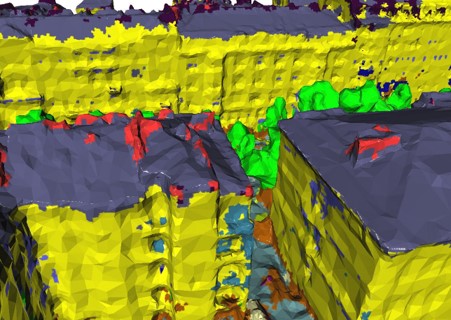} &
        \includegraphics[width=0.23\textwidth]{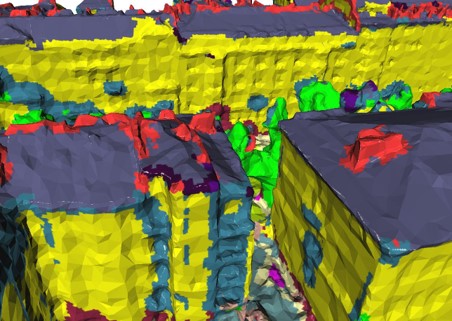} &
        \includegraphics[width=0.23\textwidth]{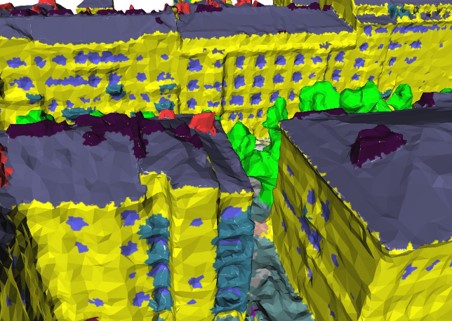} \\	
        \includegraphics[width=0.23\textwidth]{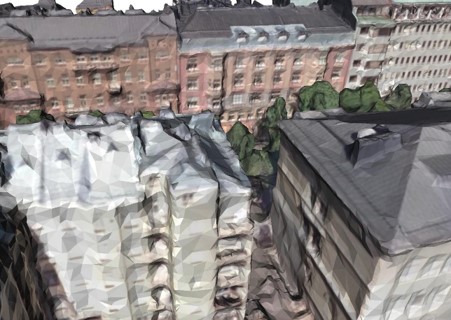} &
        \includegraphics[width=0.23\textwidth]{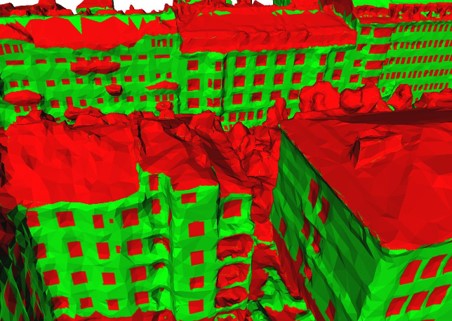} &
        \includegraphics[width=0.23\textwidth]{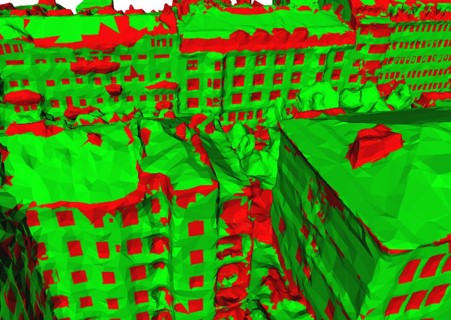} &
        \includegraphics[width=0.23\textwidth]{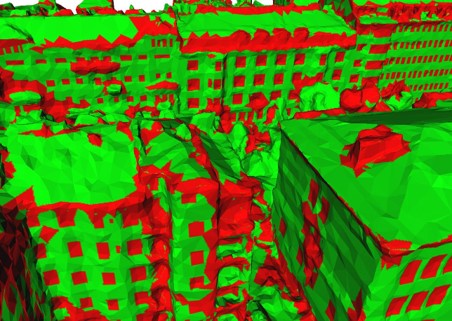} &
        \includegraphics[width=0.23\textwidth]{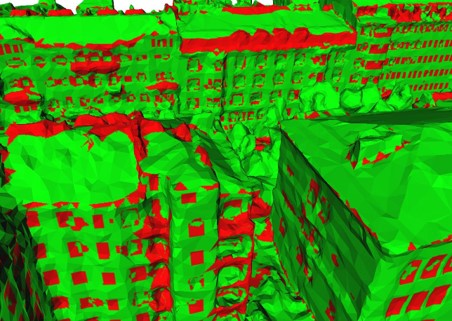} \\	
        Truth (top), Mesh (bottom)   &
		PointNet\(^{Sp.}\) &
		PointNet++\(^{Sp.}\) &
        SPG\(^{Sp.}\) & 
        SparseUNet\(^{Rd.}\) \\
        \includegraphics[width=0.23\textwidth]{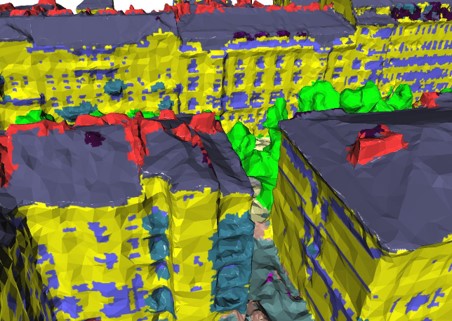} &
        \includegraphics[width=0.23\textwidth]{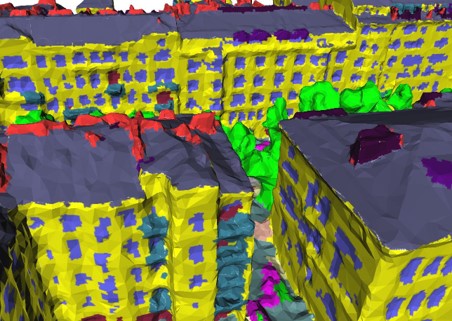} &
        \includegraphics[width=0.23\textwidth]{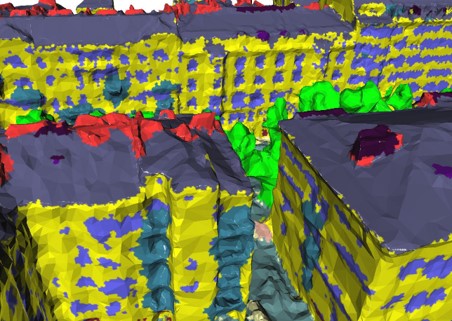} &
        \includegraphics[width=0.23\textwidth]{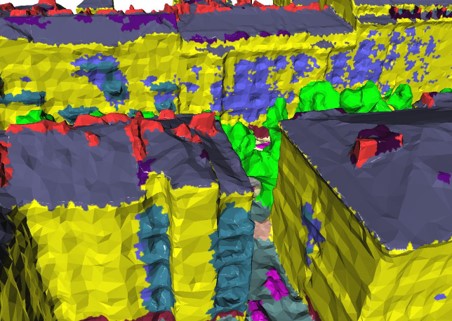} &
        \includegraphics[width=0.23\textwidth]{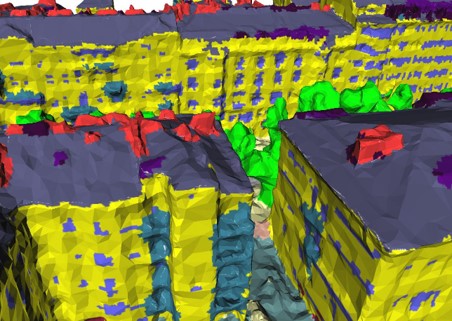} \\	
        \includegraphics[width=0.23\textwidth]{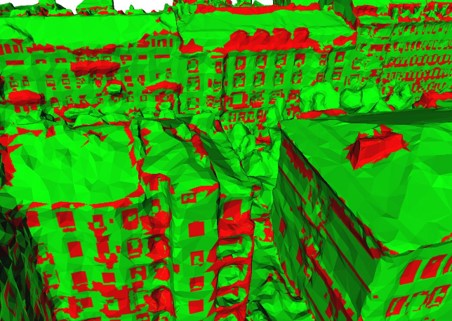} &
        \includegraphics[width=0.23\textwidth]{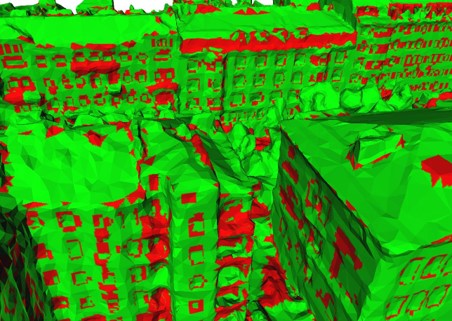} &
        \includegraphics[width=0.23\textwidth]{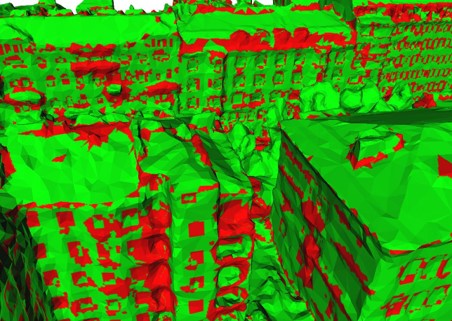} &
        \includegraphics[width=0.23\textwidth]{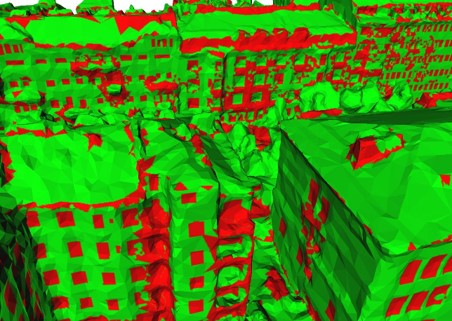} &
        \includegraphics[width=0.23\textwidth]{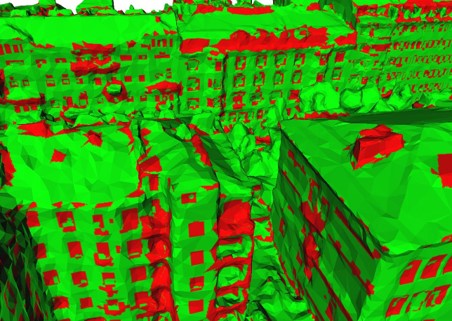} \\	
		Randla-net\(^{Sp.}\) &
		KPConv\(^{Sp.}\) &
        PointNext\(^{Po.}\) & 
        PointTransV3\(^{Rd.}\) &
        PointVector\(^{Sp.}\)\\
    \end{tabular}
    }
    \caption{Qualitative analysis of semantic segmentation and error maps in the pixel labeling track for all methods in the first scenario. \(^{Sp.}\) denotes superpixel sampling, \(^{Rd.}\) for random sampling, and \(^{Po.}\) for Poisson-disk sampling~\cite{cook1986stochastic}. The zoomed-in view direction is indicated in the input mesh image.}
    \label{fig:pixel_label_quality_scene1}
\end{figure*}

\begin{figure*}[!t]
    \centering
    \resizebox{\textwidth}{!}{
    \begin{tabular}{ccccc}
        \includegraphics[width=0.23\textwidth]{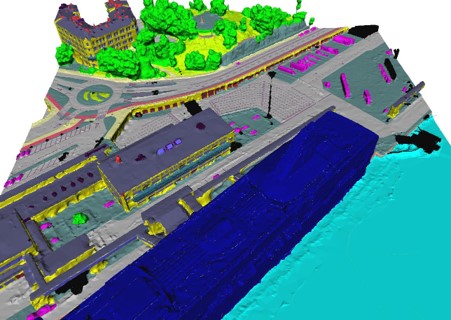} &
        \includegraphics[width=0.23\textwidth]{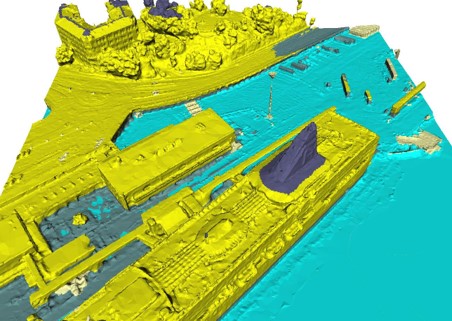} &
        \includegraphics[width=0.23\textwidth]{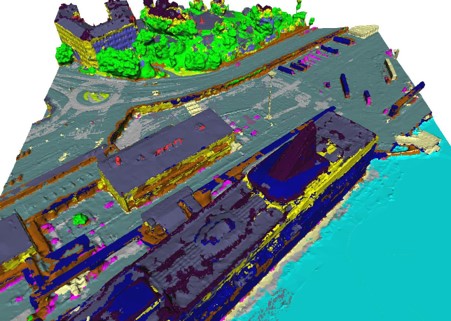} &
        \includegraphics[width=0.23\textwidth]{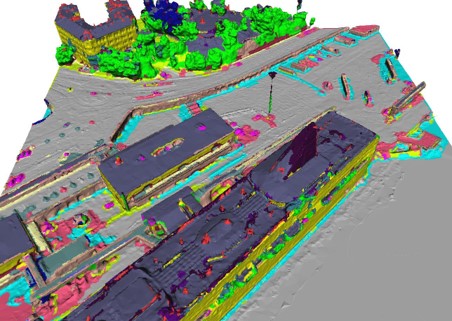} &
        \includegraphics[width=0.23\textwidth]{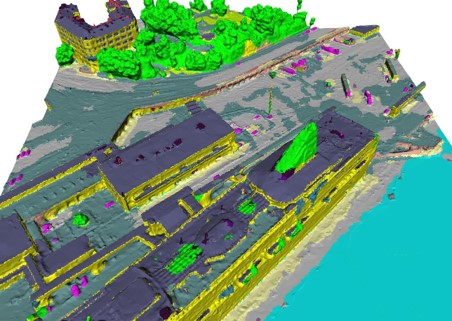} \\	
        \includegraphics[width=0.23\textwidth]{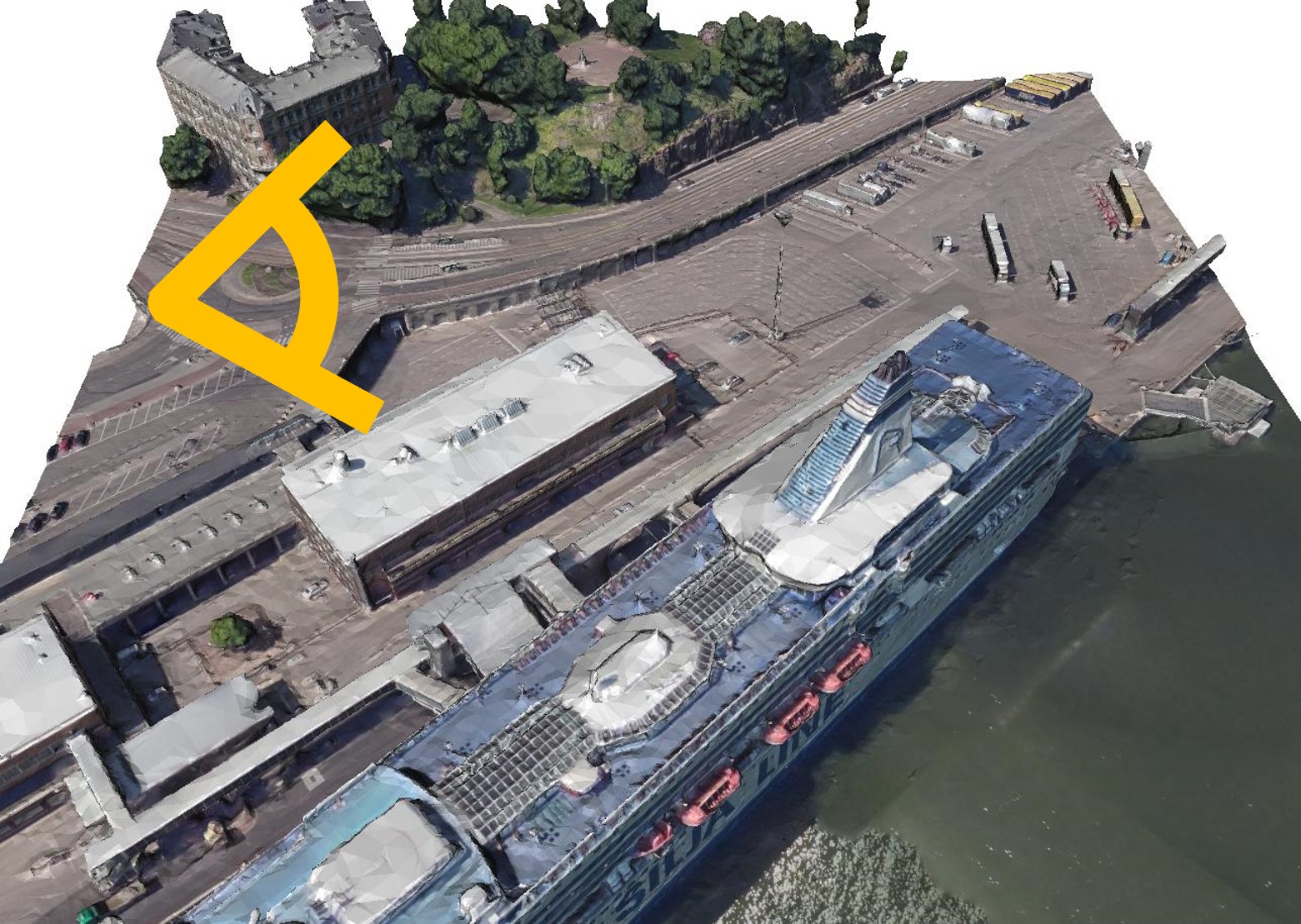} &
        \includegraphics[width=0.23\textwidth]{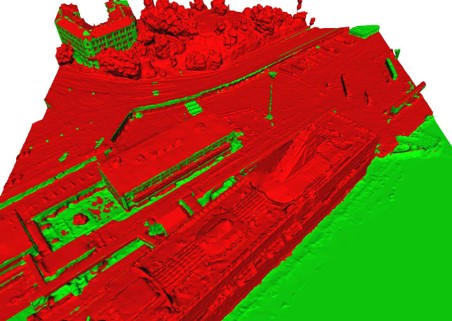} &
        \includegraphics[width=0.23\textwidth]{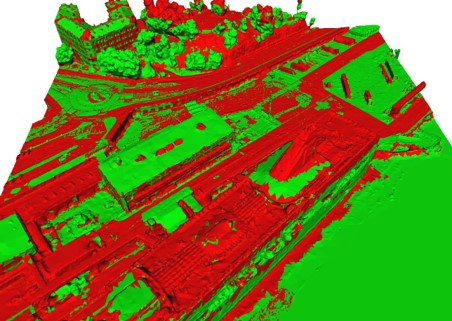} &
        \includegraphics[width=0.23\textwidth]{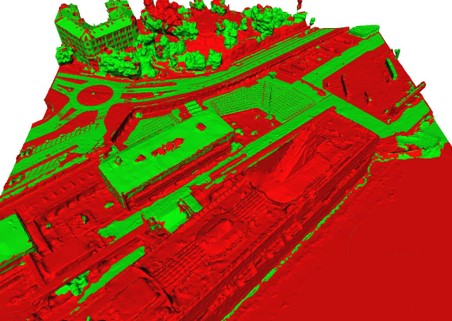} &
        \includegraphics[width=0.23\textwidth]{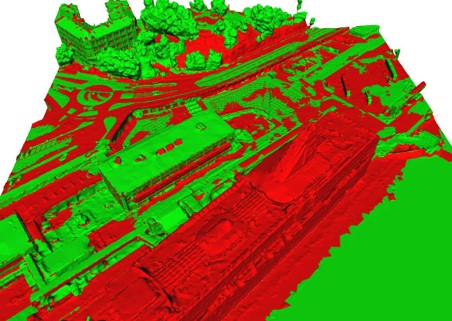} \\	
        Truth (top), Mesh (bottom)   &
		PointNet\(^{Sp.}\) &
		PointNet++\(^{Sp.}\) &
        SPG\(^{Sp.}\) & 
        SparseUNet\(^{Rd.}\) \\
        \includegraphics[width=0.23\textwidth]{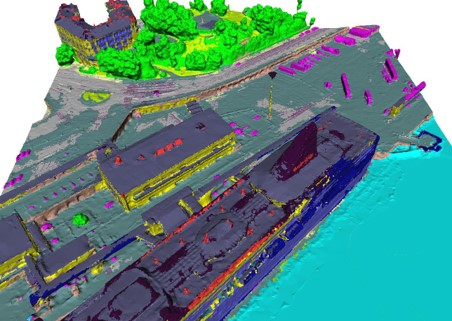} &
        \includegraphics[width=0.23\textwidth]{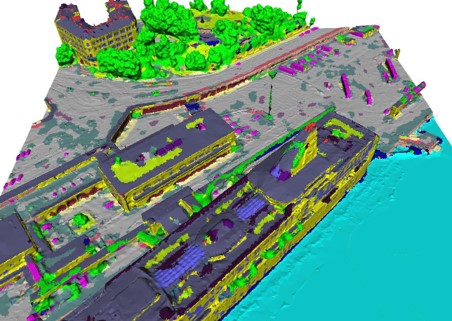} &
        \includegraphics[width=0.23\textwidth]{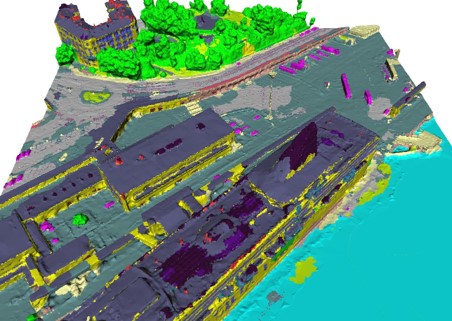} &
        \includegraphics[width=0.23\textwidth]{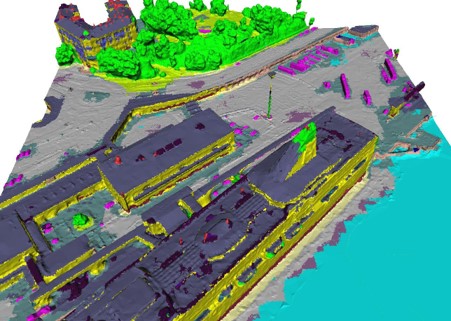} &
        \includegraphics[width=0.23\textwidth]{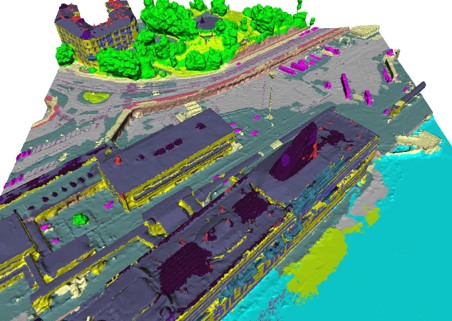} \\	
        \includegraphics[width=0.23\textwidth]{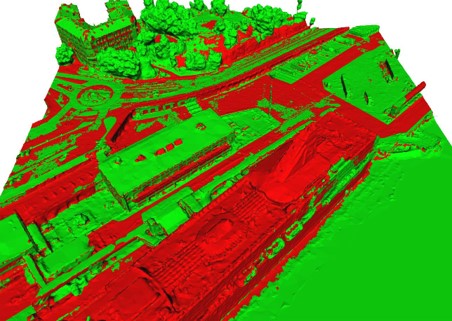} &
        \includegraphics[width=0.23\textwidth]{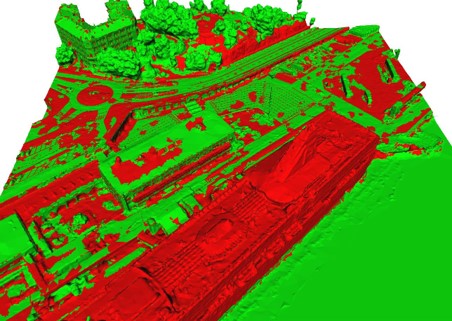} &
        \includegraphics[width=0.23\textwidth]{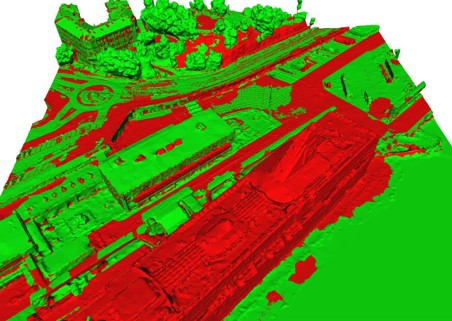} &
        \includegraphics[width=0.23\textwidth]{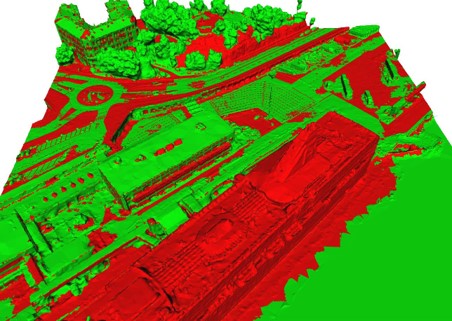} &
        \includegraphics[width=0.23\textwidth]{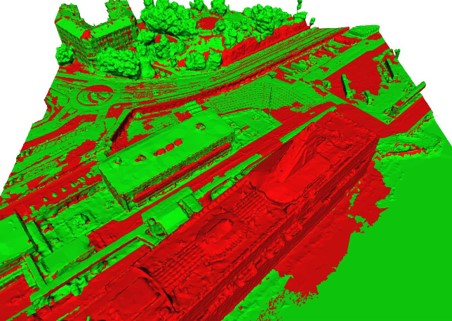} \\	
		Randla-net\(^{Sp.}\) &
		KPConv\(^{Sp.}\) &
        PointNext\(^{Po.}\) & 
        PointTransV3\(^{Rd.}\) &
        PointVector\(^{Sp.}\)\\
        \includegraphics[width=0.23\textwidth]{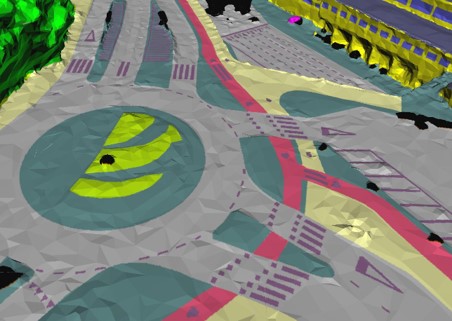} &
        \includegraphics[width=0.23\textwidth]{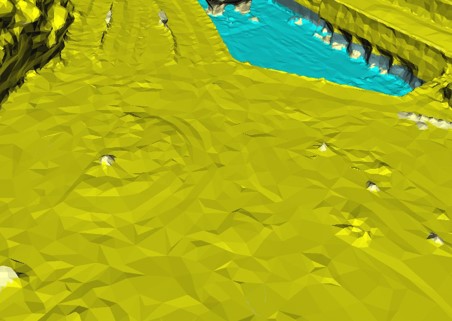} &
        \includegraphics[width=0.23\textwidth]{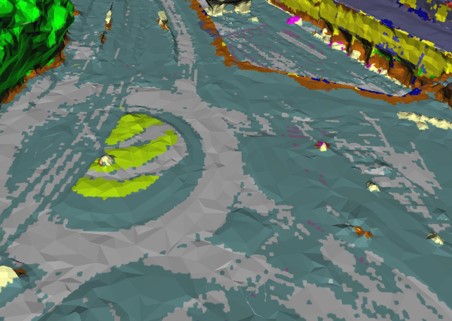} &
        \includegraphics[width=0.23\textwidth]{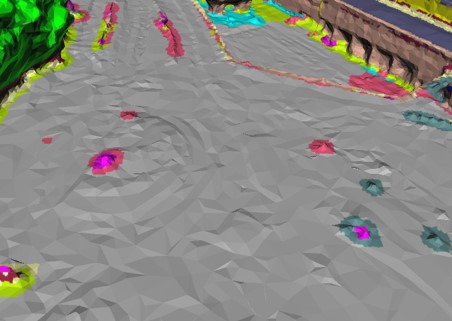} &
        \includegraphics[width=0.23\textwidth]{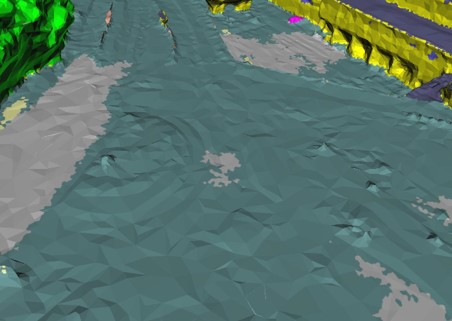} \\	
        \includegraphics[width=0.23\textwidth]{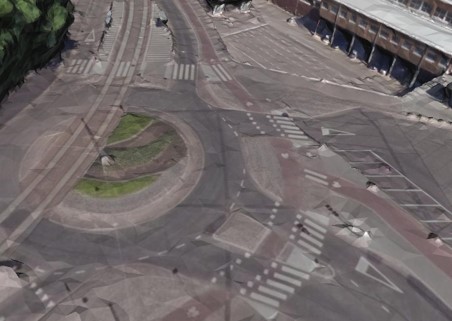} &
        \includegraphics[width=0.23\textwidth]{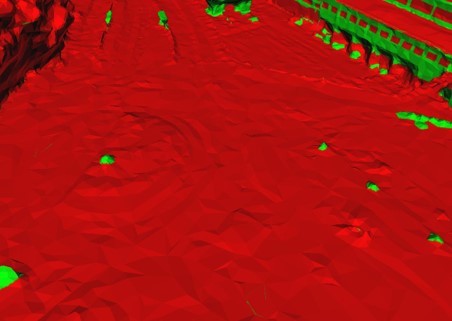} &
        \includegraphics[width=0.23\textwidth]{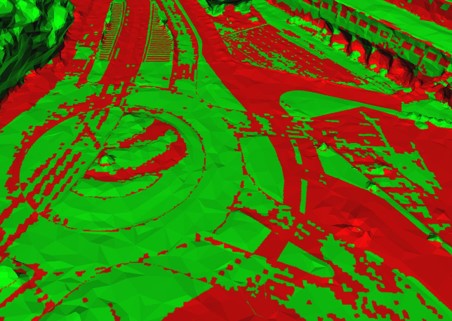} &
        \includegraphics[width=0.23\textwidth]{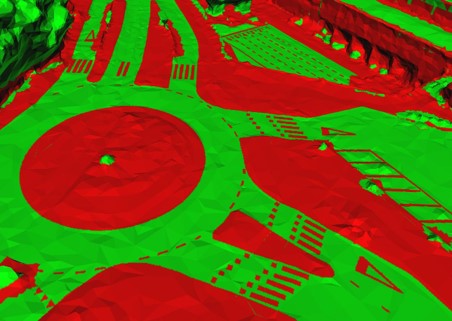} &
        \includegraphics[width=0.23\textwidth]{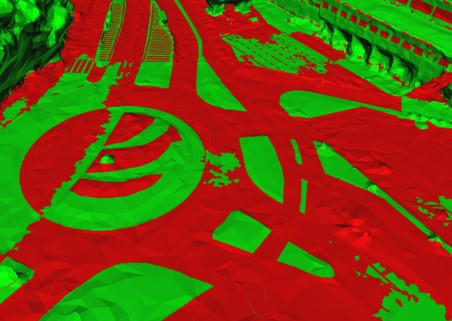} \\	
        Truth (top), Mesh (bottom)   &
		PointNet\(^{Sp.}\) &
		PointNet++\(^{Sp.}\) &
        SPG\(^{Sp.}\) & 
        SparseUNet\(^{Rd.}\) \\
        \includegraphics[width=0.23\textwidth]{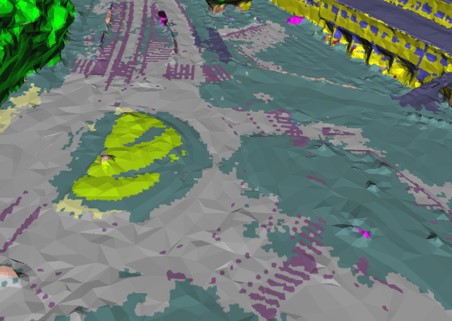} &
        \includegraphics[width=0.23\textwidth]{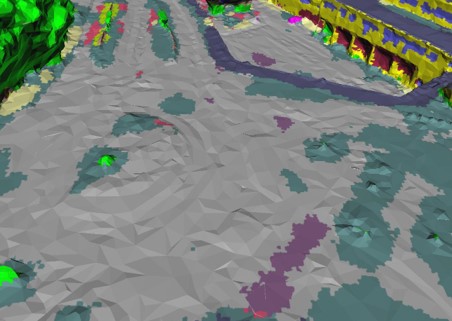} &
        \includegraphics[width=0.23\textwidth]{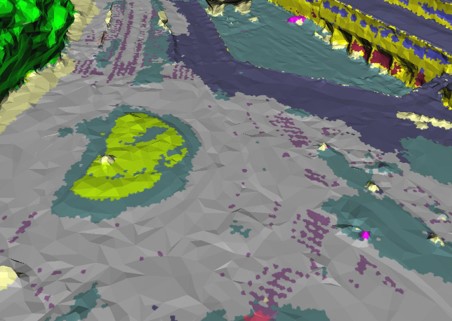} &
        \includegraphics[width=0.23\textwidth]{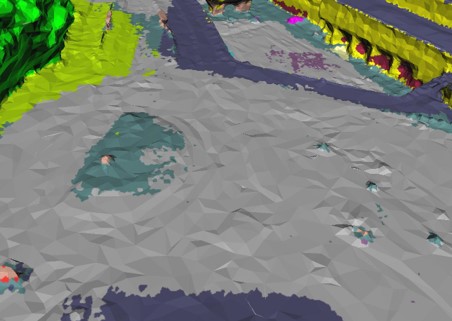} &
        \includegraphics[width=0.23\textwidth]{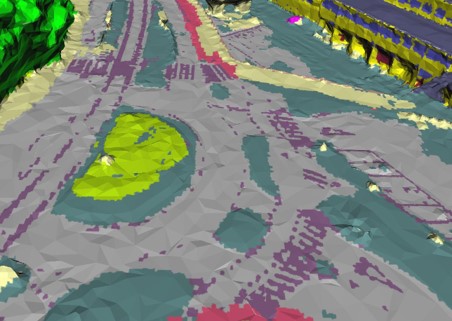} \\	
        \includegraphics[width=0.23\textwidth]{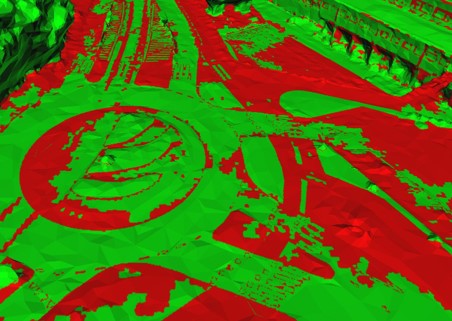} &
        \includegraphics[width=0.23\textwidth]{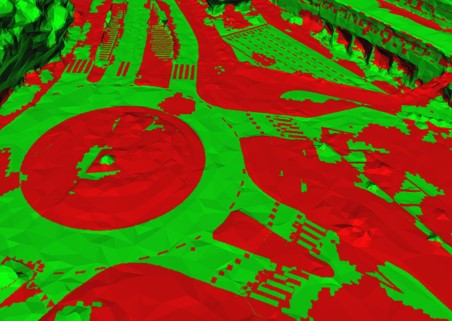} &
        \includegraphics[width=0.23\textwidth]{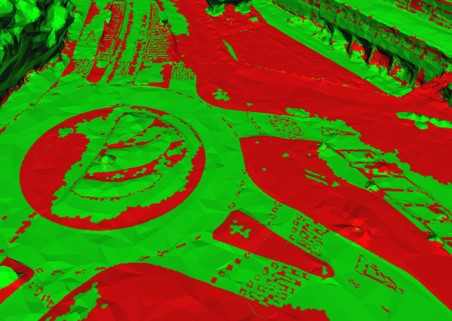} &
        \includegraphics[width=0.23\textwidth]{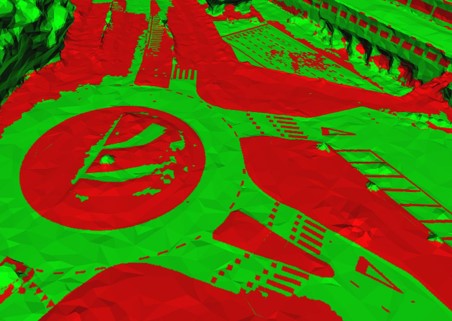} &
        \includegraphics[width=0.23\textwidth]{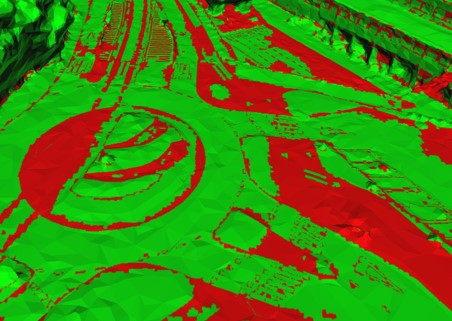} \\	
		Randla-net\(^{Sp.}\) &
		KPConv\(^{Sp.}\) &
        PointNext\(^{Po.}\) & 
        PointTransV3\(^{Rd.}\) &
        PointVector\(^{Sp.}\)\\
    \end{tabular}
    }
    \caption{Qualitative analysis of semantic segmentation and error maps in the pixel labeling track for all methods in the second scenario. \(^{Sp.}\) denotes superpixel sampling, \(^{Rd.}\) for random sampling, and \(^{Po.}\) for Poisson-disk sampling~\cite{cook1986stochastic}. The zoomed-in view direction is indicated in the input mesh image.}
    \label{fig:pixel_label_quality_scene2}
\end{figure*}

\noindent \textit{\textbf{2) Pixel labeling track.}} \cref{tab:tex_labels} provides a detailed comparison of results for all face-labeled and pixel-labeled classes.
PointVector~\cite{deng2023pointvector} surpasses other methods in all categories, particularly with pixel labels.
However, compared to the categories shared with~\cref{tab:face_labels}, the IoU results for most methods have decreased.
This is mainly because the three mesh sampling methods produce relatively uniform point clouds, failing to capture the geometric density variations inherent in adaptive meshes.
Additionally, the increase in the number of classes has exacerbated the issue of class imbalance.
We performed qualitative analyses for all methods in two scenarios, with global and zoomed-in views, as shown in~\cref{fig:pixel_label_quality_scene1} and~\cref{fig:pixel_label_quality_scene2}.
\begin{table*}[!ht]
  \centering
    \begin{tabular}{lccccccccccccccc}
    \toprule
      & \multicolumn{1}{c}{terr.} & \multicolumn{1}{c}{hveg.} & \multicolumn{1}{c}{faca.} & \multicolumn{1}{c}{wate.} & \multicolumn{1}{c}{car} & \multicolumn{1}{c}{boat} & \multicolumn{1}{c}{roof.} & \multicolumn{1}{c}{chim.} & \multicolumn{1}{c}{dorm.} & \multicolumn{1}{c}{balc.} & \multicolumn{1}{c}{roin.} & \multicolumn{1}{c}{wall} & \multicolumn{1}{c}{OA} & \multicolumn{1}{c}{mAcc} & \multicolumn{1}{c}{mIoU} \\
      \midrule
    RF\_MRF & 81.6  & 86.6  & 81.3  & 84.5  & 24.8  & 3.7  & 73.3  & 27.6  & 0.0  & 4.8  & 0.4  & 5.9  & 85.2  & 45.3  & 39.5  \\
    SUM\_RF & 84.8  & 88.1  & 84.0  & 79.0  & 42.5  & 10.6  & 77.7  & 42.4  & 3.5  & 22.2  & 4.7  & 12.7  & 86.9  & 53.6  & 46.0  \\
    PSSNet & 80.7  & 90.5  & 85.2  & 64.2  & 52.6  & 13.0  & 78.1  & 44.0  & 6.6  & 25.7  & 6.9  & 16.6  & 86.3  & 56.4  & 47.0  \\
    PoinNet\(^{Sp.}\) & 52.6  & 7.1  & 38.6  & 59.9  & 0.0  & 0.0  & 22.8  & 0.0  & 0.0  & 0.0  & 0.0  & 0.0  & 50.6  & 22.0  & 15.1  \\
    PoinNet++\(^{Sp.}\) & 67.9  & 68.7  & 59.2  & 86.1  & 24.2  & 11.1  & 51.1  & 24.9  & 0.0  & 0.0  & 3.3  & 1.1  & 69.0  & 46.9  & 33.1  \\
    SPG\(^{Sp.}\) & 53.4  & 55.3  & 62.5  & 40.5  & 27.4  & 13.1  & 64.3  & 33.9  & 5.1  & 11.3  & 3.9  & 9.9  & 64.9  & 55.0  & 31.7  \\
    SparseUNet\(^{Fc.}\) & 88.6  & 91.7  & 88.6  & 76.7  & 75.6  & 14.6  & 82.3  & 70.1  & 27.0  & 49.0  & 28.0  & 33.9  & 90.3  & 71.7  & 60.5  \\
    Randla-net\(^{Fc.}\) & 86.7  & 92.3  & 81.6  & \textbf{87.1} & 82.9  & \textbf{41.2} & 71.6  & 55.6  & 21.6  & 27.6  & 19.0  & 21.1  & 86.7  & 76.3  & 57.4  \\
    KPConv\(^{Fc.}\) & 86.9  & 90.8  & 88.3  & 81.5  & 66.4  & 16.5  & 81.9  & 66.7  & 16.1  & 45.3  & 21.2  & 28.2  & 90.1  & 64.7  & 57.5  \\
    PointNext\(^{Fc.}\) & 91.0  & 95.0  & 90.4  & 81.6  & 91.2  & 17.9  & 83.1  & 74.6  & 33.8  & 56.0  & 30.0  & 39.3  & 91.8  & 77.2  & 65.3  \\
    PointTransV3\(^{Fc.}\) & 88.6  & 90.1  & 87.9  & 78.9  & 72.1  & 16.1  & 81.0  & 66.2  & 21.4  & 45.2  & 25.0  & 36.4  & 89.9  & 70.2  & 59.1  \\
    PointVector\(^{Fc.}\) & \textbf{92.3} & \textbf{96.8} & \textbf{91.7} & 85.1  & \textbf{95.2} & 22.0  & \textbf{85.9} & \textbf{82.6} & \textbf{47.9} & \textbf{62.4} & \textbf{38.6 } & \textbf{40.0} & \textbf{93.1} & \textbf{80.7} & \textbf{70.0} \\
    \bottomrule
    \end{tabular}%
  \caption{Comparison of 3D semantic segmentation methods for face labeling using optimal sampling. Semantic categories: `terr.' (terrain), `hveg.' (high vegetation), `faca.' (facade), `wate.' (water), `roof.', `chim.' (chimney), `dorm.' (dormer), `balc.' (balcony), and `roin.' (roof installation). \(^{Fc.}\) and \(^{Sp.}\) denote face-centered and superpixel sampling, respectively. Results are presented as IoU (\%), Overall Accuracy (OA \%), mean Accuracy (mAcc \%), and mean IoU (mIoU \%). Highest values in IoU, OA, mAcc, and mIoU are highlighted in bold.}
  \label{tab:face_labels}
\end{table*}%

\begin{table*}[!t]
  \centering
  \resizebox{\textwidth}{!}{
    \begin{tabular}{lcccccccccccccccccccccc}
    \toprule
      & \multicolumn{1}{c}{\rotatebox{90}{hveg.}} & \multicolumn{1}{c}{\rotatebox{90}{faca.}} & \multicolumn{1}{c}{\rotatebox{90}{wate.}} & \multicolumn{1}{c}{\rotatebox{90}{car}} & \multicolumn{1}{c}{\rotatebox{90}{boa.}} & \multicolumn{1}{c}{\rotatebox{90}{roof.}} & \multicolumn{1}{c}{\rotatebox{90}{chim.}} & \multicolumn{1}{c}{\rotatebox{90}{dorm.}} & \multicolumn{1}{c}{\rotatebox{90}{balc.}} & \multicolumn{1}{c}{\rotatebox{90}{roin.}} & \multicolumn{1}{c}{\rotatebox{90}{wall}} & \multicolumn{1}{c}{\rotatebox{90}{wind.}} & \multicolumn{1}{c}{\rotatebox{90}{door}} & \multicolumn{1}{c}{\rotatebox{90}{lveg.}} & \multicolumn{1}{c}{\rotatebox{90}{impe.}} & \multicolumn{1}{c}{\rotatebox{90}{road}} & \multicolumn{1}{c}{\rotatebox{90}{roma.}} & \multicolumn{1}{c}{\rotatebox{90}{cycl.}} & \multicolumn{1}{c}{\rotatebox{90}{side.}} & \multicolumn{1}{c}{\rotatebox{90}{OA}} & \multicolumn{1}{c}{\rotatebox{90}{mAcc}} & \multicolumn{1}{c}{\rotatebox{90}{mIoU}} \\
      \midrule
    PoinNet\(^{Sp.}\) & 0.5  & 13.3  & 16.5  & 0.0  & 2.1  & 7.9  & 0.0  & 0.0  & 0.0  & 0.0  & 0.0  & 0.0  & 0.0  & 0.0  & 8.5  & 0.0  & 0.0  & 0.0  & 0.2  & 17.3  & 9.8  & 2.6  \\
    PoinNet++\(^{Sp.}\) & 72.7  & 47.5  & 86.4  & 34.9  & 12.4  & 52.4  & 28.1  & 0.0  & 5.3  & 5.6  & 0.4  & 13.0  & 5.0  & 42.4  & 31.2  & 14.6  & 9.5  & 0.0  & 7.3  & 55.4  & 35.2  & 24.7  \\
    SPG\(^{Sp.}\) & 58.2  & 50.8  & 18.4  & 24.1  & 2.7  & 60.4  & 39.9  & 3.1  & 13.6  & 4.4  & 10.5  & 2.4  & 4.0  & 13.4  & 14.6  & 31.0  & 0.0  & 1.7  & 12.1  & 51.5  & 34.5  & 19.2  \\
    SparseUNet\(^{Rd.}\) & 88.8  & \textbf{70.0}  & 5.9  & 51.6  & 2.5  & \textbf{79.8}  & 55.0  & 12.3  & \textbf{45.4}  & 22.6  & \textbf{31.5}  & 32.0  & 12.3  & 15.2  & 43.8  & 44.6  & 5.2  & 0.6  & 35.8  & 72.9  & 45.1  & 34.5  \\
    Randla-net\(^{Sp.}\) & 90.5  & 60.9  & 84.6  & 67.6  & \textbf{22.7}  & 74.7  & 53.3  & 0.6  & 29.3  & 16.2  & 26.3  & 33.4  & 12.8  & 59.8  & 48.8  & 50.2  & 31.5  & 0.0  & 37.1  & 73.5  & 57.7  & 42.1  \\
    KPConv\(^{Sp.}\) & 84.0  & 68.5  & 81.7  & 68.6  & 21.8  & 78.2  & \textbf{66.4}  & \textbf{25.0}  & 41.8  & \textbf{29.6}  & 31.5\(^{\ast}\)  & 36.1  & 14.9  & 21.4  & 35.8  & 50.0  & 7.3  & 13.4  & 34.1  & 74.4  & 58.3  & 42.6  \\
    PointNext\(^{Po.}\) & 90.1  & 66.2  & 87.9  & 68.1  & 16.3  & 74.5  & 59.7  & 14.9  & 35.6  & 19.1  & 31.0  & 33.2  & 13.7  & 55.5  & 51.4  & 55.5  & 29.0  & 6.9  & 40.0  & 76.0  & 57.6  & 44.7  \\
    PointTransV3\(^{Rd.}\) & 85.9  & 59.9  & 74.6  & 64.7  & 17.8  & 75.9  & 58.7  & 15.3  & 37.2  & 16.2  & 29.3  & 11.8  & 7.9  & 27.1  & 43.3  & 51.5  & 3.5  & 7.2  & 33.4  & 70.6  & 54.1  & 38.0  \\
    PointVector\(^{Sp.}\) & \textbf{92.7}  & 66.6  & \textbf{92.0}  & \textbf{70.2}  & 19.8  & 76.8  & 60.8  & 21.8  & 37.0  & 20.6  & 30.8  & \textbf{37.1}  & \textbf{16.5}  & \textbf{59.8}  & \textbf{53.9}  & \textbf{57.4}  & \textbf{35.0}  & \textbf{16.4}  & \textbf{45.0}  & \textbf{77.0}  & \textbf{63.8}  & \textbf{47.9}  \\
    \bottomrule
    \end{tabular}%
    }
    \caption{Comparison of 3D semantic segmentation methods for pixel labeling using optimal sampling strategies: `hveg.' (high vegetation), `faca.' (facade surface), `wate.' (water), `roof.' (roof surface), `chim.' (chimney), `dorm.' (dormer), `balc.' (balcony), `roin.' (roof installation), `wind.' (window), `lveg.' (low vegetation), `impe.' (impervious surfaces), `roma.' (road marking), `cycl.' (cycle lane), and `side.' (sidewalk). Additionally, \(^{Sp.}\) denotes superpixel sampling, \(^{Rd.}\) for random sampling, and \(^{Po.}\) for Poisson-disk sampling~\cite{cook1986stochastic}. Results are presented as IoU (\%), Overall Accuracy (OA \%), mean Accuracy (mAcc \%), and mean IoU (mIoU \%). \(^{\ast}\)KPConv’s IoU for wall is 31.46\%, slightly below SparseUnet’s 31.47\%. Highest values in IoU, OA, mAcc, and mIoU are highlighted in bold.}
   \label{tab:tex_labels}
\end{table*}%

\section{Comparison of related datasets}
\paragraph{Compare with SUM.} 
Our proposed SUM Parts dataset extends beyond SUM's object-level labels~\cite{gao2021sum}, offering three key benefits: \textbf{(1)} finer geometric analysis, such as evaluating heat loss at the window-level rather than at the building-scale; \textbf{(2)} support for part-aware tasks, e.g., drone navigation for precise delivery by localizing windows, doors, and rooftop solar panel planning; \textbf{(3)} seamless integration with urban digital twins and BIM workflows.

\paragraph{Compare with KITTI-360.} KITTI-360~\cite{kitti360_2023} focuses on street-view LiDAR-image fusion for autonomous driving, providing 37 Cityscapes-aligned classes, including road-accessible static and dynamic objects (\(\geq\)0.1m resolution) labeled via manual selection and trajectory-based matching. In contrast, SUM Parts addresses broader urban planning and sustainability challenges using oblique photogrammetry meshes. Key differences include: 
\textbf{(1)} Labeling granularity: SUM Parts offers both object- and part-level annotations (21 CityGML-aligned classes) for fine-grained urban infrastructure details.
\textbf{(2)} Annotation tools: Our mesh-texture semi-automatic selection tools (click, stroke, lasso) with 2D/3D template matching ensure efficient annotation.
\textbf{(3)} Coverage: SUM Parts provides full-city coverage, annotating all static objects (\(\geq\)0.5m resolution), including vehicle-inaccessible areas.
Hence, SUM Parts complements KITTI-360 for broader urban applications.

\end{document}